\documentclass{article} % For LaTeX2e
\usepackage{iclr2025_conference,times}

\usepackage{amsmath,amsfonts,bm}

\def\eqref#1{equation~\ref{#1}}
\def\1{\bm{1}}

\def\vs{{\bm{s}}}

\DeclareMathAlphabet{\mathsfit}{\encodingdefault}{\sfdefault}{m}{sl}
\SetMathAlphabet{\mathsfit}{bold}{\encodingdefault}{\sfdefault}{bx}{n}

\usepackage[utf8]{inputenc} % allow utf-8 input
\usepackage[T1]{fontenc}    % use 8-bit T1 fonts
\usepackage{url}
\usepackage{booktabs}       % professional-quality tables
\usepackage{amsfonts}       % blackboard math symbols
\usepackage{nicefrac}       % compact symbols for 1/2, etc.
\usepackage{microtype}      % microtypography

\usepackage{soul}
\usepackage{lipsum}
\usepackage{fancyhdr}

\usepackage{eccvabbrv}
\usepackage{subfig}
\usepackage{amsmath}
\usepackage{amssymb}
\usepackage{mathtools}
\usepackage{amsthm}
\usepackage{multirow}
\usepackage{colortbl}
\usepackage{adjustbox}

\usepackage{graphicx}
\usepackage[ruled]{algorithm2e}
\usepackage{upgreek}
\usepackage{wrapfig}
\usepackage{xcolor}
\usepackage{pifont}
\usepackage{graphicx}
\usepackage{multirow}

\theoremstyle{definition}

\usepackage[ruled]{algorithm2e}
\definecolor{cvprblue}{rgb}{0.21,0.49,0.74}
\definecolor{greenx}{RGB}{0,128,128}
\definecolor{maroonx}{RGB}{195,18,48}
\usepackage[colorlinks=True,
            linkcolor=maroonx,
            anchorcolor=blue,  
            pagebackref,
            citecolor=cvprblue,
            ]{hyperref}
\usepackage{enumitem}
\usepackage{tcolorbox}
\setlist{leftmargin=5.5mm}
\usepackage{wrapfig}
\usepackage{textgreek}
\usepackage{bbm}

\newcommand{\cc}{\cellcolor{gray!20}}
\setlist[itemize]{leftmargin=*,topsep=0em}
\definecolor{darkred}{RGB}{192, 0, 0}
\definecolor{darkgreen}{RGB}{103, 174, 64}

\newcommand{\RebuttalRevision}[1]{\textcolor{black}{#1}}

\title{Self-Correcting Decoding with Generative Feedback for Mitigating Hallucinations in Large Vision-Language Models}
\author{Ce Zhang\thanks{Equal contribution. \, $^{\dag}$Contact: {\small \texttt{\{cezhang, zifuw, yaqix\}@cs.cmu.edu}}.}\hspace{0.45em}\quad Zifu Wan\footnotemark[1]\hspace{0.45em}\quad Zhehan Kan\quad Martin Q. Ma\quad Simon Stepputtis\\
\textbf{Deva Ramanan\quad Russ Salakhutdinov\quad Louis-Philippe Morency\quad Katia Sycara\quad Yaqi Xie} \\[3pt]
School of Computer Science, Carnegie Mellon University
}

\iclrfinalcopy % Uncomment for camera-ready version, but NOT for submission.
\begin{document}

\maketitle
\vspace{-12pt}
\begin{abstract}

\looseness=-1
\vspace{-5pt}
While recent Large Vision-Language Models (LVLMs) have shown remarkable performance in multi-modal tasks, they are prone to generating hallucinatory text responses that do not align with the given visual input, which restricts their practical applicability in real-world scenarios. In this work, inspired by the observation that the text-to-image generation process is the inverse of image-conditioned response generation in LVLMs, we explore the potential of leveraging text-to-image generative models to assist in mitigating hallucinations in LVLMs. We discover that generative models can offer valuable self-feedback for mitigating hallucinations at both the response and token levels. Building on this insight, we introduce self-correcting Decoding with Generative Feedback (DeGF), a novel training-free algorithm that incorporates feedback from text-to-image generative models into the decoding process to effectively mitigate hallucinations in LVLMs. Specifically, DeGF generates an image from the initial response produced by LVLMs, which acts as an auxiliary visual reference and provides self-feedback to verify and correct the initial response through complementary or contrastive decoding. Extensive experimental results validate the effectiveness of our approach in mitigating diverse types of hallucinations, consistently surpassing state-of-the-art methods across six benchmarks. Code is available at \url{https://github.com/zhangce01/DeGF}.
\end{abstract}

\vspace{-10pt}

\section{Introduction}
\vspace{-4pt}
\label{sec:intro}
% \looseness=-1
Large Vision-Language Models (LVLMs) have demonstrated remarkable performance across various multi-modal tasks, such as image captioning and visual question answering, by extending the capabilities of powerful Large Language Models (LLMs) to incorporate visual inputs~\citep{liu2023visual,li2023blip,dai2024instructblip,bai2023qwen,ye2024mplug}. Despite their proficiency in interpreting both visual and textual modalities, these models often suffer from \textit{hallucinations}, where LVLMs erroneously produce responses that are inconsistent with the visual input~\citep{li2023evaluating,gunjal2024detecting,yin2023woodpecker,wu2024evaluating}.
% in their generated text response.
% \yaqi{Question: Only hallucinations in generated text response? not in generated image response? We need to find a way to scope it down naturally. Or maybe define LVLM, does the stable diffusion we used belog to LVLM?} 
% Specifically, this refers to the phenomenon where LVLMs erroneously produce responses that are inconsistent with the visual input~\citep{li2023evaluating,gunjal2024detecting,yin2023woodpecker}. 
This potential for misinformation raises significant concerns, limiting the models' reliability and restricting their broader deployment in real-world scenarios~\citep{liu2024survey,bai2024hallucination,chen2024detecting,zhao2024fact}.

\looseness=-1
Recent research has revealed that a major cause of hallucinations in LVLMs is the over-reliance on language priors due to biased training sets, which can override the visual content in response generation~\citep{bai2024hallucination,liu2024survey,leng2024mitigating}. In response, various strategies have been developed to detect and mitigate these hallucinations by directly introducing additional training~\citep{chen2024alleviating,sun2023aligning,jiang2024hallucination, chen2023mitigating,zhang2024reflective}, demonstrating promising results in reducing over-reliance. However, the need for additional data and costly training processes hinders their deployment in downstream tasks. 
More recently, a new paradigm of methods has emerged to tackle the hallucination problem in LVLMs by intervening in the decoding process~\citep{huang2024opera,deng2024seeing,kim2024code}. Among these, recent training-free contrastive decoding-based methods~\citep{li2023contrastive} have proven effective in mitigating undesired hallucinations by contrasting token predictions derived from original visual input with bias-inducing counterparts, such as no/distorted visual input~\citep{favero2024multi,leng2024mitigating}, disturbed instructions~\citep{wang2024mitigating2}, or premature layers~\citep{chuang2024dola}. 
% \yaqi{This visual cue part can be further emphasized. Maybe we can start a new paragraph here and merge the first half of the original paragarph to the previous one.}

While these contrastive decoding-based methods effectively mitigate hallucinations arising from language priors, we recognize that hallucinations can also originate beyond language bias, stemming from visual deficiencies in LVLMs~\citep{tong2024eyes}. For instance, in counting hallucinations, language does not imply any count information; instead, miscounts largely arise from visual recognition errors of LVLMs, as complex scenes include numerous, similar objects at ambiguous positions which may confuse the LVLMs, leading to incorrect visual understanding and, consequently, hallucinated answers. Therefore, we argue that current contrastive decoding-based methods may struggle to generalize effectively across different types of hallucinations.

\looseness=-1
In this work, we explore the potential of leveraging powerful text-to-image generative models (\eg, Stable Diffusion~\citep{rombach2021highresolution,podell2024sdxl}) to mitigate various types of hallucinations in LVLMs. 
Our work is based on a simple yet intuitive hypothesis: Given a visual input and a textual prompt to an LVLM, if the generated response conditioned on the original image is accurate and non-hallucinatory,  a text-to-image generative model should be capable of reversing this process to produce a similar image from that response. 
Alternatively, if there is a discrepancy between the original image and the one generated from the response, this difference can serve as valuable self-feedback, guiding the decoding process to correct potential hallucinations in the initial response.
% Therefore, the discrepancy between the original image and the generated image can serve as valuable self-feedback, guiding the decoding process to correct potential hallucinations in the initial response.
% \martin{Our approach provides token-level feedback to correct hallucination, a further improvement from image-level contrastive approaches.}
To verify this hypothesis, we conduct an empirical study (in Section~\ref{sec:diffusion}), demonstrating that \textit{generative models can provide valuable self-feedback for mitigating hallucinations at both the response and token levels}.
% \yaqi{``any'' seems too strong. There are a lot of discrepancies, but we are only using the relevant ones. And we want to also hint that, but only the discrepancy but also the similarity values}

% Our approach is based on a simple yet intuitive assumption \yaqi{assumption -> hypotheses. Simplify the assumption sentence if you could.}: given a visual input $v$ and a textual prompt $\mathbf{x}$ that generates an image-conditioned response $\boldsymbol{\tau}$ from an LVLM, a text-to-image generative model should be capable of reversing this process to produce a similar image $v'$ based on $\boldsymbol{\tau}$. We hypothesize that the discrepancy between the original image $v$ and the generated image $v'$ can serve as valuable self-feedback, guiding the decoding process to correct potential hallucinations in the initial response. 
% To verify this hypothesis, we conduct an empirical study (presented in Section~\ref{sec:diffusion}), which demonstrates that generative models can function as zero-shot hallucination detectors, offering feedback at both the response and token levels.
% To the best of our knowledge, we are the first work to explore the use of generative models for mitigating hallucinations in LVLMs.

\begin{figure}[t]
  \begin{center}
     \makebox[\textwidth]{\includegraphics[width=\textwidth]{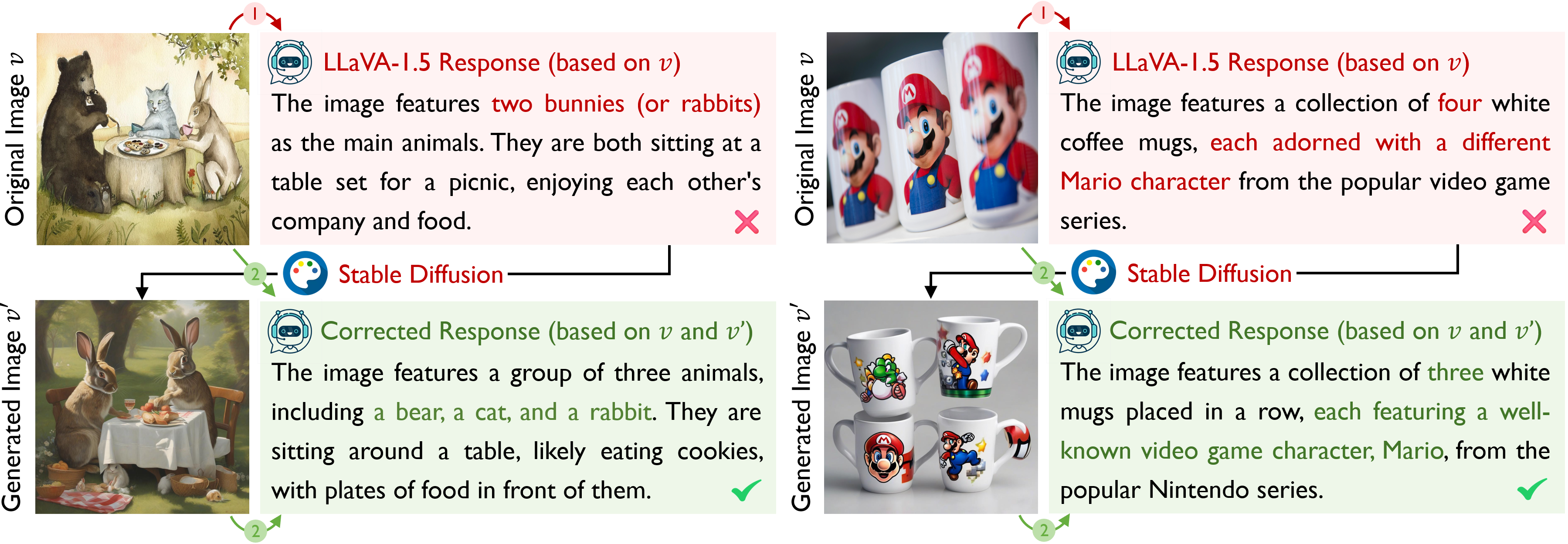}}
  \end{center}
  \vspace{-14pt}
  \caption{\looseness=-1 
  % \yaqi{The numbers are confusing as they are. Please explain what they mean in the caption. If they are used to denote the time steps, we probably need to add a 2 for stable diffusion.} 
  \textbf{Generative models can visualize and help correct various types of hallucinations in the initial response}. \raisebox{-0.07cm}{\includegraphics[height=0.35cm]{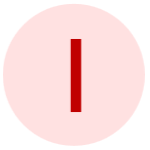}} In the first query, we provide LLaVA-1.5~\citep{liu2023visual} with the prompt ``\texttt{Describe this image in detail}'' to produce captions for two examples from LLaVA-Bench. Based on the initial response, we utilize Stable Diffusion XL~\citep{podell2024sdxl} to generate a new image $v'$, which effectively highlights hallucinations and provides valuable self-feedback. \raisebox{-0.07cm}{\includegraphics[height=0.35cm]{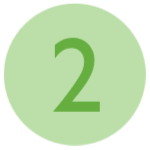}} In the second query, our approach incorporates both the original image $v$ and the generated image $v'$ into the decoding process, using the feedback to successfully correct various types of hallucinations.}
  \label{fig:intro}
   \vspace{-10pt}
\end{figure}

Building on this insight, we introduce self-correcting Decoding with Generative Feedback (DeGF), a novel training-free decoding algorithm that effectively incorporates feedback from text-to-image generative models to recursively enhance the accuracy of LVLM responses. 
Specifically, for each instance, we generate a new image based on the initial response, which serves as an \textit{auxiliary visual reference} to assess and verify the accuracy of the initial output.
% we treat the generated image based on the initial response as an \textit{auxiliary visual reference} and re-query the LVLM using both the original image and this reference. 
We propose self-correcting decoding that either enhances or contrasts predictions from the original and this reference based on the auxiliary visual reference, \textit{confirming} or \textit{revising} the initial LVLM response based on the degree of divergence between the two predictions. By integrating this additional visual reference and generative feedback, LVLMs can gain enhanced visual insights and verify the initial response to ensure accurate visual details in the text outputs.
% \martin{(define ``double-check''?)}
In Figure~\ref{fig:intro}, we demonstrate that incorporating generative feedback in our approach can reduce various types of hallucinations, including object existence, visual appearance, counting, \etc. To the best of our knowledge, we are the first work to explore the use of text-to-image generative feedback as a self-correcting mechanism for mitigating hallucinations in LVLMs.

% the use of generative models \martin{to generate visual reference} for mitigating hallucinations in LVLMs.

\looseness=-1
The effectiveness of DeGF is evaluated on LLaVA-1.5, InstructBLIP, and Qwen-VL across six benchmarks: POPE~\citep{li2023evaluating}, CHAIR~\citep{rohrbach2018object}, MME-Hallucination~\citep{fu2023mme}, MMBench~\citep{liu2025mmbench}, MMVP~\citep{tong2024eyes}, and LLaVA-Bench. Extensive experimental results validate the effectiveness of our DeGF in mitigating various types of hallucinations in LVLMs. Qualitative case studies and GPT-4V-aided evaluation on LLaVA-Bench further demonstrate that our approach enhances both the accuracy and detailedness of the LVLM responses. 

% \clearpage
The contributions of this paper are summarized as follows:
\begin{itemize}
    % \item We conduct a pilot study \yaqi{``pilot study'' sounds preliminary. The subject this contribution should be the discovery (we discovered xxx), but the pilot study. We don't need to mention it in the contribution.} discovering the potential of generative models in reflecting hallucinations in LVLMs and demonstrate that generative models can function as zero-shot hallucination detectors, offering feedback at both the response and token levels.
    \item We investigate the potential of text-to-image generative models in mitigating hallucinations in LVLMs and demonstrate that text-to-image generative models can provide valuable self-feedback for mitigating hallucinations at both the response and token levels.
    \item We propose self-correcting Decoding with Generative Feedback (DeGF), a novel training-free decoding algorithm for LVLMs that recursively enhances the accuracy of responses by integrating feedback from text-to-image generative models with complementary/contrastive decoding.
    \item Extensive experimental evaluations across six benchmarks demonstrate that our DeGF consistently outperforms state-of-the-art approaches in effectively mitigating hallucinations in LVLMs.
\end{itemize}

% \clearpage
\vspace{-3pt}
\section{Related Work}
\vspace{-2pt}
\looseness=-1
\textbf{Hallucination in LVLMs}.
% With the recent advances in LLMs' linguistic capabilities~\citep{touvron2023llama,chowdhery2023palm,brown2020language, chiang2023vicuna} and the growing potential of VLMs in multi-modal understanding~\citep{li2019visualbert,li2022blip,radford2021learning}, LVLMs have emerged as a powerful fusion of the two~\citep{chen2023shikra,liu2023visual,zhu2023minigpt,driess2023palm,alayrac2022flamingo}. 
% By training a modality connection module to align visual tokens with the textual embedding space of the LLM, LVLMs demonstrate unified decoding of visual and textual tokens, enabling their broad application in multimodal tasks such as visual question answering~\citep{liu2024survey}.
With advances of autoregressive LLMs~\citep{touvron2023llama,chowdhery2023palm,chiang2023vicuna}, researchers have extended these powerful models to process visual inputs, leading to the development of LVLMs~\citep{liu2023visual,dai2024instructblip,bai2023qwen,ye2024mplug}. These models typically train a modality alignment module to project visual tokens into the textual embedding space of the LLM, demonstrating impressive performance in various multi-modal tasks such as visual question answering and image captioning~\citep{liu2024survey,bai2024hallucination}.
However, LVLMs are prone to hallucinations, where contradictions arise between the visual content and the generated textual response~\citep{li2023evaluating, liu2024survey, bai2024hallucination}.

\looseness=-1
To mitigate hallucinations in LVLMs, early works have introduced various approaches, including reinforcement learning from human feedback (RLHF)~\citep{gunjal2024detecting, sun2023aligning}, applying auxiliary supervision~\citep{jiang2024hallucination, chen2023mitigating}, incorporating negative~\citep{liu2023mitigating} or noisy data~\citep{yue-etal-2024-less}, and training post-hoc revisors for correction~\citep{zhou2024analyzing, yin2023woodpecker}. Despite promising results, these methods often lack practicality due to their reliance on additional data and costly training processes. To address this, another line of work focuses on training-free methods that can be seamlessly integrated into existing LVLMs.
Such methods encompass contrastive decoding~\citep{leng2024mitigating, favero2024multi} and guided decoding with auxiliary information~\citep{chen2024halc,deng2024seeing,woo2024ritual}. 
In this work, we present a novel training-free approach that recursively enhances the accuracy of the LVLM response by incorporating text-to-image generative feedback. 
To the best of our knowledge, we are the first work to effectively utilize feedback from text-to-image generative models to mitigate hallucinations in LVLMs.

% With the recent advances in LLMs' linguistic capabilities~\citep{touvron2023llama,chowdhery2023palm,brown2020language, chiang2023vicuna} and the growing potential of VLMs in multi-modal understanding~\citep{li2019visualbert,li2022blip,radford2021learning}, LVLMs have emerged as a powerful fusion of the two~\citep{chen2023shikra,liu2023visual,zhu2023minigpt,driess2023palm,alayrac2022flamingo}. 
% By training a modality connection module to align visual tokens with the textual embedding space of the LLM, LVLMs demonstrate unified decoding of visual and textual tokens, enabling their broad application in multi-modal tasks such as visual question answering~\citep{liu2024survey}.

% However, the aforementioned methods fail to consider both the inclusion of reasonable visual information and the adaptation of token decoding. In contrast, \Ours{} leverages generated images as effective visual references, explicitly assisting LVLMs in mitigating hallucinations.

% \todo{Mention previous work: Img-Diff: Contrastive Data Synthesis for Multimodal Large Language Models}

\textbf{Text-to-Image Synthesis}.
Text-to-image synthesis aims to create realistic images from textual descriptions~\citep{zhu2019dm, ge2023expressive}.
In recent years, significant progress has been achieved in this area, largely due to the advent of deep generative models~\citep{zhan2023multimodal, goodfellow2014generative}. 
These advances include Generative Adversarial Networks (GAN)~\citep{sauer2023stylegan, kang2023scaling}, autoregressive models~\citep{chang2023muse, yu2022scaling}, and diffusion models~\citep{ho2020denoising,karras2022elucidating,nichol2022glide,saharia2022photorealistic, rombach2021highresolution}.
Among these, diffusion-based methods have been particularly distinguished due to their ability to generate high-quality, detailed images with fine-grained control over the synthesis process~\citep{yang2023diffusion,croitoru2023diffusion}.
Pre-trained on large-scale text-image datasets such as LAION~\citep{schuhmann2022laion}, diffusion-based methods have demonstrated strong vision-language alignment, making them valuable for downstream tasks such as classification~\citep{li2023your} and semantic segmentation~\citep{amit2021segdiff, wolleb2022diffusion}. 
% \zifu{More recently, diffusion models have been used to generate contrastive data for fine-tuning LVLMs, resulting in comprehensive improvements~\citep{jiao2024img}. However, these improvements stem from the data itself rather than advancements in LVLMs.} \martin{Previous sentence needs clarification.}

More recently, \citet{jiao2024img} incorporate text-to-image generative models to enhance fine-grained image recognition in LVLMs by introducing the Img-Diff dataset, which generates pairs of similar images using Stable Diffusion XL~\citep{podell2024sdxl}. Their results demonstrate that fine-tuning LVLMs with this additional data leads to improved performance on several VQA tasks.
In contrast, in this work, we directly leverage a pre-trained diffusion model to provide valuable self-feedback for refining the generated responses of LVLMs in the decoding process, dynamically improving the accuracy and consistency of the model’s response without modifying the underlying LVLMs. 
% Through empirical evaluations, we demonstrate that with generative feedback, our proposed training-free DeGF method effectively mitigates various types of hallucinations in LVLMs.
% \input{sections/diffusion}
\vspace{-3pt}
\section{Method}
\begin{figure}[t]
  \begin{center}
     \makebox[\textwidth]{\includegraphics[width=\textwidth]{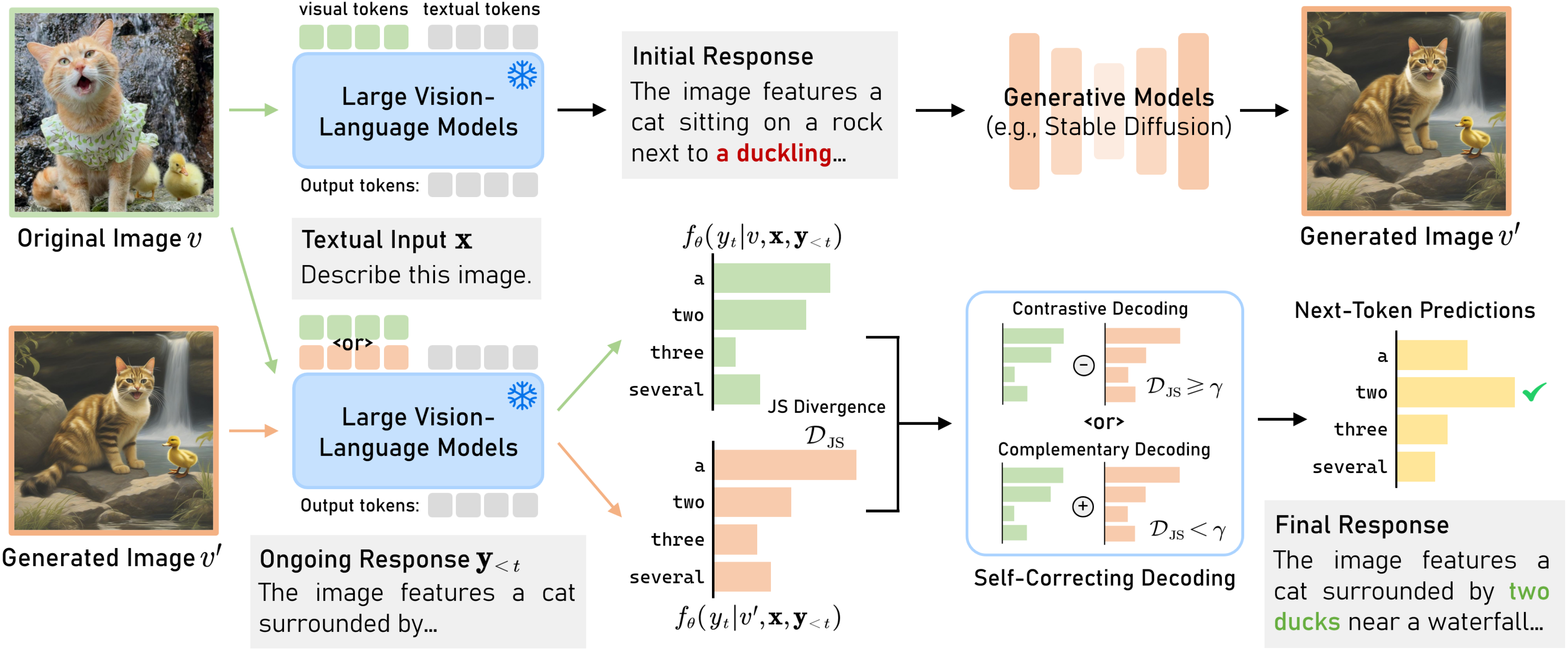}}
  \end{center}
  \vspace{-10pt}
  \caption{\textbf{Overview of our proposed DeGF}. Our method follows a two-step process: first, a generative model produces a high-quality image based on the initial response; second, this image acts as an auxiliary visual reference, providing feedback to refine the next-token predictions. Additionally, we introduce self-correcting decoding, which either enhances or contrasts the next-token predictions conditioned on the original and generated images to mitigate hallucinations in the LVLM response.} %\martin{(In the figure, we could make it clear it is a two step process by labeling step 1 and step 2, and separate top and bottom using dashed line.)}}
  \label{fig:overview}
\end{figure}

\vspace{-2pt}
\label{sec:method}
In this work, we present DeGF, a novel training-free algorithm that recursively improves the accuracy of LVLM responses using text-to-image generative feedback, as illustrated in Figure~\ref{fig:overview}. 

\subsection{Preliminary: Decoding of LVLMs}
\looseness=-1
We consider an LVLM parameterized by $\theta$, which processes an input image $v$ and a textual query $\mathbf{x}$, aiming to autoregressively generate a fluent sequence of textual responses $\mathbf{y}$. The visual input $v$ is first processed by a vision encoder and then projected into visual tokens within the textual input space using a vision-language alignment module (\eg, Q-Former~\citep{li2023blip} or linear projection~\citep{liu2023visual}). These visual tokens, along with the textual query tokens, are then fed into the language encoder for conditioned autoregressive generation. We denote the autoregressive generation process as
\begin{equation}
y_t \sim p_{\theta}(y_t | v, \mathbf{x}, \mathbf{y}_{<t}) \propto \exp f_{\theta}(y_t | v, \mathbf{x}, \mathbf{y}_{<t}),
\end{equation}
where $y_t$ represents the token at time step $t$, $\mathbf{y}_{<t} \triangleq [y_0, \dots, y_{t-1}]$ denotes the sequence of tokens generated before time step $t$, and $f_{\theta}$ is the logit distribution (unnormalized log-probabilities) produced by the LVLM over a vocabulary of textual tokens $\mathcal{V}$. At each step $t \in [0, \dots, T]$, the response token $y_t$ is sampled from the probability distribution $p_{\theta}(y_t | v, \mathbf{x}, \mathbf{y}_{<t})$, and this generative process continues iteratively until the response sequence $\mathbf{y} \triangleq [y_0, \dots, y_{T}]$ is complete.

     % \vspace{-10pt}

\subsection{Visual Reference Generation}
\label{sec:diffusion}
In our method, we incorporate generative feedback from diffusion models to guide the decoding process. Specifically, given a visual input $v$ and a textual query $\mathbf{x}$, we first prompt the LVLMs to generate an initial response $\boldsymbol{\tau}$, which includes relevant descriptions of the visual input with potential hallucinations. Subsequently, we leverage a pre-trained diffusion model $\mathcal{G}$ to generate a new image $v'$ based on the initial response:
\begin{equation}
v' = \mathcal{G}(\boldsymbol{\tau}, x_T), \quad\text{where} \,\,x_T \sim \mathcal{N}(0, \mathbf{I}).
\end{equation}
Here, $x_T$ denotes a sample from the standard Gaussian distribution, which serves as the initial noisy input to the diffusion model. Starting from this pure noise image $x_T$, the diffusion model $\mathcal{G}$ iteratively applies $T$ steps of the denoising process to obtain $x_T, x_{T-1}, \dots, x_0$, where the final output $x_0$ corresponds to the final generated image $v'$. Through this diffusion process, the generative model visualizes the initial response, providing a visual reference that helps mitigate potential hallucinations and produce a more accurate and consistent output.

\begin{wrapfigure}[18]{r}{0.6\textwidth}
  \begin{center}
  \vspace{-3pt}
    \includegraphics[width=\linewidth]{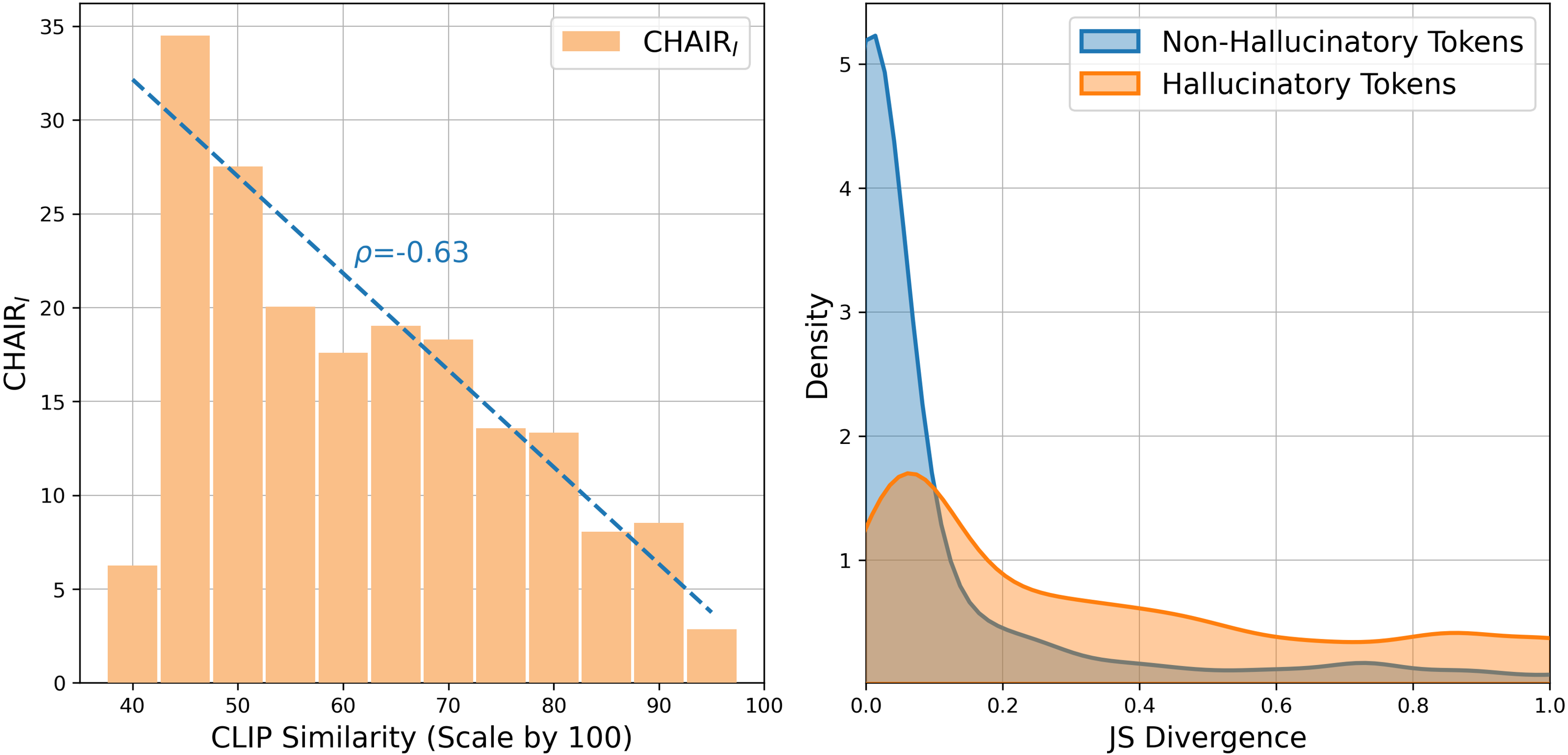}
  \end{center}
  \vspace{-10pt}
  \caption{\textbf{Text-to-image generative models can provide feedback for reflecting hallucinations}. (\textit{Left}) Bar plot of average $\text{CHAIR}_I$ scores binned by CLIP similarity (scaled by 100) on the CHAIR benchmark; (\textit{Right}) Density plots of token-level JS divergence for both hallucinatory and non-hallucinatory tokens on the POPE benchmark.}
  \label{fig:feedback}
\end{wrapfigure}

\textbf{Effectiveness of Text-to-Image Generative Models in Reflecting Hallucinations}. 
% \martin{Further clarify motivation, setup, and implication of results in this section can be crucial and can greatly improve chance of acceptance.}
We validate the effectiveness of generative models in reflecting hallucinations through an empirical study, as shown in Figure~\ref{fig:feedback}.\footnote{\RebuttalRevision{For Figure~\ref{fig:feedback}, we evaluate 1,000 CHAIR samples (\textit{Left}) and 3,000 POPE samples (\textit{Right}).}} The experimental results demonstrate that \textit{text-to-image generative models can provide valuable self-feedback for mitigating hallucinations} at both the response and token levels.

\looseness=-1
We conduct the following two experiments:
(1) We generate an image $v'$ using diffusion model based on the initial caption provided by LLaVA-1.5 and compute the CLIP image similarities between the original image $v$ and the generated image $v'$ using OpenCLIP~\citep{cherti2023reproducible} ViT-H/14 backbone. Following prior work, we use the CHAIR~\citep{rohrbach2018object} benchmark, a rule-based metric on MS-COCO~\citep{lin2014microsoft} for evaluating object hallucination from generated captions. We report the average per-instance metric $\text{CHAIR}_I$ within each bin of CLIP similarity, which evaluates the object hallucination rates in the entire initial response. 
As shown in Figure~\ref{fig:feedback} (\textit{Left}), a clear negative correlation between hallucination rates and CLIP similarities is observed (with a correlation coefficient of $\rho=-0.63$).  This indicates that \textit{lower similarity between the original image and generated image corresponds to higher rates of hallucinations at the response level}.
(2) Similarly, we generate an image $v'$ based on the initial response given by LLaVA-1.5 for each instance on the POPE~\citep{li2023evaluating} benchmark. In Figure~\ref{fig:feedback} (\textit{Right}), we present the density plot of Jensen-Shannon (JS) divergence between the predicted probabilities for both images, \ie, $p_{\theta}(y_t | v, \mathbf{x}, \mathbf{y}_{<t})$ and $p_{\theta}(y_t | v', \mathbf{x}, \mathbf{y}_{<t})$, for hallucinatory and non-hallucinatory tokens.\footnote{Note that POPE benchmark contains yes-or-no questions about object existence. In this experiment, we evaluate only the first response token (\ie, \texttt{yes} or \texttt{no}) to determine the presence of hallucinations.} The results show that the density of JS divergence follows a long-tail distribution, with hallucinatory tokens exhibiting significantly longer tails and higher JS divergence. This shows \textit{JS divergence between probabilities derived from the original and the generated image corresponds well to hallucinations at the token level.}
% \martin{Alternative: \textit{``This shows JS divergence between probabilities of the original and the generated image corresponds well to hallucinations at the token level.''}}
% \martin{Needs clarification.}
These observations provide insights into the effectiveness of generative models in reflecting hallucinations, and motivate us to incorporate the generative feedback during the decoding process.

% As a preliminary experiment, we first describe the images in the CHAIR benchmark~\citep{rohrbach2018object} using an LVLM. Based on these generated descriptions, we employ a text-to-image generative model to synthesize images that closely align with the descriptions. We then calculate the CLIP similarity~\citep{radford2021learning} between the original images and the synthesized images. To assess the level of hallucination, we compute the $\text{CHAIR}_I$ and $\text{CHAIR}_S$ metrics (as described in Eq.~\ref{eq:chair metrics}) for each image and visualize the results in a bar chart. As shown in Fig.~\ref{fig:feedback}, a strong correlation is evident between the hallucination level and the image similarity. Higher CLIP scores, indicating greater similarity, are associated with lower levels of hallucination.

\subsection{Self-Correcting Decoding with Generative Feedback}
\label{sec:token}
In this section, we focus on effectively utilizing generative feedback during the decoding process to mitigate potential hallucinations. Specifically, we propose a self-correcting decoding approach that leverages generative feedback to \textit{confirm} or \textit{revise} the initial response by selectively enhancing or contrasting the logits for each generated token based on the measured divergence between the two predicted probability distributions.

% During the token decoding phase, unlike previous methods such as VCD~\citep{leng2024mitigating}, which contrast outputs from the original input and a distorted input, we aim to adaptively utilize the synthesized visual reference to confirm or revise the LVLM's token predictions.
%
Specifically, to predict a specific token $y_t$, we utilize LVLMs to generate two output distributions, each conditioned on either the original image $v$ or the synthesized visual reference $v'$, expressed as:
\begin{equation}
    p_{\theta}(y_t | v, \mathbf{x}, \mathbf{y}_{<t})\!=\!\mathtt{Softmax}\!\left[f_\theta(y_{t}| v,\mathbf{x},\mathbf{y}_{<t})\right]\!, \,\,\, p_{\theta}(y_t | v', \mathbf{x}, \mathbf{y}_{<t})\!=\!\mathtt{Softmax}\!\left[f_\theta(y_{t}| v',\mathbf{x},\mathbf{y}_{<t})\right].
\end{equation}
We define and compute the following distance metric based on Jensen-Shannon (JS) divergence at each timestep $t$ to quantify the discrepancy between two next-token probability distributions:
\begin{gather} 
    d_t(v, v') = \mathcal{D}_{\mathrm{JS}} \left(p_{\theta}\left(y_t | v, \mathbf{x},\mathbf{y}_{<t}\right) \parallel p_{\theta}\left(y_t | v', \mathbf{x},\mathbf{y}_{<t}\right) \right), \nonumber\\
\text{where} \,\, \mathcal{D}_{\mathrm{JS}}(P \parallel Q) = \frac{1}{2} \mathcal{D}_{\mathrm{KL}}(P \parallel M) + \frac{1}{2} \mathcal{D}_{\mathrm{KL}}(Q \parallel M), \,\text{and} \,\, M=\frac{1}{2}(P+Q).
\end{gather}
Here, $\mathcal{D}_{\mathrm{KL}}$ represents the Kullback-Leibler (KL) divergence. Note that $d_t(v, v') \in [0, 1]$ is a symmetric metric, providing a fine-grained measure of how closely the two distributions align as the model predicts each subsequent token. 

We consider two scenarios based on the token-level generative feedback: (1) If the two predictions are aligned and both images agree on a specific token prediction, we \textit{confirm} the original prediction as correct, and the auxiliary prediction from the generated image can be combined with the original prediction for enhancement (complementary decoding~\citep{woo2024ritual}). (2) Conversely, if there is a significant discrepancy between the predictions, indicating that the original prediction is likely hallucinatory, we \textit{revise} the original response by using the generated visual input as a contrasting reference to refine the initial next-token prediction (contrastive decoding \citep{leng2024mitigating}). To implement this, we introduce a distance threshold $\gamma$ and develop two corresponding decoding approaches as follows:
\begin{equation}
y_t \sim p_{\theta}(y_t) = 
\begin{cases} 
\mathtt{Softmax}\left[f_\theta(y_{t}| v,\mathbf{x},\mathbf{y}_{<t}) + \alpha_{1} \, f_\theta(y_{t}| v',\mathbf{x},\mathbf{y}_{<t})\right], & \text{if } d_t(v, v') < \gamma; \\[5pt]
\mathtt{Softmax}\left[(1 + \alpha_{2}) \, f_\theta(y_{t}| v,\mathbf{x},\mathbf{y}_{<t}) - \alpha_{2} \, f_\theta(y_{t}| v',\mathbf{x},\mathbf{y}_{<t})\right], & \text{if } d_t(v, v') \geq \gamma,
\end{cases}
\end{equation}
where $\alpha_1$ and $\alpha_2$ are hyperparameters that control the influence of the generated visual reference in the final prediction. Note that setting $\alpha_1=0$ or $\alpha_2=0$ degrades this process to regular decoding. The final generated token $y_t$ is sampled from the multinomial distribution with probabilities $p_{\theta}(y_t)$.

% Then we calculate the Jensen-Shannon divergence (JS-divergence)~\citep{menendez1997jensen} of the two probability distributions. The JS-divergence $\mathcal{D}_{\mathrm{JS}}(p_{ori} \parallel p_{syn})$ is calculated as:
% \begin{gather}
% M = \frac{1}{2}(p_{ori} + p_{syn}),\\
% \mathcal{D}_{\mathrm{JS}}(p_{ori} \parallel p_{syn}) = \frac{1}{2} \mathcal{D}_{\mathrm{KL}}(p_{ori} \parallel M) + \frac{1}{2} \mathcal{D}_{\mathrm{KL}}(p_{syn} \parallel M),
% \label{eq:js_divergence}
% \end{gather}
% where $M$ is the mixture distribution of the two predictions and $\mathcal{D}_{\mathrm{KL}}$ stands for the KL-divergence of two distributions.

% The JS-divergence measures the distance between two distributions, making it suitable for adaptive intensity token decoding. Specifically, if the distance is below a certain threshold, we confirm that the two distributions are similar, and the visual reference should be used to validate the original prediction. Conversely, if a significant discrepancy between the distributions is detected, the visual reference is used to revise the original prediction. This process can be mathematically formulated as:

% \clearpage

\section{Experiments}
\label{sec:experiment}
In this section, we evaluate the effectiveness of our method in mitigating hallucinations in LVLMs across a range of benchmarking scenarios, comparing it with existing state-of-the-art approaches. 

\looseness=-1
\textbf{Evaluated LVLMs}. \RebuttalRevision{We evaluate the effectiveness of our method on three state-of-the-art open-source LVLMs: LLaVA-1.5~\citep{liu2024improved}, InstructBLIP~\citep{dai2024instructblip}, and Qwen-VL~\citep{bai2023qwen}}. Both LLaVA-1.5 and InstructBLIP utilize Vicuna-7B~\citep{chiang2023vicuna} as the language encoder, which is instruction-tuned from LLaMA~\citep{touvron2023llama}. In contrast, Qwen-VL~\citep{bai2023qwen} is based on the Qwen 7B backbone. Specifically, we implement our approach using weights of the Qwen-VL-Chat model.

% LLaVA-1.5~\citep{liu2023visual} employs a pre-trained CLIP ViT-L/14~\citep{radford2021learning} as the vision encoder, and trains a linear mapping layer to connect the vision and language modalities. In contrast, InstructBLIP~\citep{dai2024instructblip} builds on a pre-trained BLIP-2~\citep{li2023blip} and incorporates an instruction-aware Q-Former module to bridge the modalities.

\textbf{Benchmarks}. We conduct extensive experiments on six benchmarks: (1) \textbf{POPE~\citep{li2023evaluating}} is a widely used benchmark for assessing object hallucinations in LVLMs, which tests the models with yes-or-no questions regarding the presence of specific objects, such as, ``\texttt{Is there a \{object\} in the image?}'' (2) \textbf{CHAIR~\citep{rohrbach2018object}} evaluates object hallucinations in open-ended captioning tasks. It prompts the LVLMs to describe specific images selected from a random sample of 500 images from the MSCOCO validation set; (3) \textbf{MME-Hallucination~\citep{fu2023mme}} is a comprehensive benchmark for LVLMs consisting of four subsets: \textit{existence} and \textit{count} for object-level hallucinations, and \textit{position} and \textit{color} for attribute-level hallucinations; (4) \RebuttalRevision{\textbf{MMBench~\citep{liu2025mmbench}} serves as a comprehensive benchmark designed to assess the multi-modal understanding capabilities of LVLMs across 20 dimensions}; (5) \textbf{MMVP~\citep{tong2024eyes}} collects CLIP-blind pairs and evaluates the fine-grained visual recognition capabilities of LVLMs. It consists of 150 image pairs, each accompanied by a binary-option question; (6) \textbf{LLaVA-Bench} provides 24 images featuring complex scenes, memes, paintings, and sketches, along with 60 challenging questions.

\begin{table}[t]
    \renewcommand{\arraystretch}{1.05}
    \centering
    \small
    \caption{
        \RebuttalRevision{\textbf{Results on POPE~\citep{li2023evaluating} benchmark}. Higher ($\uparrow$) accuracy, precision, recall, and F1 indicate better performance. The best results are \textbf{bolded}, and the second-best are \underline{underlined}.}
    }
    % \vspace{2pt}
    \label{tab:POPE}
    \setlength{\tabcolsep}{7pt} % base value: 6pt
    \resizebox{\textwidth}{!}{
    \begin{tabular}{cclccccccccc}
    \toprule
     & \multirow{2}{*}[-2pt]{\textbf{Setup}} & \multirow{2}{*}[-2pt]{\textbf{Method}} & \multicolumn{3}{c}{\textbf{LLaVA-1.5}} & \multicolumn{3}{c}{\textbf{InstructBLIP}} & \multicolumn{3}{c}{\RebuttalRevision{\textbf{Qwen-VL}}} \\
    \arrayrulecolor{gray} \cmidrule(lr){4-6} \cmidrule(lr){7-9} \cmidrule(lr){10-12}
     &  &  & {Acc.} $\uparrow$ & {Prec.}  $\uparrow$ & {F1} $\uparrow$ & {Acc.} $\uparrow$ & {Prec.} $\uparrow$ & {F1} $\uparrow$ & \RebuttalRevision{{Acc.} $\uparrow$} & \RebuttalRevision{{Prec.} $\uparrow$} & \RebuttalRevision{{F1} $\uparrow$} \\
    \midrule
    \multirow{15}{*}[-5pt]{\rotatebox{90}{\textbf{\normalsize MS-COCO}}} & \multirow{5}{*}{Random} 
    & Regular & 83.13 & 81.94  & 83.44 & 83.07 & 83.02  & 83.08 & 87.43 & 93.56 & 86.48 \\
     &  & VCD  & 87.00 & 86.13  & 87.15 & 86.23 & 88.14  & 85.88 & 88.80 & 93.89 & 88.11 \\
     &  & M3ID  &  87.50 &  87.38  &  87.52 &  86.67 &  88.09  &  86.41 & \textbf{89.83} & \underline{95.44} & \underline{89.17}  \\
     &  & RITUAL  &  \underline{88.87}	&  \underline{89.23} &   \textbf{88.81} &  \textbf{88.83} &  \underline{90.48}  &  \textbf{88.60} & 89.47 & \textbf{96.32} & 88.62 \\
     &  & \cc \textbf{Ours} &\cc \textbf{89.03} &\cc \textbf{91.20}  &\cc \underline{88.74}  &\cc \textbf{88.83}  &\cc \textbf{93.73}  &\cc \underline{87.71}  & \cc \underline{89.73} & \cc 93.19 & \cc \textbf{89.31} \\
     \arrayrulecolor{gray}\cmidrule(lr){2-12}
      &  \multirow{5}{*}{Popular} & Regular & 81.17 &	78.28  &	82.08 & 77.00	& 73.82  &	78.44 & 84.70 & 88.24 & 83.96 \\
     &  & VCD  & 83.10 & 79.96  & 83.94 &  80.07 &  77.67 & 80.89 & 85.13 & 87.27 & 84.69 \\
     &  & M3ID  &  84.30 &  81.58  &  84.95 & 80.97 & 77.93  &  81.85 & \underline{86.27} & \underline{89.19} & \textbf{85.73} \\
     &  & RITUAL  &  \underline{85.83} &  \underline{84.17}  &  \underline{86.17} &  \underline{81.97} &  \underline{78.90}  &  \textbf{82.87}  & 84.57 & 84.09 & 84.67 \\
     &  & \cc \textbf{Ours} &\cc \textbf{86.63} &\cc \textbf{87.75}  &\cc \textbf{86.28} &\cc \textbf{82.73} &\cc \textbf{84.02} &\cc \underline{82.10} & \cc \textbf{86.50} & \cc \textbf{89.87} & \cc \underline{85.71} \\
     \arrayrulecolor{gray}\cmidrule(lr){2-12}
      &  \multirow{5}{*}{Adversarial} & Regular & 77.43 & 73.31  & 79.26 & 74.60 & 71.26  & 76.45 & 79.83 & 80.13 & 79.73 \\
     &  & VCD  & 77.17 & 72.18  & 79.47 &  77.20 &  74.29  &  78.49 & 81.33 & 80.60 & 81.55 \\
     &  & M3ID  &  78.23 &  73.51  &  80.22 & 77.47 & 73.68  & 79.14 & 82.03 & 81.47 & 82.19 \\
     &  & RITUAL  &  \underline{78.80} &  \underline{74.43}  &  \underline{80.54} &  \underline{78.73} &  \underline{74.57}  &  \textbf{80.39}  & \underline{82.80} & \underline{83.15} & \underline{82.71} \\
     &  & \cc \textbf{Ours} &\cc \textbf{81.63} &\cc \textbf{80.59}   &\cc \textbf{81.94} &\cc \textbf{80.30} &\cc \textbf{80.90}  &\cc \underline{80.11} &\cc \textbf{83.47} & \cc \textbf{84.49} & \cc \textbf{82.98} \\
     \arrayrulecolor{gray}\midrule
     \multirow{15}{*}[-5pt]{\rotatebox{90}{\textbf{\normalsize A-OKVQA}}} &  \multirow{5}{*}{Random} & Regular & 81.90 & 76.63  & 83.53 & 80.63 & 76.82  & 81.92 & 86.27 & 90.66 & 85.48 \\
     &  & VCD  & 83.83 & 78.05  & 85.34 & 84.20 &  80.90  & 85.00 & 87.87 & 90.06 & 87.53 \\
     &  & M3ID  &  84.67 &  79.25  &  85.97 &  85.43 & 81.77  &  86.23 & \textbf{88.13} & \underline{92.06} & \underline{87.55} \\
     &  & RITUAL  &  \underline{85.17} &  \underline{79.79}  &  \underline{86.40} &  \underline{87.13} & \underline{83.92} &  \textbf{87.71} & 87.73 & \textbf{92.49} & 87.01 \\
     &  & \cc \textbf{Ours} &\cc \textbf{86.93} &\cc \textbf{84.28}  &\cc \textbf{87.42} &\cc \textbf{87.40} &\cc \textbf{87.67}  &\cc \underline{87.26} & \cc \underline{87.90} & \cc 89.16 & \cc \textbf{87.58} \\
     \arrayrulecolor{gray}\cmidrule(lr){2-12}
      &  \multirow{5}{*}{Popular} & Regular & 75.07 &  68.58 & 78.77 & 75.17 & 70.15  & 77.91 & 84.60 & 87.99 & 83.88 \\
     &  & VCD  & 76.63 & 69.59 & 80.19 &  78.63 &  \underline{73.53} &  80.72 & 86.23 & 87.30 & 86.03 \\
     &  & M3ID  &  77.80 & 70.98  &  80.91 & \underline{78.80} & 73.38 & 81.00 & \textbf{86.50} & \underline{89.59} & 85.95 \\
     &  & RITUAL  &  \underline{78.83} &  \underline{71.99} &  \underline{81.68} &  {78.73} &  {72.83} &  \underline{81.17} & 86.36 & 88.73 & \underline{86.20} \\
     &  & \cc \textbf{Ours} &\cc \textbf{80.90} &\cc \textbf{75.68} &\cc \textbf{82.66}  &\cc \textbf{81.47} &\cc \textbf{78.61}  &\cc \textbf{82.35} & \cc \underline{86.43} & \cc \textbf{90.74} & \cc \textbf{86.52} \\
     \arrayrulecolor{gray}\cmidrule(lr){2-12}
      &  \multirow{5}{*}{Adversarial} & Regular & 67.23 & 61.56 & 73.70 & 69.87 & 64.54 & 74.54 & 76.90 & 75.59 & 77.48 \\
     &  & VCD  & 67.40 & 61.39  & 74.21 &  \underline{71.00} &  \underline{65.41}  &  75.45 & 79.13 & 76.04 & 80.30 \\
     &  & M3ID  &  \underline{68.60} &  62.22  &  \underline{75.11} & 70.10 &  64.28  & 75.16 & 79.50 & 77.54 & 80.21 \\
     &  & RITUAL  &  {68.57} &  \underline{62.26}  &  {74.99} &  {70.27} & 64.15 & \underline{75.55} & \underline{80.20} & \underline{79.08} & \textbf{80.58} \\
     &  & \cc \textbf{Ours} &\cc \textbf{72.70} &\cc \textbf{66.70}  &\cc \textbf{76.86} &\cc \textbf{73.93} &\cc \textbf{69.36}   &\cc \textbf{76.67} & \cc \textbf{80.75} & \cc \textbf{80.37} & \cc \underline{80.46} \\
     \arrayrulecolor{gray}\midrule
     \multirow{15}{*}[-5pt]{\rotatebox{90}{\textbf{\normalsize GQA}}} &  \multirow{5}{*}{Random} & Regular & 82.23 & 76.32  & 84.03 & 79.67 & 76.05  & 80.99 & 84.90 & 89.51 & 83.96 \\
     &  & VCD  & 83.23 & 76.73  & 85.05 &  82.83 &  80.16  &  83.56 & 85.21 & 92.05 & 84.21 \\
     &  & M3ID  &  84.20 &  78.00  &  85.77 & 83.07 & 80.06 & 83.87 & 85.69 & 93.11 & 84.67 \\
     &  & RITUAL  &  \underline{86.10} &  \underline{80.30} &  \underline{87.31} &  \underline{84.87} &  \underline{82.52}  &  \textbf{85.39} & \textbf{86.13} & \underline{93.78} & \underline{84.81} \\
     &  & \cc \textbf{Ours} &\cc \textbf{87.40} &\cc \textbf{83.51}   &\cc \textbf{88.09} &\cc \textbf{85.40} &\cc \textbf{85.64}  &\cc \underline{85.12} & \cc \underline{85.95} & \cc \textbf{94.22} & \cc \textbf{85.08} \\
     \arrayrulecolor{gray}\cmidrule(lr){2-12}
      &  \multirow{5}{*}{Popular} & Regular & 73.47 & 66.83  & 77.84 & 73.33 & 68.72  & 76.26 & 81.33 & 83.38 & 80.74 \\
     &  & VCD  & 72.37 & 65.27 & 77.58 & \underline{76.13} & \underline{71.10}  & \textbf{78.68} & 81.97 & 82.82 & 81.73 \\
     &  & M3ID  &  73.87 &  66.70  &  78.49 &  75.17 &  69.94  &  78.04 & \textbf{82.13} & 84.58 & \underline{81.48} \\
     &  & RITUAL  &  \underline{74.80} &  \underline{67.50}  &  \underline{79.15} &  {74.50} &  {69.17}  &  {77.61} & 81.13 & \underline{85.48} & 81.03 \\
     &  & \cc \textbf{Ours} &\cc \textbf{78.10} &\cc \textbf{71.56}   &\cc \textbf{80.98}  &\cc \textbf{76.90} &\cc \textbf{73.89} &\cc \underline{78.27} & \cc \underline{82.10} & \cc \textbf{86.39} & \cc \textbf{81.85} \\
     \arrayrulecolor{gray}\cmidrule(lr){2-12}
      &  \multirow{5}{*}{Adversarial} & Regular & 68.60 & \underline{62.43}  & 74.84 &  68.60 &  63.94  &  73.10 & 79.03 & 80.43 & 78.54 \\
     &  & VCD  & \underline{68.83} & 62.26  & \underline{75.43} & 71.00 & 65.75  & \underline{75.14} & 80.87 & 81.07 & 80.80 \\
     &  & M3ID  &  68.67 &  62.16  &  75.28 & \underline{71.17} & \underline{65.79}  & \textbf{75.36} & 81.03 & 82.93 & \textbf{80.94} \\
     &  & RITUAL  &  {68.23} &  {61.75} &  {75.10} &  {70.17} &  {64.76}  &  {74.78} & \underline{81.07} & \underline{83.29} & 80.41 \\
     &  & \cc \textbf{Ours} &\cc \textbf{74.07} &\cc \textbf{67.42}  &\cc \textbf{78.22} &\cc \textbf{73.63} &\cc \textbf{70.08}  &\cc 75.11 & \cc \textbf{81.13} & \cc \textbf{84.18} & \cc \underline{80.57} \\
    \bottomrule
    \end{tabular}
    }
    \vspace{-5pt}
\end{table}

\textbf{Baselines}. \RebuttalRevision{As a simple baseline, we include results from regular decoding, where the next token is sampled directly from the post-softmax probability distribution.} Additionally, we compare the performance of our method with three state-of-the-art decoding approaches: VCD~\citep{leng2024mitigating}, M3ID~\citep{favero2024multi}, and RITUAL~\citep{woo2024ritual}. \RebuttalRevision{For evaluations on the CHAIR~\citep{rohrbach2018object} and MME-Hallucination~\citep{fu2023mme} benchmark, we further include comparisons with Woodpecker~\citep{yin2023woodpecker}, HALC~\citep{chen2024halc}, DoLa~\citep{chuang2024dola} and OPERA~\citep{huang2024opera}.} We report the performance of these baselines based on our re-implementation using their released code bases.

\looseness=-1
\textbf{Implementation Details}. In our experiments, we adhere to the default query format for the input data used in both LLaVA-1.5~\citep{liu2024improved} and InstructBLIP~\citep{dai2024instructblip}. Additionally, we set $\alpha_1 = 3$, $\alpha_2 = 1$, and $\gamma = 0.1$ by default in our decoding process. We follow VCD~\citep{leng2024mitigating} to implement adaptive plausibility constraints~\citep{li2023contrastive}, where we set $\beta=0.1$ in open-ended CHAIR benchmark and $\beta=0.25$ for other tasks.
To ensure the reliability of our results, we conduct MME experiments three times with different initialization seeds and report the mean accuracy along with the standard deviation. All experiments are conducted on a single 48GB NVIDIA RTX 6000 Ada GPU. More implementation details are provided in Section~\ref{sec:detail} of the Appendix.

\subsection{Results and Discussions}
\looseness=-1
\textbf{Results on POPE}. In Table~\ref{tab:POPE}, we compare the performance of our method against other baselines on the POPE benchmark under three different negative sampling settings, across three datasets. 
As shown in the table, our method consistently outperforms other decoding methods on three  LVLMs, achieving state-of-the-art accuracies across all settings, with improvements of up to 5.24\% in accuracy, 6.33\% in precision, and 2.79\% in F1 score compared to the second-best approach. This suggests that incorporating a generative reference enables the LVLMs to perceive more fine-grained visual details, thereby effectively addressing object hallucinations. Moreover, while most decoding methods tend to be overconfident in their responses, the self-correcting decoding mechanism in our method makes it more conservative in responding \texttt{Yes}, as evidenced by significantly higher precision across all settings. This highlights its enhanced performance in filtering out false positives and suppressing misinformation.

Another notable finding is that our method shows significantly improved performance in the \textit{popular} and \textit{adversarial} settings, which are more challenging than the \textit{random} setting. In the \textit{popular} and \textit{adversarial} settings, non-existent negative objects frequently appear and co-occur with other objects~\citep{li2023evaluating}, making them more susceptible to hallucination by LVLMs, as evidenced by the varying degrees of performance degradation across all baselines. However, our method exhibits a lower performance drop compared to other baselines, demonstrating its effectiveness in addressing hallucinations arising from object co-occurrence.

\begin{table}[t]
    % \begin{minipage}[t!]{0.5\textwidth}
        \begin{center}
        \begin{small}
        
        \setlength{\tabcolsep}{5pt} % base value: 6pt
        \caption{\textbf{Results on CHAIR~\citep{rohrbach2018object} benchmark.} We limit the maximum number of new tokens to 64. Lower ($\downarrow$) CHAIR$_S$, CHAIR$_I$ and higher ($\uparrow$) recall and length indicate better performance. The best results in each setting are \textbf{bolded}, and the second-best are \underline{underlined}.}
        \vspace{-6pt}
        \label{tab:CHAIR}
        \resizebox{\textwidth}{!}{
        \begin{tabular}{lcccccccc}
            \toprule
              \multirow{2}{*}[-0.5ex]{\textbf{Method}}  &  \multicolumn{4}{c}{\textbf{LLaVA-1.5}} & \multicolumn{4}{c}{\textbf{InstructBLIP}} \\
             \cmidrule(lr){2-5}\cmidrule(lr){6-9}
                 & CHAIR$_S$ $\downarrow$ & CHAIR$_I$ $\downarrow$ & Recall $\uparrow$ & Length $\uparrow$   & CHAIR$_S$ $\downarrow$ & CHAIR$_I$ $\downarrow$ & Recall $\uparrow$ & Length $\uparrow$ \\
             \midrule
             Regular & 26.2 & 9.4 & 58.5 & 53.4 &  31.2 & 11.1  & 59.0 & 53.6\\
              VCD & 24.4 & 7.9 & 63.3 & \underline{54.2} &  {30.0} & {10.1}  & 61.8 & 54.2\\
              M3ID & \underline{21.4} & \underline{6.3}  & \textbf{64.4} & 53.5 & 30.8 & 10.4 & 62.6 & 53.4\\
              RITUAL & {22.4} & {6.9}  & 63.0 & \textbf{54.9} &  26.6 & 8.9 & 63.4 & \underline{55.3}\\
              \RebuttalRevision{Woodpecker} & {24.9} & {7.5}  & 60.8 & 49.7 &  31.2 & 10.8 & 62.3 & 51.3\\
              \RebuttalRevision{HALC} & {21.7} & {7.1}  & \underline{63.4} & 53.4 &  \underline{24.5} & \underline{8.0} & \underline{63.8} & 55.1\\
              \cc \textbf{Ours} & \cc \textbf{18.4} & \cc \textbf{6.1}  & \cc {62.7} & \cc {54.1} & \cc  \textbf{24.0} & \cc \textbf{7.7}  & \cc \textbf{67.2}  & \cc \textbf{55.5} \\
            \bottomrule
        \end{tabular}
        
        }
        \end{small}
        \end{center}
        \vspace{-10pt}
\end{table}
\begin{table*}[t!]
    % \begin{minipage}[t!]{0.5\textwidth}
        \begin{center}
        \begin{small}
        \setlength{\tabcolsep}{6pt} % base value: 6pt
        \caption{\looseness=-1\textbf{Results on MME-Hallucination~\citep{fu2023mme} and MMBench~\citep{liu2025mmbench} benchmark.} We report the average MME scores along with the standard deviation across three random seeds for each subset. \RebuttalRevision{We also report the overall accuracy achieved by the different methods on the MMBench benchmark in the final column}. Higher scores ($\uparrow$) indicate better performance. The best results  are \textbf{bolded}, and the second-best are \underline{underlined}. }
        % \martin{consistent with confidence intervals; either include error bars in all tables, or include none in main text and add in Appendix?}}
        \label{tab:MME}
        \vspace{-5pt}
        \resizebox{\textwidth}{!}{
        \begin{tabular}{lcccccc}
            \toprule
             \multirow{2}{*}[-0.5ex]{\textbf{Method}}  &  \multicolumn{2}{c}{\textbf{Object-level}} & \multicolumn{2}{c}{\textbf{Attribute-level}} & \multirow{2}{*}[-0.5ex]{\textbf{MME Score $\uparrow$}} & \multirow{2}{*}[-0.5ex] {\RebuttalRevision{\textbf{MMBench $\uparrow$}}}\\
             \cmidrule(lr){2-3}\cmidrule(lr){4-5}
                & Existence $\uparrow$ & Count $\uparrow$ & Position $\uparrow$ & Color $\uparrow$   & \\
             \midrule
             Regular &  173.75 {\tiny ($\pm4.79$)} & 121.67 {\tiny ($\pm12.47$)} & 117.92 {\tiny ($\pm3.69$)\phantom{0}} & 149.17 {\tiny ($\pm7.51$)\phantom{0}} &  562.50 {\tiny ($\pm3.96$)\phantom{0}} & \RebuttalRevision{64.1} \\
              DoLa  & 176.67 {\tiny ($\pm2.89$)} & 113.33 {\tiny ($\pm10.41$)} & 90.55 {\tiny ($\pm8.22$)} & 141.67 {\tiny ($\pm7.64$)\phantom{0}} &  522.22 {\tiny ($\pm16.78$)} & \RebuttalRevision{63.8} \\
              OPERA  & 183.33 {\tiny ($\pm6.45$)} & 137.22 {\tiny ($\pm6.31$)\phantom{0}} & 122.78 {\tiny ($\pm2.55$)\phantom{0}} &155.00 {\tiny ($\pm5.00$)\phantom{0}}  &  598.33 {\tiny ($\pm10.41$)} & \RebuttalRevision{64.4} \\
              VCD  & 186.67 {\tiny ($\pm5.77$)} & 125.56 {\tiny ($\pm3.47$)\phantom{0}} & 128.89 {\tiny ($\pm6.73$)\phantom{0}} & 139.45 {\tiny ($\pm12.51$)} &  580.56 {\tiny ($\pm15.13$)} & \RebuttalRevision{\underline{64.6}} \\
              M3ID  & 186.67 {\tiny ($\pm5.77$)} &  128.33 {\tiny ($\pm10.41$)} & \underline{131.67} {\tiny ($\pm5.00$)\phantom{0}} & 151.67 {\tiny ($\pm20.88$)} &  598.11 {\tiny ($\pm20.35$)}  & \RebuttalRevision{64.4} \\
             RITUAL  & \underline{187.50} {\tiny ($\pm2.89$)} & \underline{139.58} {\tiny ($\pm7.64$)\phantom{0}}  & 125.00 {\tiny ($\pm10.27$)} & \underline{164.17} {\tiny ($\pm6.87$)\phantom{0}} & \underline{616.25} {\tiny ($\pm20.38$)} & \RebuttalRevision{63.8}    \\
            \RebuttalRevision{Woodpecker} & \RebuttalRevision{\underline{187.50} {\tiny ($\pm2.89$)}} & \RebuttalRevision{125.00 {\tiny ($\pm0.00$)\phantom{0}}} & \RebuttalRevision{126.66 {\tiny ($\pm2.89$)\phantom{0}}} & \RebuttalRevision{149.17 {\tiny ($\pm17.34$)}} &  \RebuttalRevision{588.33 {\tiny ($\pm10.00$)}} & \RebuttalRevision{64.0}\\
            \RebuttalRevision{HALC} & \RebuttalRevision{183.33 {\tiny ($\pm0.00$)}} & \RebuttalRevision{133.33 {\tiny ($\pm5.77$)\phantom{0}}} & \RebuttalRevision{107.92 {\tiny ($\pm3.69$)\phantom{0}}} &\RebuttalRevision{155.00 {\tiny ($\pm5.00$)\phantom{0}}}  &  \RebuttalRevision{579.58 {\tiny ($\pm9.07$)\phantom{0}}} & \RebuttalRevision{64.2}\\
             
             \cc \textbf{Ours} & \cc \textbf{188.33} {\tiny ($\pm2.89$)} & \cc \textbf{150.00} {\tiny ($\pm7.64$)\phantom{0}}  & \cc \textbf{133.89} {\tiny ($\pm3.85$)\phantom{0}} & \cc \textbf{172.22} {\tiny ($\pm3.47$)\phantom{0}} & \cc \textbf{644.44} {\tiny ($\pm9.18$)\phantom{0}} & \cc \RebuttalRevision{\textbf{65.5}} \\
            %  \arrayrulecolor{gray!50}\cmidrule(lr){2-4}
            %  & \textbf{\Ours}+VCD & 20.0 & 6.8 \\
            %  & \textbf{\Ours}+M3ID & 18.0 & 5.7 \\
            \bottomrule
        \end{tabular}
        }
        \vspace{-15pt}
        \end{small}
        \end{center}
        
\end{table*}

\textbf{Results on CHAIR}. We also compare the performance of our methods and other state-of-the-art methods in the open-ended captioning task and report the CHAIR scores, recall, and the average length of responses in Table~\ref{tab:CHAIR}, \RebuttalRevision{Table~\ref{tab:CHAIR-128}, and Table~\ref{tab:CHAIR-256}}. The results, evaluated across two different LVLMs, consistently demonstrate performance improvements achieved by our method over the compared approaches. Specifically, our method outperforms the second-best approach by 3.0\% and 2.6\% on the CHAIR$_S$ metric, while also enhancing the detailedness of generated responses compared to regular decoding, as indicated by the higher recall and increased response length.
These results demonstrate that by incorporating generative feedback into the decoding process of LVLMs, our method effectively mitigates object hallucinations in open-ended captioning tasks.

% \clearpage
\begin{figure}[t]
\begin{minipage}[t]{0.49\linewidth}
\includegraphics[width=\linewidth]{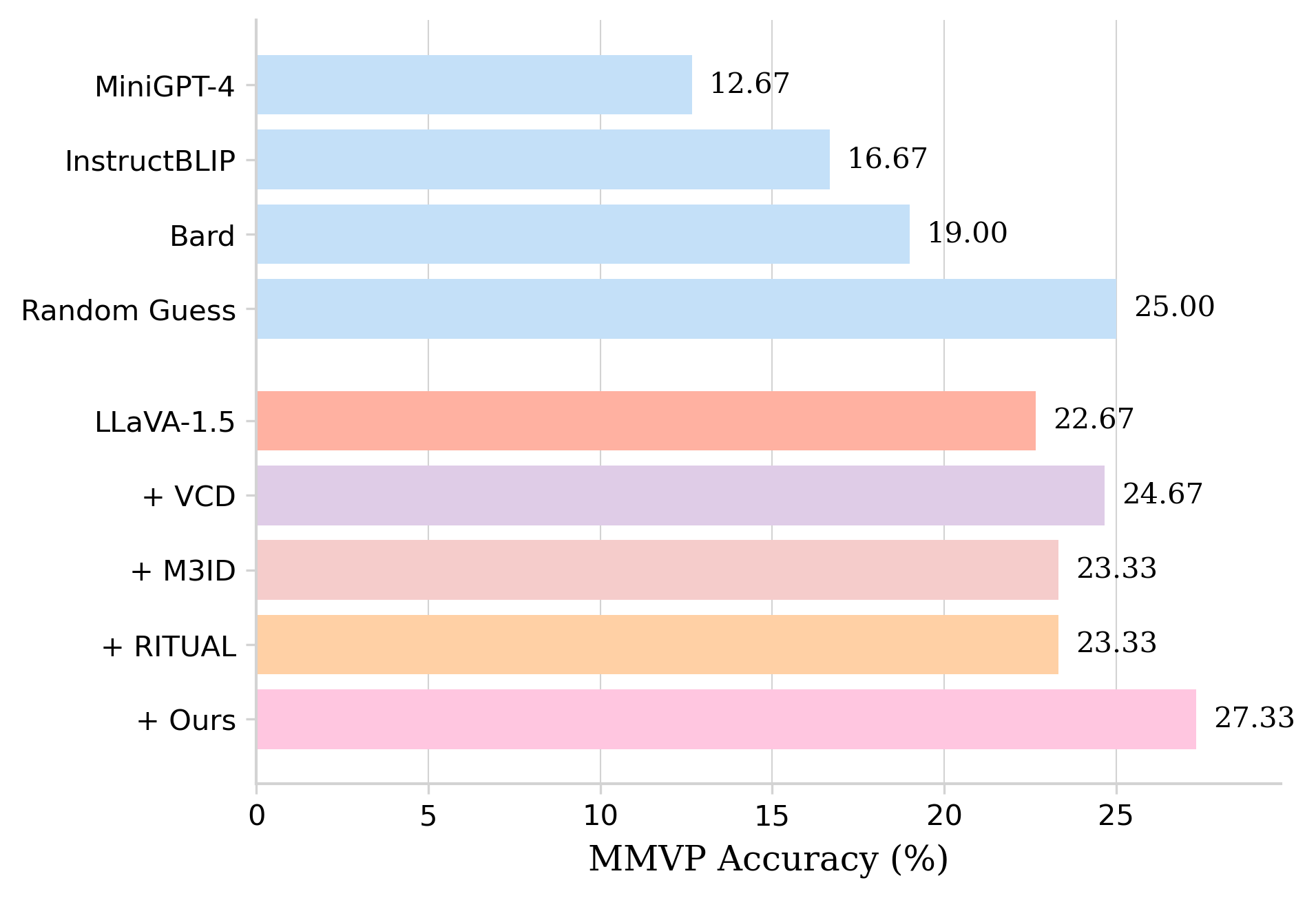}
\vspace{-21pt}
\caption{\looseness=-1\textbf{Results on MMVP~\citep{tong2024eyes}}. We apply our approach to LLaVA-1.5~\citep{liu2024improved} and compare its performance against other hallucination mitigation methods.}
\label{fig:mmvp}
\end{minipage}
\hspace{2pt}
\begin{minipage}[t]{0.49\linewidth}
\makeatletter\def\@captype{table}
\vspace{-128pt}
\small
\begin{center}
\resizebox{\linewidth}{!}{
\renewcommand{\arraystretch}{1.05}
\begin{tabular}{lcccc}
    \toprule
      \multirow{2}{*}[-0.5ex]{\textbf{Method}}  &  \multicolumn{2}{c}{\textbf{LLaVA-1.5}} & \multicolumn{2}{c}{\textbf{InstructBLIP}} \\
     \cmidrule(lr){2-3}\cmidrule(lr){4-5}
         & Acc. $\uparrow$ & Det. $\uparrow$ & Acc. $\uparrow$ & Det. $\uparrow$ \\
     \midrule
     Regular & 2.88 & 3.29 & 3.42 & 3.96 \\
     \cc \textbf{Ours} & \cc \textbf{4.29} & \cc \textbf{4.54}  & \cc \textbf{4.38} & \cc \textbf{4.79} \\
     \midrule
      VCD & 3.62 & 3.83  & 3.71 & 4.21\\
      \cc \textbf{Ours} & \cc \textbf{4.04} & \cc \textbf{4.38}  & \cc \textbf{4.17} & \cc \textbf{4.58} \\
      \midrule
      M3ID & 3.88 & 4.08 & 4.00 & 4.33 \\
      \cc \textbf{Ours} & \cc \textbf{4.04} & \cc \textbf{4.29}  & \cc \textbf{4.08} & \cc \textbf{4.50} \\
    \bottomrule
\end{tabular}
}
\vspace{2pt}
\caption{\RebuttalRevision{\textbf{GPT-4V-aided evaluation on LLaVA-Bench}. Higher accuracy and detailedness ($\uparrow$) indicate better performance. The evaluation is performed on LLaVA-1.5~\citep{liu2024improved}.}}
\label{table:gpt4v}
\end{center}
\end{minipage}
\vspace{-3pt}
\end{figure}

\textbf{Results on MME-Hallucination and MMBench}. Beyond object hallucinations, we further compare the performance of our method with other approaches using the more comprehensive MME-Hallucination benchmark, which includes both object-level and attribute-level hallucinations. The results in Table~\ref{tab:MME} and Table~\ref{tab:MME-full}
demonstrate that our method significantly outperforms the compared methods, with substantial margins in the total score metric (\eg, +18.19 on LLaVA-1.5 and +21.11 on InstructBLIP) and consistently superior performance across various evaluation settings, achieving the best results in 6 out of 8 settings. Moreover, our method shows notable improvements on the attribute-level \textit{color} subset,  which is particularly challenging as it requires models to accurately capture subtle attribute information. This further illustrates the effectiveness of our approach in addressing a wide range of hallucinations, both at the object existence level and in finer-grained attribute recognition. \RebuttalRevision{Additionally, our proposed DeGF enhances the general multi-modal understanding capabilities of LVLMs, as evidenced by its superior performance on the MMBench benchmark.}

% \begin{wrapfigure}[18]{r}{0.5\textwidth}
% \centering
% \vspace{-15pt}
% \includegraphics[width=\linewidth]{figs/mmvp.png}
% \vspace{-18pt}
% \caption{\textbf{Results on MMVP~\citep{tong2024eyes} benchmark}. We apply our approach to LLaVA-1.5~\citep{liu2023visual} and compare its performance against other hallucination mitigation methods (VCD, M3ID, and RITUAL). For reference, we also report the performance of other LVLMs.}
% \label{fig:mmvp}
% \end{wrapfigure}
\textbf{Results on MMVP}. We conduct experiments on the MMVP benchmark to assess the fine-grained visual recognition capabilities of LVLMs. As shown in Figure~\ref{fig:mmvp}, applying our self-correcting decoding approach to LLaVA-1.5 significantly improves performance from 22.67\% to 27.33\%.
Our approach also demonstrates notable advantages over other hallucination mitigation baselines, further showcasing its superiority in handling nuanced visual recognition tasks.
These results suggest that our approach significantly enhances the model’s capacity to discern and correctly interpret fine-grained distinctions between images with similar appearances but different contents. By integrating generative feedback, our approach effectively reduces misinterpretations and improves the precision of visual recognition tasks, contributing to more reliable and accurate performance in complex scenarios.

% \begin{figure}[t]
% \centering
% \includegraphics[width=0.6\linewidth]{figs/mmvp.png}
% \caption{\textbf{Results on MMVP~\citep{tong2024eyes} benchmark}. We apply our approach to LLaVA-1.5~\citep{liu2023visual} and compare its performance against other hallucination mitigation methods. For reference, we also report the performance of other LVLMs: MiniGPT-4~\citep{zhu2023minigpt}, InstructBLIP~\citep{dai2024instructblip}, and Bard.}
% \label{fig:mmvp}
% \end{figure}

\begin{figure}[t]
\centering
\includegraphics[width=\linewidth]{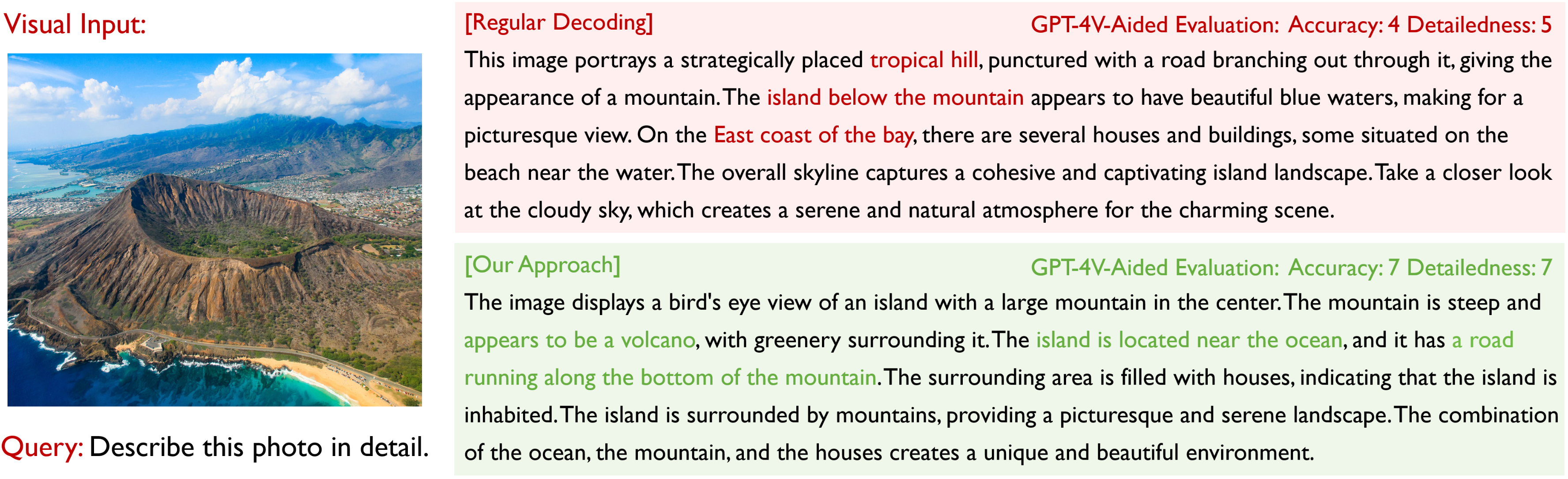}
\vspace{-18pt}
\caption{\textbf{Case study on the LLaVA-Bench benchmark}. We compare the responses generated by regular decoding and our method using LLaVA-1.5. GPT-4V-aided evaluation results are also provided alongside the responses. Hallucinated and accurate content is highlighted in \textcolor{darkred}{red} and \textcolor{darkgreen}{green}.} 
% \martin{Green-code correct parts corresponding to the red parts, \eg, island is located near the ocean?}
\vspace{-12pt}
\label{fig:llavabench}
\end{figure}

% \begin{wraptable}[11]{r}{0.5\textwidth}
% \small
% \begin{center}
% \vspace{-24pt}
% \caption{\textbf{GPT-4V-aided evaluation on LLaVA-Bench}. Higher scores ($\uparrow$) indicate better performance. }
% \vspace{-5pt}
% \label{table:gpt4v}
% \resizebox{\linewidth}{!}{
% \begin{tabular}{lcccc}
%     \toprule
%       \multirow{2}{*}[-0.5ex]{\textbf{Method}}  &  \multicolumn{2}{c}{\textbf{LLaVA-1.5}} & \multicolumn{2}{c}{\textbf{InstructBLIP}} \\
%      \cmidrule(lr){2-3}\cmidrule(lr){4-5}
%          & Acc. $\uparrow$ & Det. $\uparrow$ & Acc. $\uparrow$ & Det. $\uparrow$ \\
%      \midrule
%      Regular & 2.88 & 3.29 & 3.42 & 3.96 \\
%      \cc \textbf{Ours} & \cc \textbf{4.29} & \cc \textbf{4.54}  & \cc \textbf{4.38} & \cc \textbf{4.79} \\
%      \midrule
%       VCD & 24.4 & 7.9  & 61.8 & 54.2\\
%       \cc \textbf{Ours} & \cc \textbf{4.29} & \cc \textbf{4.54}  & \cc \textbf{4.38} & \cc \textbf{4.79} \\
%       \midrule
%       M3ID & 53.5 & 30.8 & 10.4 & 62.6 \\
%       \cc \textbf{Ours} & \cc \textbf{18.4} & \cc \textbf{6.1}  & \cc {62.7} & \cc {62.7} \\
%     \bottomrule
% \end{tabular}
% }
% \end{center}
% \end{wraptable}
\looseness=-1
\textbf{Results on LLaVA-Bench}. In Figure~\ref{fig:llavabench}, we present a case study on LLaVA-Bench comparing our method's response with the response generated by regular decoding using the LLaVA-1.5 model. Specifically, regular decoding often leads to hallucinated or inaccurate content, such as describing ``\texttt{the island below the mountain}''. 
Besides, the response generated by regular decoding tends to focus on elements like the ``\texttt{cloudy sky}'' and ``\texttt{cohesive and captivating island landscape}'' without providing specific information about the central features of the image. In contrast, our response is more detailed, mentioning the volcano, the road, the surrounding greenery, and the inhabited areas, which gives a clearer understanding of the image's content. \RebuttalRevision{The GPT-4V-aided evaluation shown in Table~\ref{table:gpt4v} further confirms that our method enhances both the accuracy and detailedness of the generated response, outperforming other hallucination mitigation approaches such as VCD and M3ID.}
Due to the page limit, please refer to Section~\ref{sec:case} of the Appendix for more case studies.

\begin{figure}[t]
\begin{minipage}[t]{0.49\linewidth}
\makeatletter\def\@captype{table}
\setlength\tabcolsep{3pt}
\caption{\textbf{Sensitivity analysis of distance threshold $\gamma$.} We present the performance of our approach, based on the LLaVA-1.5 backbone, across three benchmarks for varying values of $\gamma$.}
% \vspace{3.3pt}
    \label{tab:gamma}
    \resizebox{\textwidth}{!}{
    \begin{tabular}{lcccc}
        \toprule
        Values of $\gamma$ & POPE Acc. & CHAIR$_S$ & CHAIR$_I$ & MME Score \\
        \midrule
        $\gamma = 0$ & 87.93 & 21.0 & 7.2 & 622.50 \\
        $\gamma = 0.01$ & 88.07 & 21.0 & 6.8 & 632.22 \\
        $\gamma = 0.05$ & 88.67 & 19.4 & 6.3 & 637.50 \\
        \cc $\gamma = 0.1$ & \cc\textbf{89.03} & \cc\textbf{18.4} & \cc\textbf{6.1} & \cc644.44 \\
        $\gamma = 0.5$ & 88.73 & 19.8 & 6.4 & \textbf{646.67} \\
        $\gamma = 1$ & 88.43 & 21.6 & 6.6 & 638.33 \\
        \bottomrule
    \end{tabular}
    }
\end{minipage}
\hspace{2pt}
\begin{minipage}[t]{0.49\linewidth}
\makeatletter\def\@captype{table}
\setlength\tabcolsep{2.5pt}
\caption{\textbf{Effects of different generative models.} We report the performance of different variants of our method, utilizing various stable diffusion models, on the LLaVA-1.5 backbone.}
    \vspace{1pt}
    \label{tab:generative}
    \resizebox{\textwidth}{!}{
    \begin{tabular}{lcccc}
        \toprule
        Models & POPE Acc. & CHAIR$_S$ & CHAIR$_I$ & MME Score \\
        \midrule
        Regular & 83.13 & 26.2 & 9.4 & 562.50  \\
        SD-v1.1 & 88.37 & 19.3 & 6.5 & 638.33 \\
        \cc SD-v1.5 & \cc\textbf{89.03} & \cc18.4 & \cc6.1 & \cc644.44 \\
        SD-v2.1 & 88.70 & 18.8 & 6.7 & 632.22 \\
        SD-XL-v0.9 & 88.87 & 18.6 & 6.1 & 642.50 \\
        SD-XL-v1.0 & 88.60 & \textbf{17.9} & \textbf{5.8} & \textbf{648.33} \\
        \bottomrule
    \end{tabular}
}
\end{minipage}
\vspace{-8pt}
\end{figure}

% \vspace{-5pt}
\subsection{Ablation Studies}
% \vspace{-3pt}
\textbf{Analysis of Distance Threshold $\gamma$}. In Section~\ref{sec:token}, we introduce a distance threshold $\gamma$ to determine the appropriate decoding algorithm for each generated token. Table~\ref{tab:gamma} presents an analysis of our method's performance with various values of $\gamma$ across three benchmarks. For simplicity, we report the performance on the MS-COCO dataset with \textit{random} setting for all POPE results in the ablation studies.
Notably, when $\gamma$ is set to either 0 or 1—corresponding to the exclusive use of contrastive or complementary decoding for all tokens—the performance exhibits a significant decline, by 0.6\% and 1.1\% in POPE accuracy, respectively. Moreover, our default setting of $\gamma=0.1$ achieves the optimal performance in 3 out of 4 evaluated metrics. Additional sensitivity analyses for other hyperparameters are provided in Section~\ref{sec:moreablation} of the Appendix.

\textbf{Effects of Different Generative Models}. Table~\ref{tab:generative} presents the performance of various variants of our method that incorporate different generative models (\ie, different versions of Stable Diffusion) while using the same LLaVA-1.5 backbone. The results indicate that the effectiveness of our DeGF is robust to the choice of generative models, as performance remains largely unaffected by the specific model used, and all variants demonstrate consistent improvements over the original regular decoding approach.
Although utilizing SD-XL-v1.0~\citep{podell2024sdxl} yields slightly better performance, we opt for SD-v1.5 as the default due to its faster image generation speed (3.8 s/image \vs 11.3 s/image).

\subsection{Efficiency Comparison}
\begin{wraptable}[11]{r}{0.5\textwidth}
\small
\begin{center}
\vspace{-16pt}
\caption{\RebuttalRevision{\textbf{Efficiency comparison}. For each method, we present the average inference latency per instance and peak GPU memory. Experiments are conducted on a single RTX A6000 Ada GPU.}}
\label{tab:efficiency}
\vspace{-5pt}
\resizebox{\linewidth}{!}{
\begin{tabular}{lcccc}
\toprule
Method  & Avg. Latency $\downarrow$ & GPU Memory $\downarrow$ & CHAIR$_S$ $\downarrow$  \\ 
\midrule
Regular  &  3.44 s {\tiny ($\times$1.00)} &  15778 MB  {\tiny ($\times$1.00)} & 55.0\\
VCD &  6.91 s {\tiny ($\times$2.01)} &  16634 MB  {\tiny ($\times$1.05)}  & 54.4 \\
OPERA & 24.70 s  {\tiny ($\times$7.18)} &  22706 MB  {\tiny ($\times$1.44)}& 52.6 \\
Woodpecker & 10.68 s {\tiny ($\times$3.10)} & 22199 MB  {\tiny ($\times$1.41)} & 57.6 \\
HALC & 22.61 s {\tiny ($\times$6.51)} &  23084 MB {\tiny ($\times$1.46)}& 51.0 \\
\rowcolor{gray!20}
\textbf{Ours} &  13.89 s {\tiny ($\times$4.04)}  & 19119 MB  {\tiny ($\times$1.21)}& 48.8\\
\bottomrule
\end{tabular}
}
\end{center}
\end{wraptable}

\RebuttalRevision{
In Table~\ref{tab:efficiency}, we compare the efficiency of our approach with other methods on the CHAIR benchmark using the LLaVA-1.5 model, with the maximum token length set to 128. Our approach involves two queries and incorporates a text-to-image generative model to mitigate hallucinations, resulting in a 4.04$\times$ increase in latency and a 1.21$\times$ increase in GPU memory usage. Specifically, our method consists of three stages: initial response generation, image generation, and response self-correction, which take an average of 3.4 seconds, 3.8 seconds, and 6.6 seconds per instance, respectively. Compared to other approaches, while our method is slower than regular decoding and contrastive decoding-based methods, it demonstrates efficiency advantages over OPERA and HALC. Note that our approach also achieves the lowest hallucination rates among all compared methods. In Appendix~\ref{subsec:speedup}, we discuss several strategies to accelerate our approach, such as limiting the length of the initial response and reducing the number of inference steps in the diffusion process.
}
% \vspace{-4pt}
\section{Conclusion}
% \vspace{-3pt}
In this work, we present self-correcting Decoding with Generative Feedback (DeGF), a novel training-free approach that leverages feedback from text-to-image generative models to recursively improve the accuracy of generated responses. Specifically, we generate a new image based on the initial response given by LVLMs, which serves as a visual reference and provides token-level feedback for mitigating hallucinations. Building on this, we propose a corresponding self-correcting decoding algorithm that measures the discrepancy between next-token predictions conditioned on the original and generated images, selecting either contrastive or complementary decoding to reduce the likelihood of hallucinatory responses. Extensive experimental results across six benchmarks demonstrate that our DeGF consistently outperforms state-of-the-art methods in mitigating hallucinations in LVLMs.

\clearpage

\section*{Acknowledgements}
This work has been funded in part by the Army Research Laboratory (ARL) award W911NF-23-2-0007 and W911QX-24-F-0049, DARPA award FA8750-23-2-1015, and ONR award N00014-23-1-2840. MM and LPM are partially supported by Meta and National Institutes of Health awards R01MH125740, R01MH132225, and R21MH130767. RS is supported in part by the ONR grant N00014-23-1-2368.

\section*{Ethics Statement}
Our work focuses on enhancing the reliability of current large vision-language models. Our research does not involve human subjects, sensitive data, or any practices that pose privacy or security concerns. We also discuss the broader ethical and societal implications of this work in Appendix~\ref{sec:limitation}.

\section*{Reproducibility Statement}
The large vision-language models utilized in our experiments, such as LLaVA and InstructBLIP, are open-source and publicly available. We have detailed our experimental setup, including hyperparameter configurations, prompts, and other key design choices, in Section~\ref{sec:experiment} of the main paper and Section~\ref{sec:detail} of the Appendix to ensure reproducibility. Code is publicly available at \url{https://github.com/zhangce01/DeGF}.



\bibliography{iclr2025_conference}
\bibliographystyle{iclr2025_conference}

\clearpage
\appendix
\renewcommand{\thesection}{\Alph{section}}
\renewcommand\thefigure{\Alph{section}\arabic{figure}} 
\renewcommand\thetable{\Alph{section}\arabic{table}}  
\setcounter{section}{0}
\setcounter{figure}{0} 
\setcounter{table}{0} 

{\LARGE\sc Self-Correcting Decoding with Generative Feedback for Mitigating Hallucinations in Large Vision-Language Models\par}

{\LARGE\sc Appendix\par} \vspace{10pt}

In this supplementary document, we provide additional details and experimental results to enhance understanding and insights into our method.
This supplementary document is organized as follows:
\begin{itemize}[leftmargin=0.5cm, itemindent=0cm, itemsep=4pt,topsep=4pt,parsep=0pt]
    \item The limitations and broader impacts of this work are discussed in Section~\ref{sec:limitation}.
    \item Additional experimental details, including further implementation details, descriptions of other implemented baselines, and license information for the utilized code and datasets, are provided in Section~\ref{sec:detail}.
    \item Additional experimental results are presented in Section~\ref{sec:moreablation}.
    \item More case studies and GPT-4V-aided evaluations are provided in Section~\ref{sec:case}.
    \item Potential directions for future work are discussed in Section~\ref{sec:future}.
\end{itemize}

\section{Limitations and Broader Impacts}
\label{sec:limitation}
\textbf{Limitations}. Although our method effectively mitigates hallucinations in LVLMs, it relies on pre-trained text-to-image generative models, which introduces additional computational complexity. The process of generating images also adds time, potentially slowing down LVLM response generation and making it less suitable for real-time applications. However, our method is training-free, reducing the overhead typically associated with fine-tuning large models, and offering broader applicability across various tasks. Moreover, the use of generative feedback improves the model's ability to verify and correct responses, particularly in complex scenarios. Thus, while the computational trade-offs may limit real-time performance, our method excels in settings where accuracy and reliability are prioritized over speed. We also hope that advances in efficient diffusion-based models will improve the feasibility of our approach in real-world applications in the future.

\textbf{Broader Impacts}. In this work, our goal is to develop more reliable large vision-language models (LVLMs) by incorporating feedback from generative models. By using this feedback mechanism, we aim to address a critical issue faced by current multi-modal models: hallucinations, where models produce responses that are inconsistent with the visual input. Hallucinations not only degrade model performance but also pose risks in real-world applications by generating inaccurate or misleading information. Our approach leverages the strengths of generative models to detect and mitigate these hallucinations, improving the overall accuracy and reliability of LVLMs. In doing so, we contribute to enhancing trustworthiness and reducing the spread of misinformation in systems that rely on multi-modal AI, making them safer and more effective for a wide range of applications.

\section{More Experimental Details}
\label{sec:detail}
\subsection{Benchmarks and Metrics}
We conduct extensive experiments on the following benchmarks:
\begin{itemize}
    \item \looseness=-1 \textbf{POPE~\citep{li2023evaluating}} is a widely used benchmark for assessing object hallucinations in LVLMs. It tests the models with yes-or-no questions regarding the presence of specific objects, such as, ``\texttt{Is there a \{object\} in the image?}'' The benchmark draws data from three existing datasets: MSCOCO~\citep{lin2014microsoft}, A-OKVQA~\citep{schwenk2022okvqa}, and GQA~\citep{hudson2019gqa}, and comprises three distinct subsets—\textit{random}, \textit{popular}, and \textit{adversarial}—based on how the negative samples are generated. For each dataset setting, the benchmark provides 6 questions per image, resulting in 3,000 test instances. We evaluate the performance of different methods using four metrics: accuracy, precision, recall, and F1 score.
    \item \textbf{CHAIR~\citep{rohrbach2018object}} evaluates object hallucinations in open-ended captioning tasks. It prompts the LVLMs to describe specific images selected from a random sample of 500 images from the MSCOCO validation set and assesses performance based on two metrics:
    \begin{equation}
    \label{eq:chair metrics}
        \text{CHAIR}_I =  \frac{\text{\# hallucinated objects}}{\text{\# all objects mentioned}}, \quad\text{CHAIR}_S =  \frac{\text{\# sentences with hallucinated object}}{\text{\# all sentences}}.
    \end{equation}
    Additionally, we assess the recall and the average length of the generated responses.
    \item \textbf{MME-Hallucination~\citep{fu2023mme}} is a comprehensive benchmark for LVLMs consisting of four subsets: \textit{existence} and \textit{count} for object-level hallucinations, and \textit{position} and \textit{color} for attribute-level hallucinations. Each subset includes 30 images and 60 questions, with two questions per image. Similar to POPE~\citep{li2023evaluating}, these questions are structured as yes-or-no queries, and performance is assessed based on binary accuracy. Following the official implementation, the reported score is calculated by combining accuracy and accuracy+, where accuracy is based on individual questions, and accuracy+ is based on images where both questions are answered correctly.
    \item \textbf{MMBench~\citep{liu2025mmbench}} is a comprehensive evaluation benchmark designed to assess the multimodal understanding and reasoning capabilities of AI models. It focuses on tasks requiring the integration of visual and textual information, testing a model's ability to handle diverse, real-world scenarios. In particular, MMBench employs a hierarchical ability taxonomy, designating Perception and Reasoning as Level-1 (L-1) abilities. It further refines the taxonomy by incorporating more detailed ability dimensions, organizing them into six Level-2 (L-2) and twenty Level-3 (L-3) dimensions.
    \item \textbf{MMVP~\citep{tong2024eyes}} collects CLIP-blind pairs and evaluates the fine-grained visual recognition capabilities of LVLMs. It consists of 150 image pairs, each accompanied by a binary-option question. Each image is queried independently, and for a given pair, the LVLM's response is considered correct only if both associated questions are answered accurately.
    \item \textbf{LLaVA-Bench\footnote{\url{https://huggingface.co/datasets/liuhaotian/llava-bench-in-the-wild}.}} provides 24 images featuring complex scenes, memes, paintings, and sketches, along with 60 challenging questions. We select examples from this dataset to provide qualitative comparisons between the responses generated by different decoding methods. We also follow \citet{yin2023woodpecker} to evaluate the accuracy and detailedness of generated responses of different methods using the advanced LVLM, GPT-4V\footnote{\url{https://openai.com/index/gpt-4v-system-card}.}.
\end{itemize}

\subsection{More Implementation Details}
In our experiments, we adhere to the default query format for the input data used in both LLaVA-1.5~\citep{liu2023visual} and InstructBLIP~\citep{dai2024instructblip}. Additionally, we set $\alpha_1 = 3$, $\alpha_2 = 1$, and $\gamma = 0.1$ by default in our decoding process. We follow VCD~\citep{leng2024mitigating} to implement adaptive plausibility constraints~\citep{li2023contrastive}:
\begin{gather}
p_{\theta}(y_t) = 0, \quad \text{if} \,\,\, y_t \notin \mathcal{V}(y_{<t})\nonumber\\
    \text{where}\,\,\,\mathcal{V}(y_{<t}) = \{y_t \in \mathcal{V}: p_{\theta}(y_t | v, \mathbf{x}, \mathbf{y}_{<t}) \geq \beta \max_w  p_{\theta}(w | v, \mathbf{x}, \mathbf{y}_{<t})\}
    \label{eq:adaptive}
\end{gather}
Here, $\mathcal{V}$ is the whole vocabulary of LVLM, and hyperparameter $\beta \in [0, 1]$ controls the truncation of the next token distribution. A larger $\beta$ indicates more aggressive truncation, keeping only the high-probability tokens. In our implementation, we set the logits for $y_t \notin \mathcal{V}(y_{<t})$ to $-\infty$. By default, we set $\beta=0.1$ in the open-ended CHAIR benchmark and $\beta=0.25$ for other tasks. All experiments are conducted on a single 48GB NVIDIA RTX 6000 Ada GPU. 

Recall that in our method, we use a text-to-image generative model to reverse the image-to-text response generation process by producing a new image from the initial response. To ensure the new image is both high-quality and relevant, we aim to generate specific descriptions for the given visual content. Thus, we slightly modify the initial query prompt for each evaluated benchmark: 
\begin{itemize}
    \item \textbf{POPE~\citep{li2023evaluating}, MME-Hallucination~\citep{fu2023mme}, and MMVP~\citep{tong2024eyes}}. In POPE, MME-Hallucination, and MMVP benchmarks, models are tested with yes-or-no/binary selection questions, such as, ``\texttt{Is there a \{object\} in the image?}'' To obtain more detailed explanations and descriptions of the original image, we modify the prompt by adding, ``\texttt{Briefly describe relevant details.}'' This encourages the model to provide not only a yes-or-no answer but also additional visual information.
    \item \textbf{CHAIR~\citep{rohrbach2018object}}. For the CHAIR benchmark, we retain the original prompt, ``\texttt{Please describe this image in detail.}'' as it effectively prompts the model to provide comprehensive visual details from the original image.
\end{itemize}
Note that for the second query, where both the original and generated images are used as input, we apply the original prompt to ensure a fair comparison.
% \clearpage

\subsection{Details of Other Baselines}
In this work, we mainly compare the performance of our DeGF with three state-of-the-art approaches: VCD~\citep{leng2024mitigating}, M3ID~\citep{favero2024multi}, and RITUAL~\citep{woo2024ritual}. The method and implementation details for these approaches are provided below:
\begin{itemize}
\item \textbf{VCD~\citep{leng2024mitigating}} contrasts output distributions derived from original and distorted visual inputs. Specifically, given a textual query ${x}$ and a visual input ${v}$, the model generates two distinct output distributions: one conditioned on the original ${v}$ and the other on the distorted visual input ${v'}$, which is derived by applying pre-defined distortions (i.e., Gaussian noise mask) to ${v}$. 
Then, a new contrastive probability distribution is computed by:
\begin{gather}
p_{vcd}\left(y_t\right) =\mathsf{Softmax}\left[ (1+\alpha) 
f_\theta\left(y | v, \mathbf{x}, \mathbf{y}_{<t}\right) -\alpha f_\theta\left(y | v', \mathbf{x}, \mathbf{y}_{<t}\right)\right].
\end{gather}
In our implementation, we follow the default setting in VCD~\citep{leng2024mitigating} and set $\alpha=1$ for reproduction. To generate $v'$, we use a total of 500 noise steps.

\item \textbf{M3ID~\citep{favero2024multi}} contrasts output distributions derived from original visual inputs and pure text inputs without visual information. The final probability distribution is
\begin{equation}
    p_{m3id}\left(y_t\right) = \mathsf{Softmax}\left[f_\theta\left(y | v, \mathbf{x}, \mathbf{y}_{<t}\right) + \frac{1-e^{-\lambda t}}{e^{-\lambda t}} \left( f_\theta\left(y | v, \mathbf{x}, \mathbf{y}_{<t}\right) - f_\theta\left(y |  \mathbf{x}, \mathbf{y}_{<t}\right)\right) \right].
\end{equation}
Similarly, we follow their recommended best practice and set the hyperparameter $\lambda$, which balances the conditioned model and unconditioned model, to $0.02$.

\item \textbf{RITUAL~\citep{woo2024ritual}} applies common image transformations (\eg, crop, flip, color jitter, \etc) to the original visual input $v$,
This results in a transformed version of the visual input, $v^{(T)}$. Then,  RITUAL utilizes both the original and transformed images to generate the response and this dual-input approach significantly reduces the likelihood of hallucinatory outputs.
The probability distribution is calculated as follows:
\begin{equation}
p_{ritual}\left(y_t\right) =\mathsf{Softmax}\left[f_\theta\left(y | v, \mathbf{x}, \mathbf{y}_{<t}\right) +\kappa f_\theta\left(y | v^{(T)}, \mathbf{x}, \mathbf{y}_{<t}\right)\right].
\label{eq:ritual_sampling}
\end{equation}
Here, $\kappa$ is a balancing hyperparameter, adjusting the contribution of the transformed input relative to the original. We follow their official implementation to set $\kappa=3$ in default.

\end{itemize}

\subsection{Dataset and Code Licensing}
\textbf{Datasets}. We list the known license information for the datasets below: POPE~\citep{li2023evaluating} and MMVP~\citep{tong2024eyes} benchmarks are licensed under MIT License.
CHAIR~\citep{rohrbach2018object} is made available under the BSD 2-Clause License.
LLaVA-Bench is available under Apache-2.0 License.
MME-Hallucination~\citep{fu2023mme} benchmark dataset is collected by Xiamen University for academic research only.

\textbf{Code}. In this work, we also use some code implementations from the existing codebase: LLaVA~\citep{liu2023visual} and VCD~\citep{leng2024mitigating} are licensed under the Apache-2.0 License.
InstructBLIP~\citep{dai2024instructblip} is under BSD-3-Clause License.  RITUAL~\citep{woo2024ritual} is licensed under MIT License.

\section{More Experimental Results and Analysis}
\label{sec:moreablation}

\begin{table}[t]
    % \begin{minipage}[t!]{0.5\textwidth}
        \begin{center}
        \begin{small}
        
        \setlength{\tabcolsep}{5pt} % base value: 6pt
        \caption{\RebuttalRevision{\textbf{Results on CHAIR~\citep{rohrbach2018object} benchmark.} We limit the maximum number of new tokens to 128. Lower ($\downarrow$) CHAIR$_S$, CHAIR$_I$ and higher ($\uparrow$) recall and length indicate better performance. The best results in each setting are \textbf{bolded}, and the second-best are \underline{underlined}.}}
        \vspace{-6pt}
        \label{tab:CHAIR-128}
        \resizebox{\textwidth}{!}{
        \begin{tabular}{lcccccccc}
            \toprule
              \multirow{2}{*}[-0.5ex]{\textbf{Method}}  &  \multicolumn{4}{c}{\textbf{LLaVA-1.5}} & \multicolumn{4}{c}{\textbf{InstructBLIP}} \\
             \cmidrule(lr){2-5}\cmidrule(lr){6-9}
                 & CHAIR$_S$ $\downarrow$ & CHAIR$_I$ $\downarrow$ & Recall $\uparrow$ & Length $\uparrow$   & CHAIR$_S$ $\downarrow$ & CHAIR$_I$ $\downarrow$ & Recall $\uparrow$ & Length $\uparrow$ \\
             \midrule
             Regular & 55.0 & 16.3 & 71.9 & \textbf{97.3} &  57.0 & 17.6  & 68.3 & \textbf{100.4}\\
              VCD & 54.4 & 16.6 & 75.1 & \underline{97.0} &  60.4 & 17.8  & \textbf{72.5} & 99.9\\
              M3ID & 56.6 & 15.7 & \textbf{76.8} & 94.5 & 62.2 & 18.1 & 71.9 & 99.8\\
              RITUAL & \underline{49.6} & \underline{14.8}  & 74.7 & 96.2 &  \textbf{48.4} & \underline{14.5} & \underline{72.2} & \underline{100.0}\\
              Woodpecker & 57.6 & 16.7  & 70.3 &93.2 & 60.8 & 17.6 & 69.7 & 97.6 \\
              HALC & 51.0 &  \underline{14.8} & 75.3& 95.8 & 53.8  & 15.7 & 71.9 & 99.1 \\
              \cc \textbf{Ours} & \cc \textbf{48.8} & \cc \textbf{14.6}  & \cc \underline{76.0} & \cc 96.4 & \cc  \underline{49.2} & \cc \textbf{14.4}  & \cc \underline{72.2}  & \cc 98.9 \\
            \bottomrule
        \end{tabular}
        
        }
        \end{small}
        \end{center}
        \vspace{-10pt}
\end{table}

\begin{table}[t]
    % \begin{minipage}[t!]{0.5\textwidth}
        \begin{center}
        \begin{small}
        
        \setlength{\tabcolsep}{5pt} % base value: 6pt
        \caption{\RebuttalRevision{\textbf{Results on CHAIR~\citep{rohrbach2018object} benchmark.} We limit the maximum number of new tokens to 256. Lower ($\downarrow$) CHAIR$_S$, CHAIR$_I$ and higher ($\uparrow$) recall and length indicate better performance. The best results in each setting are \textbf{bolded}, and the second-best are \underline{underlined}.}}
        \vspace{-6pt}
        \label{tab:CHAIR-256}
        \resizebox{\textwidth}{!}{
        \begin{tabular}{lcccccccc}
            \toprule
              \multirow{2}{*}[-0.5ex]{\textbf{Method}}  &  \multicolumn{4}{c}{\textbf{LLaVA-1.5}} & \multicolumn{4}{c}{\textbf{InstructBLIP}} \\
             \cmidrule(lr){2-5}\cmidrule(lr){6-9}
                 & CHAIR$_S$ $\downarrow$ & CHAIR$_I$ $\downarrow$ & Recall $\uparrow$ & Length $\uparrow$   & CHAIR$_S$ $\downarrow$ & CHAIR$_I$ $\downarrow$ & Recall $\uparrow$ & Length $\uparrow$ \\
             \midrule
             Regular & 58.0 & 17.7 & 74.1 & \textbf{106.3} & 61.0 &  18.2 & 68.9 & \textbf{112.0}\\
              VCD & 58.2 & 16.7 & \underline{78.0}& \underline{103.5} & 63.0& 18.6  & \textbf{72.9}  & \underline{106.3} \\
              M3ID & 56.8 & 16.1& \textbf{80.7} & 98.2 &65.8  & 19.9& \underline{72.4} & 102.7\\
              RITUAL & \underline{51.0} & \underline{15.1}  & 76.0 & 100.9 & \underline{50.4} & \underline{15.3} & 72.0 & 102.0\\
              \cc \textbf{Ours} & \cc \textbf{49.8} & \cc \textbf{14.7}  & \cc 77.2 & \cc 103.3 & \cc \textbf{49.8}   & \cc \textbf{15.1}  & \cc 72.3  & \cc 103.3  \\
            \bottomrule
        \end{tabular}
        
        }
        \end{small}
        \end{center}
        \vspace{-10pt}
\end{table}

\begin{table*}[t!]
    % \begin{minipage}[t!]{0.5\textwidth}
        \begin{center}
        \begin{small}
        \setlength{\tabcolsep}{6pt} % base value: 6pt
        \caption{\looseness=-1\RebuttalRevision{\textbf{Results on MME-Hallucination~\citep{fu2023mme} benchmark.} We report the average MME scores along with the standard deviation across three random seeds for each subset. We also report the total scores achieved by the different methods across all four subsets in the final column. Higher scores ($\uparrow$) indicate better performance. The best results  are \textbf{bolded}, and the second-best are \underline{underlined}.}}
        % \martin{consistent with confidence intervals; either include error bars in all tables, or include none in main text and add in Appendix?}}
        \label{tab:MME-full}
        \vspace{-5pt}
        \resizebox{\textwidth}{!}{
        \begin{tabular}{llccccc}
            \toprule
             \multirow{2}{*}[-0.5ex]{\textbf{Model}} & \multirow{2}{*}[-0.5ex]{\textbf{Method}}  &  \multicolumn{2}{c}{\textbf{Object-level}} & \multicolumn{2}{c}{\textbf{Attribute-level}} & \multirow{2}{*}[-0.5ex]{\textbf{Total Score $\uparrow$}}\\
             \cmidrule(lr){3-4}\cmidrule(lr){5-6}
              &   & Existence $\uparrow$ & Count $\uparrow$ & Position $\uparrow$ & Color $\uparrow$   & \\
             \midrule
            \multirow{7}{*}{{\textbf{LLaVA-1.5}}} & Regular &  173.75 {\tiny ($\pm4.79$)} & 121.67 {\tiny ($\pm12.47$)} & 117.92 {\tiny ($\pm3.69$)\phantom{0}} & 149.17 {\tiny ($\pm7.51$)\phantom{0}} &  562.50 {\tiny ($\pm3.96$)\phantom{0}} \\
             & DoLa  & 176.67 {\tiny ($\pm2.89$)} & 113.33 {\tiny ($\pm10.41$)} & 90.55 {\tiny ($\pm8.22$)} & 141.67 {\tiny ($\pm7.64$)\phantom{0}} &  522.22 {\tiny ($\pm16.78$)}  \\
             & OPERA  & 183.33 {\tiny ($\pm6.45$)} & 137.22 {\tiny ($\pm6.31$)\phantom{0}} & 122.78 {\tiny ($\pm2.55$)\phantom{0}} &155.00 {\tiny ($\pm5.00$)\phantom{0}}  &  598.33 {\tiny ($\pm10.41$)}  \\
             & VCD  & 186.67 {\tiny ($\pm5.77$)} & 125.56 {\tiny ($\pm3.47$)\phantom{0}} & 128.89 {\tiny ($\pm6.73$)\phantom{0}} & 139.45 {\tiny ($\pm12.51$)} &  580.56 {\tiny ($\pm15.13$)}  \\
             & M3ID  & 186.67 {\tiny ($\pm5.77$)} &  128.33 {\tiny ($\pm10.41$)} & \underline{131.67} {\tiny ($\pm5.00$)\phantom{0}} & 151.67 {\tiny ($\pm20.88$)} &  598.11 {\tiny ($\pm20.35$)}  \\
             & RITUAL  & \underline{187.50} {\tiny ($\pm2.89$)} & \underline{139.58} {\tiny ($\pm7.64$)\phantom{0}}  & 125.00 {\tiny ($\pm10.27$)} & \underline{164.17} {\tiny ($\pm6.87$)\phantom{0}} & \underline{616.25} {\tiny ($\pm20.38$)}     \\
             & \cc \textbf{Ours} & \cc \textbf{188.33} {\tiny ($\pm2.89$)} & \cc \textbf{150.00} {\tiny ($\pm7.64$)\phantom{0}}  & \cc \textbf{133.89} {\tiny ($\pm3.85$)\phantom{0}} & \cc \textbf{172.22} {\tiny ($\pm3.47$)\phantom{0}} & \cc \textbf{644.44} {\tiny ($\pm9.18$)\phantom{0}} \\
            %  \arrayrulecolor{gray!50}\cmidrule(lr){2-4}
            %  & \textbf{\Ours}+VCD & 20.0 & 6.8 \\
            %  & \textbf{\Ours}+M3ID & 18.0 & 5.7 \\
            \midrule
            \multirow{7}{*}{{\textbf{InstructBLIP}}} & Regular & 160.42 {\tiny ($\pm5.16$)} & 79.17 {\tiny ($\pm8.22$)\phantom{0}} & \textbf{79.58} {\tiny ($\pm8.54$)\phantom{0}} & 130.42 {\tiny ($\pm17.34$)}  &  449.58 {\tiny ($\pm24.09$)}    \\
             & DoLa  & 175.00 {\tiny ($\pm5.00$)} & 55.00 {\tiny ($\pm5.00$)\phantom{0}} & 48.89 {\tiny ($\pm3.47$)\phantom{0}} & 113.33 {\tiny ($\pm6.67$)\phantom{0}}  &  392.22 {\tiny ($\pm7.88$)\phantom{0}}   \\
             & OPERA  & 175.00 {\tiny ($\pm3.33$)} & 61.11 {\tiny ($\pm3.47$)\phantom{0}} & 53.89 {\tiny ($\pm1.92$)\phantom{0}} &120.55 {\tiny ($\pm2.55$)\phantom{0}}  &  410.56 {\tiny ($\pm9.07$)\phantom{0}}     \\
             & VCD  & 158.89 {\tiny ($\pm5.85$)} &  \textbf{91.67} {\tiny ($\pm18.34$)} & 66.11 {\tiny ($\pm9.76$)\phantom{0}} & 121.67 {\tiny ($\pm12.58$)} & 438.33 {\tiny ($\pm16.07$)}  \\
             & M3ID  & 160.00 {\tiny ($\pm5.00$)} & 87.22 {\tiny ($\pm22.63$)} & \underline{69.44} {\tiny ($\pm9.18$)\phantom{0}} & 125.00 {\tiny ($\pm7.64$)\phantom{0}} & 441.67 {\tiny ($\pm17.32$)}    \\
             & RITUAL  & \underline{182.50} {\tiny ($\pm6.45$)} & 74.58 {\tiny ($\pm5.99$)\phantom{0}}  & 67.08 {\tiny ($\pm10.31$)} & \underline{139.17} {\tiny ($\pm0.96$)\phantom{0}} & \underline{463.33} {\tiny ($\pm12.40$)}   \\
             & \cc \textbf{Ours} & \cc \textbf{186.67} {\tiny ($\pm2.89$)} & \cc \underline{89.44} {\tiny ($\pm8.22$)\phantom{0}}  & \cc 58.33 {\tiny ($\pm4.41$)\phantom{0}} & \cc \textbf{150.00} {\tiny ($\pm1.89$)\phantom{0}} & \cc \textbf{484.44} {\tiny ($\pm11.34$)} \\
             \midrule
            \multirow{5}{*}{{\textbf{Qwen-VL}}} & Regular & 155.00 {\tiny ($\pm3.54$)} & 127.67 {\tiny ($\pm13.36$)} & {131.67} {\tiny ($\pm7.73$)\phantom{0}} & 173.00 {\tiny ($\pm9.75$)\phantom{0}}  &  587.33 {\tiny ($\pm31.06$)}    \\
             % & DoLa  & 175.00 {\tiny ($\pm5.00$)} & 55.00 {\tiny ($\pm5.00$)\phantom{0}} & 48.89 {\tiny ($\pm3.47$)\phantom{0}} & 113.33 {\tiny ($\pm6.67$)\phantom{0}}  &  392.22 {\tiny ($\pm7.88$)\phantom{0}}   \\
             % & OPERA  & 175.00 {\tiny ($\pm3.33$)} & 61.11 {\tiny ($\pm3.47$)\phantom{0}} & 53.89 {\tiny ($\pm1.92$)\phantom{0}} &120.55 {\tiny ($\pm2.55$)\phantom{0}}  &  410.56 {\tiny ($\pm9.07$)\phantom{0}}     \\
             & VCD  & 156.00 {\tiny ($\pm6.52$)} &  131.00 {\tiny ($\pm6.19$)\phantom{0}} & 128.00 {\tiny ($\pm3.61$)\phantom{0}} & 181.67 {\tiny ($\pm5.14$)\phantom{0}} & 596.67 {\tiny ($\pm11.61$)}  \\
             & M3ID  & \underline{178.33} {\tiny ($\pm2.89$)} & 143.33 {\tiny ($\pm2.89$)\phantom{0}} & {150.00} {\tiny ($\pm2.89$)\phantom{0}} & 175.00 {\tiny ($\pm5.00$)\phantom{0}} & 646.66 {\tiny ($\pm8.50$)\phantom{0}}    \\
             & RITUAL  & \underline{178.33} {\tiny ($\pm2.89$)} & \underline{142.22} {\tiny ($\pm16.19$)}  & \textbf{156.66} {\tiny ($\pm2.89$)\phantom{0}} & \textbf{178.33} {\tiny ($\pm2.89$)\phantom{0}} & \underline{655.55} {\tiny ($\pm14.99$)}   \\
             & \cc \textbf{Ours} & \cc \textbf{180.00} {\tiny ($\pm0.00$)} & \cc \textbf{148.89} {\tiny ($\pm6.74$)\phantom{0}}  & \cc \underline{155.00} {\tiny ($\pm7.64$)\phantom{0}} & \cc \textbf{178.33} {\tiny ($\pm2.89$)\phantom{0}} & \cc \textbf{662.22} {\tiny ($\pm4.37$)\phantom{0}} \\
            \bottomrule
        \end{tabular}
        }
        \vspace{-15pt}
        \end{small}
        \end{center}
        
\end{table*}

\RebuttalRevision{
\subsection{Additional Results on CHAIR}
In Table~\ref{tab:CHAIR-128} and Table~\ref{tab:CHAIR-256}, we present performance comparisons on the CHAIR benchmark with maximum number of tokens set to 128 and 256. The results indicate that our approach also achieves competitive performance across two LVLMs in mitigating hallucinations during long-sequence generation scenarios.
}
\RebuttalRevision{
\subsection{Full Results on MME-Hallucination}
In Table~\ref{tab:MME-full}, we present the full results on the MME-Hallucination benchmark across three LVLMs.
}

\begin{table}[t]
\centering
\caption{
\RebuttalRevision{
\textbf{Detailed results on MMBench benchmark}. Abbreviations adopted: LR for Logical Reasoning; AR for Attribute Reasoning; RR for Relation Reasoning; FP-S for Fine-grained Perception (Single Instance); FP-C for Fine-grained Perception (Cross Instance); CP for Coarse Perception. The best results are \textbf{bolded}.}}
\vspace{5pt}
\label{tab:MMBench Full}
\begin{tabular}{lccccccc}
\toprule
Method & LR & AR & RR & FP-S & FP-C & CP & Overall \\
\midrule
Regular & 30.51 & 71.36 & 52.17 & 67.58 & \textbf{58.74} & 76.35 & 64.09 \\
VCD     & 30.51 & \textbf{73.37} & 53.04 & 67.92 & 57.34 & 77.03 & 64.60 \\
M3ID    & 30.51 & 72.36 & 53.04 & 67.58 & 57.34 & \textbf{77.36} & 64.43 \\
RITUAL  & 28.81 & 72.86 & 54.78 & 65.87 & 58.04 & 76.01 & 63.83 \\
\cc \textbf{Ours}    & \cc \textbf{31.36} & \cc 70.85 & \cc \textbf{60.87} & \cc \textbf{68.60} & \cc \textbf{58.74} & \cc \textbf{77.36} & \cc \textbf{65.46} \\
\bottomrule
\end{tabular}
\end{table}

\begin{table}[t]
\centering
\caption{
\RebuttalRevision{\textbf{Detailed results on MMVet benchmark with regular sampling}. Abbreviations adopted: Rec for Recognition, OCR for Optical Character Recognition, Know for Knowledge, Gen for Language Generation, Spat for Spatial Awareness, Math for Mathematics. The best results are \textbf{bolded}, and the second best are \underline{underlined}.}}
\vspace{5pt}
\label{tab:mmvet_sample}
\begin{tabular}{lccccccc}
\toprule
Method & Rec & OCR & Know & Gen & Spat & Math & Total \\
\midrule
Regular & 30.8 & 19.0 & 14.5 & 17.9 & 26.9 & \textbf{11.5} & 26.1 \\
VCD     & 35.6 & 21.9 & 18.3 & \underline{21.9} & \underline{28.9} & 3.8 & \underline{30.9} \\
M3ID    & 35.0 & 19.7 & 18.8 & 19.0 & 26.0 & 7.7 & 29.9 \\
RITUAL  & \textbf{36.3} & \underline{20.6} & \textbf{19.5} & 21.1 & 24.7 & 7.7 & 30.6 \\
\textbf{Ours} & \underline{35.9} & \textbf{27.2} & \underline{19.2} & \textbf{22.4} & \textbf{30.4} & \textbf{11.5} & \textbf{33.0} \\
\bottomrule
\end{tabular}
\end{table}

\begin{table}[t]
\centering
\caption{
\RebuttalRevision{\textbf{Detailed results on MMVet benchmark with greedy decoding}. Abbreviations adopted: Rec for Recognition, OCR for Optical Character Recognition, Know for Knowledge, Gen for Language Generation, Spat for Spatial Awareness, Math for Mathematics. The best results are \textbf{bolded}, and the second best are \underline{underlined}.}}
\vspace{5pt}
\label{tab:mmvet_greedy}
\begin{tabular}{lccccccc}
\toprule
Method & Rec & OCR & Know & Gen & Spat & Math & Total \\
\midrule
Greedy  & 37.0 & 22.6 & 17.5 & 20.2 & 24.9 & 7.7 & 31.8 \\
VCD     & \textbf{38.2} & 22.8 & \textbf{22.5} & \textbf{24.6} & 25.1 & 3.8 & \underline{33.4} \\
M3ID    & \underline{37.9} & \textbf{23.6} & \underline{20.4} & \underline{20.7} & \underline{26.0} & \underline{11.5} & 33.2 \\
RITUAL  & 35.6 & 21.7 & 18.9 & 19.9 & 24.7 & 7.7 & 30.6 \\
\textbf{Ours} & \underline{37.9} & \textbf{25.0} & 20.2 & 19.5 & \textbf{32.8} & \textbf{15.0} & \textbf{34.0} \\
\bottomrule
\end{tabular}
\end{table}

\begin{table}[t]
\centering
% \vspace{-10pt}
\caption{\RebuttalRevision{\textbf{Results on POPE using greedy decoding}}.}
% \vspace{5pt}
    \label{tab:greedy}
    \resizebox{0.5\textwidth}{!}{
    \begin{tabular}{lcccccc}
            \toprule
              \multirow{2}{*}[-0.5ex]{\textbf{Values}}  &  \multicolumn{4}{c}{\textbf{POPE}} \\
             \cmidrule(lr){2-5}
                 & {Acc.}  & {Prec.}  & {Rec.}  & {F1} \\
             \midrule
             Greedy & 87.73 & 88.19 & 87.13 & 87.66  \\
             VCD & 87.47 & 86.64 & \textbf{88.60} & 87.61 \\
             M3ID & 89.07 & 89.54 & 88.47 & 89.00 \\
             RITUAL & 89.23 & 90.17 & 88.07 & 89.11 \\
             \cc \textbf{Ours} & \cc \textbf{89.40} & \cc \textbf{94.44} & \cc 83.73 & \cc \textbf{88.76} \\
            \bottomrule
        \end{tabular}
}
% \vspace{-5pt}
\end{table}

\RebuttalRevision{
\subsection{Full Results on MMBench}
In Table~\ref{tab:MMBench Full}, we present the overall performance on the MMBench benchmark, as well as the detailed performance across six Level-2 abilities: Logical Reasoning (LR), Attribute Reasoning (AR), Relation Reasoning (RR), Fine-grained Perception - Single Instance (FP-S), Fine-grained Perception - Cross Instance (FP-C), and Coarse Perception (CP). We follow VCD~\citep{leng2024mitigating} to conduct experiments on the MMBench-\texttt{dev} set.
}
\RebuttalRevision{
\subsection{Results on MM-Vet}
In Table~\ref{tab:mmvet_sample} and Table~\ref{tab:mmvet_greedy}, we present the overall performance on the MM-Vet~\citep{yu2024mmvet} benchmark with random sampling decoding and greedy decoding strategies, respectively. We use LLaVA-1.5 as the LVLM backbone. From the results, we observed that our method consistently outperforms others on the MMVet benchmark. Notably, it significantly excels in the OCR, spatial awareness, and math subsets.
}
\RebuttalRevision{
\subsection{Results on POPE Using Greedy Decoding}
In Table~\ref{tab:greedy}, we present performance comparisons on the POPE benchmark with random sampling from the MS-COCO dataset. The experiment is conducted using the LLaVA-1.5 backbone.
}

\subsection{Effects of \texorpdfstring{$\alpha_1$ and $\alpha_2$}{alpha1 and alpha2} in Self-Correcting Decoding}
In Section~\ref{sec:method}, we present two decoding approaches: complementary decoding and contrastive decoding. We also introduce two balancing hyperparameters, $\alpha_1$ and $\alpha_2$, which control the relative influence of the original and generated images in next-token prediction.
In Table~\ref{tab:alpha1} and Table~\ref{tab:alpha2}, we analyze the effect of varying $\alpha_1$ or $\alpha_2$ while keeping all other hyperparameters at their default settings. The results indicate that our default choice of $\alpha_1=3$ and $\alpha_2=1$ consistently yields the best performance across two benchmarks. Moreover, compared to setting these hyperparameters to 0, which effectively reduces complementary/contrastive decoding to standard decoding, the performance improvements demonstrate that our proposed decoding approaches significantly contribute to the overall effectiveness of DeGF in mitigating hallucinations in LVLMs.

\begin{table}[t]
\centering
% \vspace{-10pt}
\caption{\textbf{Sensitivity analysis of hyperparameter $\alpha_1$}. We present the performance of our approach, based on the LLaVA-1.5 backbone, across two benchmarks for varying values of $\alpha_1$. Note that we fix $\alpha_2=1$ in this experiment.}
% \vspace{5pt}
    \label{tab:alpha1}
    \resizebox{0.7\textwidth}{!}{
    \begin{tabular}{lcccccccc}
            \toprule
              \multirow{2}{*}[-0.5ex]{\textbf{Values}}  &  \multicolumn{4}{c}{\textbf{POPE}} & \multicolumn{2}{c}{\textbf{CHAIR}} \\
             \cmidrule(lr){2-5}\cmidrule(lr){6-7}
                 & {Acc.}  & {Prec.}  & {Rec.}  & {F1}  & CHAIR$_S$  & CHAIR$_I$ \\
             \midrule
             $\alpha_1 =0$ & 87.50 & 86.71 & 87.49 & 87.10 & 22.8 & 7.6 \\
             $\alpha_1 =1$ & 87.97 & 87.28 & 87.34 & 87.31 & 20.6 & 6.9 \\
             $\alpha_1 =2$ & 88.90 & 89.39 & \textbf{87.75} & 88.56 & 19.4 & 6.3 \\
             \cc $\alpha_1 =3$ & \textbf{89.03}\cc & \textbf{91.20}\cc & 86.40\cc & \textbf{88.74}\cc & \textbf{18.4}\cc & \textbf{6.1}\cc  \\
             $\alpha_1 =4$ & 88.67 & 90.56 & 85.28 & 87.84 & 22.6 & 8.1  \\
            \bottomrule
        \end{tabular}
    }
    % \vspace{-10pt}
\end{table}

\begin{table}[t]
\centering
\caption{\textbf{Sensitivity analysis of hyperparameter $\alpha_2$}. We present the performance of our approach, based on the LLaVA-1.5 backbone, across two benchmarks for varying values of $\alpha_2$. Note that we fix $\alpha_1=3$ in this experiment.}
% \vspace{5pt}
    \label{tab:alpha2}
    \resizebox{0.7\textwidth}{!}{
    \begin{tabular}{lcccccccc}
            \toprule
              \multirow{2}{*}[-0.5ex]{\textbf{Values}}  &  \multicolumn{4}{c}{\textbf{POPE}} & \multicolumn{2}{c}{\textbf{CHAIR}} \\
             \cmidrule(lr){2-5}\cmidrule(lr){6-7}
                 & {Acc.}  & {Prec.}  & {Rec.}  & {F1}  & CHAIR$_S$  & CHAIR$_I$ \\
             \midrule
             $\alpha_2 =0$ & 86.77 & 85.17 & 86.58 & 85.87 & 23.6 & 8.2 \\
             \cc $\alpha_2 =1$ & \textbf{89.03}\cc & \textbf{91.20}\cc & 86.40\cc & \textbf{88.74}\cc & \textbf{18.4}\cc & \textbf{6.1}\cc \\
             $\alpha_2 =2$ & 88.73 & 89.86 & \textbf{86.66} & 88.23 & 21.8 & 7.5 \\
             $\alpha_2 =3$ & 88.03 & 87.97 & 86.28 & 87.12 & 22.8 & 7.3 \\
             $\alpha_2 =4$ & 87.13 & 86.52 & 86.16 & 86.34 & 23.6 & 7.9 \\
            \bottomrule
        \end{tabular}
}
\end{table}

\begin{table}[t]
        \begin{center}
        \begin{small}
        \vspace{-10pt}
        \caption{\textbf{Sensitivity analysis of hyperparameter $\beta$}. We present the performance of our approach, based on the LLaVA-1.5 backbone, across two benchmarks for varying values of $\beta$.}
        \vspace{5pt}
        \label{tab:beta}
        \resizebox{0.7\textwidth}{!}{
        \begin{tabular}{lcccccccc}
            \toprule
              \multirow{2}{*}[-0.5ex]{\textbf{Values}}  &  \multicolumn{4}{c}{\textbf{POPE}} & \multicolumn{2}{c}{\textbf{CHAIR}} \\
             \cmidrule(lr){2-5}\cmidrule(lr){6-7}
                 & {Acc.}  & {Prec.}  & {Rec.}  & {F1}  & CHAIR$_S$  & CHAIR$_I$ \\
             \midrule
             $\beta =0$ & 87.17 & 87.45 & 85.30 & 86.36 & 21.2 & 7.1  \\
             $\beta =0.05$ & 88.27 & 89.85 & 86.12 & 87.95 & 19.1 & 6.3 \\
             $\beta =0.1$ & 88.33 & 89.04 & 86.04 & 87.52 & \textbf{18.4}\cc & \textbf{6.1}\cc \\
             $\beta =0.25$ & \textbf{89.03}\cc & \textbf{91.20}\cc & \textbf{86.40}\cc & \textbf{88.74}\cc & 19.3 & 6.5 \\
             $\beta =0.5$ & 87.80 & 88.79 & 85.48 & 87.10 & 20.2 & 6.9 \\
            \bottomrule
        \end{tabular}
        
        }
        \end{small}
        \end{center}
        \vspace{-5pt}
\end{table}

% , we examine the effect of $\alpha_1$ while maintaining $\alpha_2$ at a value of 1. Similarly, Table~\ref{tab:alpha2} presents results with $\alpha_1$ fixed at 3 while varying $\alpha_2$. This ablation study aims to evaluate the effectiveness of our complementary decoding ($\alpha_1$) and contrastive decoding ($\alpha_2$). The results on the CHAIR benchmark indicate that $\alpha_2$ has a more significant impact on the outcomes than $\alpha_1$, as evidenced by the greater variance observed when modifying $\alpha_2$. This suggests that both our complementary and contrastive decoding methods significantly mitigate hallucinations, with contrastive decoding contributing more substantially to the results.

\subsection{Effect of \texorpdfstring{$\beta$}{beta} in Adaptive Plausibility Constraint}
We further conduct an ablation study on $\beta$ introduced in Equation~(\ref{eq:adaptive}), where we vary $\beta$ from 0 to 0.5 while keeping all other hyperparameters fixed. The results in Table~\ref{tab:beta} show that setting $\beta=0$, which imposes no constraint, results in suboptimal performance across both benchmarks. Additionally, in the POPE benchmark, where LVLMs handle yes-or-no questions, a more aggressive truncation with $\beta=0.25$ yields the best performance. In contrast, for the open-ended CHAIR benchmark, a lower value of $\beta=0.1$ leads to the best results.

\subsection{Scaling Up the LVLMs}
We further extend our evaluation to larger-scale 13B variants of the LLaVA-1.5 model to assess the scalability of our approach. Table~\ref{tab:POPE13} compares our experimental results with other state-of-the-art approaches across all three subsets of the POPE benchmark using the 13B-sized LLaVA-1.5 model. We observe that scaling up the LLaVA-1.5 model does not alleviate the hallucination issues, as evidenced by the comparable performance of both the 7B and 13B models. Using the 13B-sized model, our DeGF consistently achieves improved performance across all subsets compared to other approaches, demonstrating its general effectiveness and scalability.

\begin{table}[t]
    \centering
    \small
    \caption{
        \textbf{Results on POPE~\citep{li2023evaluating} benchmark using 13B-sized LLaVA-1.5}. Higher ($\uparrow$) accuracy, precision, recall, and F1 indicate better performance. 
    }
    \vspace{5pt}
    \label{tab:POPE13}
    \setlength{\tabcolsep}{8pt} % base value: 6pt
    \resizebox{0.8\textwidth}{!}{
    \begin{tabular}{cclcccc}
    \toprule
     & \multirow{2}{*}[-2pt]{\textbf{Setup}} & \multirow{2}{*}[-2pt]{\textbf{Method}} & \multicolumn{4}{c}{\textbf{LLaVA-1.5}}  \\
    \arrayrulecolor{gray} \cmidrule(lr){4-7} 
     &  &  & {Acc.} $\uparrow$ & {Prec.} $\uparrow$ & {Rec.} $\uparrow$ & {F1} $\uparrow$  \\
    \midrule
    \multirow{15}{*}[-5pt]{\rotatebox{90}{\textbf{\normalsize MS-COCO}}} & \multirow{5}{*}{Random} 
    & Regular &  82.53 & 78.57 & 89.47 & 83.67 \\
     &  & VCD  &  84.80 & 80.67 & 91.53 & 85.76 \\
     &  & M3ID  & 85.37  & 81.30 & 91.87 & 86.26  \\
     &  & RITUAL  & 87.80  & 84.45 & \textbf{92.67} & \textbf{88.37} \\
     &  & \cc \textbf{Ours} &\cc \textbf{88.40} &\cc \textbf{88.14} &\cc 88.61 &\cc \textbf{88.37}  \\
     \arrayrulecolor{gray}\cmidrule(lr){2-7}
      &  \multirow{5}{*}{Popular} & Regular & 80.53 & 76.17 & 88.87 & 82.03\\
     &  & VCD  &  82.23 & 76.88 & 92.20 & 83.84 \\
     &  & M3ID  &  82.60 & 77.91 & 91.00 & 83.95 \\
     &  & RITUAL  & 84.07  & 79.00 & \textbf{92.80} & 85.35 \\
     &  & \cc \textbf{Ours} &\cc \textbf{85.30} &\cc \textbf{84.18}  &\cc 86.93 &\cc \textbf{85.53}  \\
     \arrayrulecolor{gray}\cmidrule(lr){2-7}
      &  \multirow{5}{*}{Adversarial} & Regular & 75.80 & 70.41 & 89.00 & 78.62 \\
     &  & VCD  &  77.33 & 71.44 & 91.07 & 80.07 \\
     &  & M3ID  & 77.43  & 71.65 & 90.80 & 80.09 \\
     &  & RITUAL  & 78.00  & 71.72 & \textbf{92.47} & 80.78 \\
     &  & \cc \textbf{Ours} &\cc \textbf{81.43}   &\cc \textbf{78.61} &\cc 87.04 &\cc \textbf{82.61}    \\
    \bottomrule
    \end{tabular}
}
    % \vspace{-15pt}
\end{table}

\RebuttalRevision{
\subsection{Speeding Up Our Approach}
\label{subsec:speedup}
In this section, we propose two strategies to accelerate our approach: limiting the length of the initial response and reducing the number of inference steps in the diffusion process.
\begin{itemize}
    \item \textbf{Reducing Diffusion Inference Steps}. By default, we set the number of diffusion inference steps to 50 to ensure high-quality image generation. To improve the response generation speed, we can reduce the number of diffusion steps.
    In Table~\ref{tab:diffstep}, we report the performance on the CHAIR benchmark after reducing the diffusion inference steps in the model. By reducing the diffusion inference steps from 50 to 10, the average latency decreases by 2.85 seconds per instance, while the performance on CHAIR remains robust. This demonstrates that reducing the inference steps of the diffusion model is an effective way to speed up our approach.
    \item \textbf{Restricting Length of Initial Response}. Our method involves two queries to the LVLM for self-correcting decoding. To enhance efficiency, we can limit the length of the initial response. In Table~\ref{tab:initialnumber}, we present the efficiency and CHAIR performance results after decreasing the maximum token limit for the initial response. We can see that reducing the maximum number of tokens in the initial response from 128 to 96 decreases the latency by 0.72 seconds per instance while maintaining competitive performance. However, further reductions result in performance degradation, as a shorter initial response fails to adequately cover the entire scene, limiting its ability to generate an image that effectively reflects and mitigates hallucinations.
\end{itemize}
Note that these two strategies are not conflicting; instead, they are complementary. Setting the diffusion steps to 10 and limiting the maximum number of tokens in the initial response to 96 further reduces the inference latency to 10.21 seconds per instance while maintaining robust performance.
}

\begin{figure}[t]
\begin{minipage}[t]{0.49\linewidth}
\makeatletter\def\@captype{table}
\setlength\tabcolsep{2.8pt}
% \vspace{-50pt}
\caption{\RebuttalRevision{\textbf{Effect of reducing diffusion inference steps.}} }
\vspace{3.3pt}
    \label{tab:diffstep}
    \resizebox{\textwidth}{!}{
    \begin{tabular}{ccccc}
        \toprule
        Diff. Steps & Avg. Latency $\downarrow$ & CHAIR$_S$ $\downarrow$ & CHAIR$_I$ $\downarrow$\\
        \midrule
        50 & 13.89 s & 48.8 & 14.6  \\
        30 & 12.56 s & 48.9 & 14.7 \\
        20 & 11.87 s & 49.2 & 14.8  \\
        10 & 11.04 s & 48.8 & 14.9  \\
        \bottomrule
    \end{tabular}
    }
\end{minipage}
\hspace{2pt}
\begin{minipage}[t]{0.49\linewidth}
\makeatletter\def\@captype{table}
\setlength\tabcolsep{3pt}
\caption{\RebuttalRevision{\textbf{Effect of restricting the number of tokens in the initial response.}}}
    \vspace{4pt}
    \label{tab:initialnumber}
    \resizebox{\textwidth}{!}{
    \begin{tabular}{ccccc}
        \toprule
        \# Tokens & Avg. Latency $\downarrow$ & CHAIR$_S$ $\downarrow$ & CHAIR$_I$ $\downarrow$\\
        \midrule
        128 & 13.89 s & 48.8 & 14.6   \\
        96 & 13.17 s & 48.8 & 14.9  \\
        64 & 12.20 s & 49.5 &  14.8  \\
        32 & 11.33 s & 51.2 & 14.9  \\
        \bottomrule
    \end{tabular}
}
\end{minipage}
\vspace{-3pt}
\end{figure}

\RebuttalRevision{
\subsection{Quantitative Assessment of Generated Image Quality}
\label{subsec:quant_diffusion}
Our approach incorporates a text-to-image generation model to mitigate hallucinations. We evaluate the quality of the generated images on all 4 subsets on the MME benchmark using CLIPScore~\citep{hessel2021clipscore}. Specifically, we utilize the CLIP backbone with ViT-B/32 backbone for our evaluation. We list the results in Table~\ref{tab:mme_clipscores}. As we can see from the table, our text-to-image generative model (specifically, SD-v1.5) achieves an average CLIPScore of over 30 across all subsets. For comparison, the advanced DALL-E 3 model achieves a score of 32.0, while DALL-E 2 achieves 31.4.\footnote{These results are sourced from the technical report on DALL-E 3, available at: \url{https://cdn.openai.com/papers/dall-e-3.pdf}.} These results highlight the capability of our model to generate high-quality images that closely align with the initial response.
}

\begin{table}[t]
\centering
\caption{\RebuttalRevision{\textbf{CLIPScore evaluation across different MME subsets}}.}
\vspace{5pt}
\begin{tabular}{lcccc}
\toprule
\textbf{MME Subset}     & \textbf{Existence} & \textbf{Count} & \textbf{Position} & \textbf{Color} \\ \midrule
\textbf{Avg. CLIPScore} & 31.34              & 30.69          & 30.09             & 31.69          \\ \bottomrule
\end{tabular}
\label{tab:mme_clipscores}
\end{table}

\section{More Case Studies}
\label{sec:case}
\subsection{Details about GPT-4V-Aided Evaluation}
Following VCD~\citep{leng2024mitigating}, we use GPT-4V to evaluate responses in open-ended generation scenarios, scoring them based on accuracy and detailedness. Leveraging the strong human-like capabilities of GPT-4V, it can detect incorrect colors, positions, and relationships, providing a comprehensive evaluation of the responses.
Specifically, we apply the prompt provided in Table~\ref{tab:prompt_evaluation} to instruct GPT-4V to rate the two responses on a scale of 1 to 10 for both accuracy and detailedness:
\begin{itemize}
    \item \textbf{Accuracy} measures the consistency between the responses/descriptions generated by the LVLMs and the given image. A lower score is assigned if GPT-4V detects any inconsistencies in the content of the responses.
    \item \textbf{Detailedness} evaluates the depth and specificity of the responses provided by the LVLMs. A higher score is awarded if the response includes comprehensive descriptions, captures fine-grained details of the image, and provides well-elaborated explanations. Conversely, a lower score is given if the response is vague or lacks sufficient detail.
\end{itemize}

\begin{table*}[t]\centering
\begin{minipage}{0.95\textwidth}
%\vspace{0mm}    
\centering
\begin{tcolorbox} 
    \centering
   
     %\hspace{-4mm}
      \small
    \begin{tabular}{p{0.95\textwidth}} \hline \\
   \textbf{Description:} \\    
   
   AI that scores image description accuracy and detailedness.

   \\ \midrule

   \textbf{Instructions:} \\   
   
You are an AI designed to evaluate and score the performance of two AI assistants in describing a given image. Your primary focus is on the accuracy and detailedness of their descriptions. You will assess the accuracy by checking for hallucinations - any part of the description that is inconsistent with the image content. For detailedness, you will consider how rich the response is in necessary details, excluding any hallucinated parts. You will provide scores on a scale from 1 to 10 for each assistant separately, based on these criteria. After scoring, you will offer an explanation for your evaluation, ensuring it is free from bias and not influenced by the order of presentation of the responses.
\\ \\
Input format: \\ \\
\lbrack{}Assistant 1\rbrack{}\\
 \{Response 1\}  \\
\lbrack{}End of Assistant 1\rbrack{} \\
\\
\lbrack{}Assistant 2\rbrack{} \\
 \{Response 2\}\\
\lbrack{}End of Assistant 2\rbrack{} \\
\\
Output format:\\
\\
Accuracy:\\
Scores of the two answers:\\
Reason:\\
\\
Detailedness:\\
Scores of the two answers:\\
Reason:\\ \\

\bottomrule
    \end{tabular}
\end{tcolorbox}
\caption{\textbf{GPT-4V-aided evaluation setup}. We present the prompt we provided to GPT-4V to evaluate the LVLM responses based on accuracy and detailedness.}
\label{tab:prompt_evaluation}
\end{minipage}
\end{table*}

\subsection{More Qualitative Results}
\begin{figure}[t]

\includegraphics[width=\linewidth]{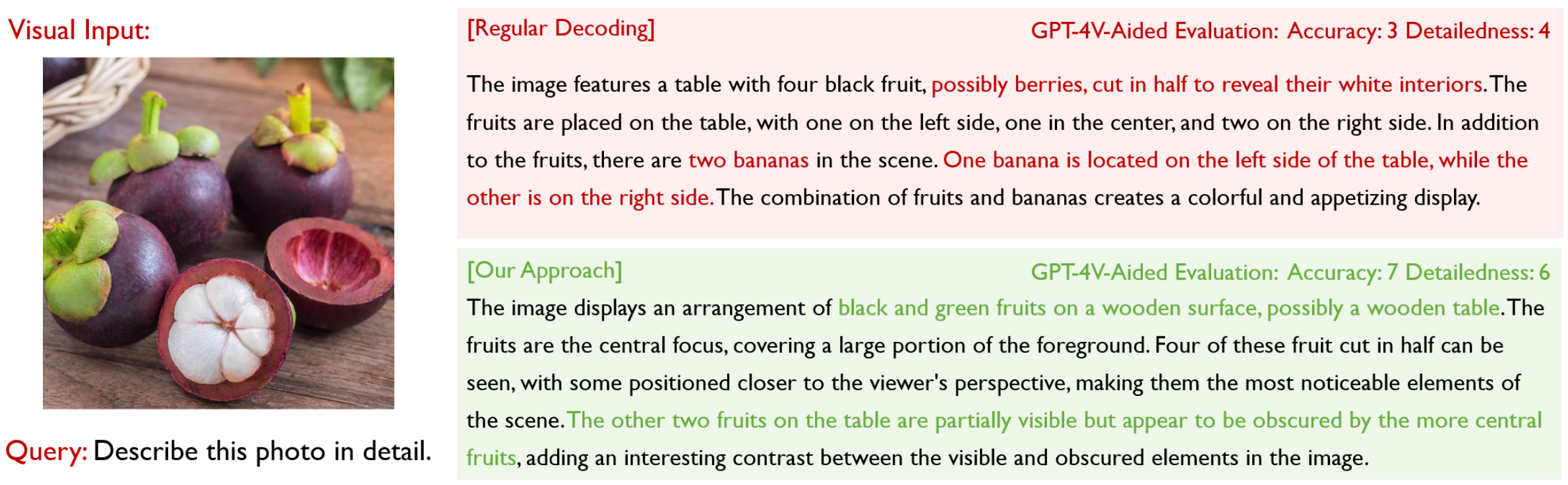}
\\[10pt]
\includegraphics[width=\linewidth]{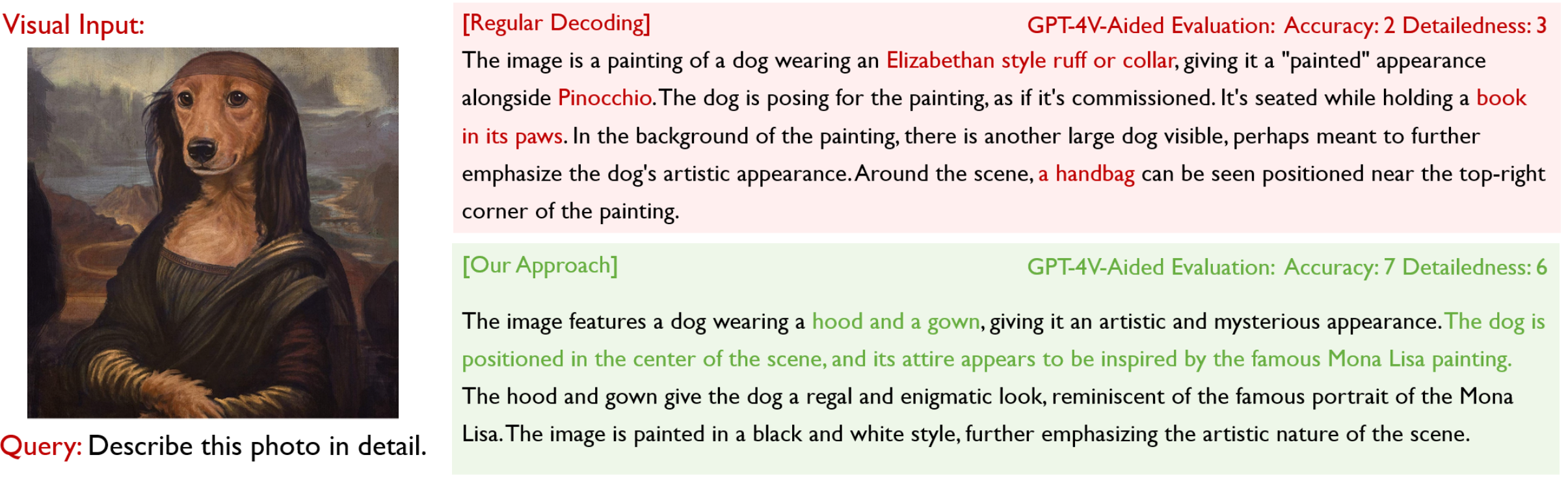}
\\[10pt]
\includegraphics[width=\linewidth]{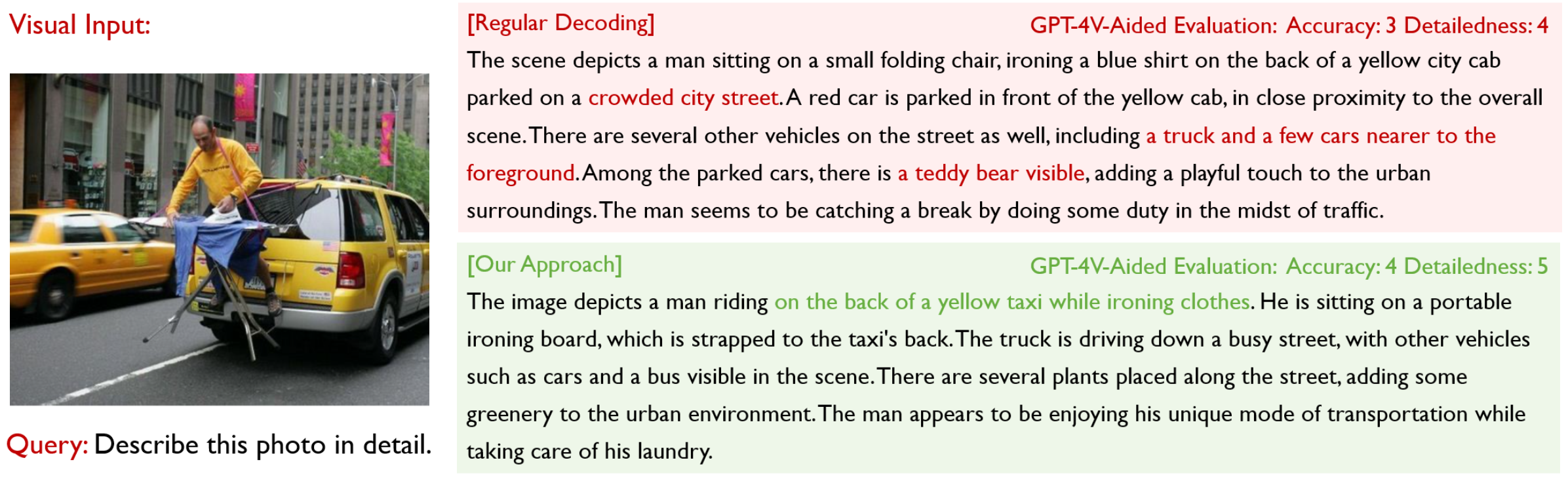}
\\[10pt]
\includegraphics[width=\linewidth]{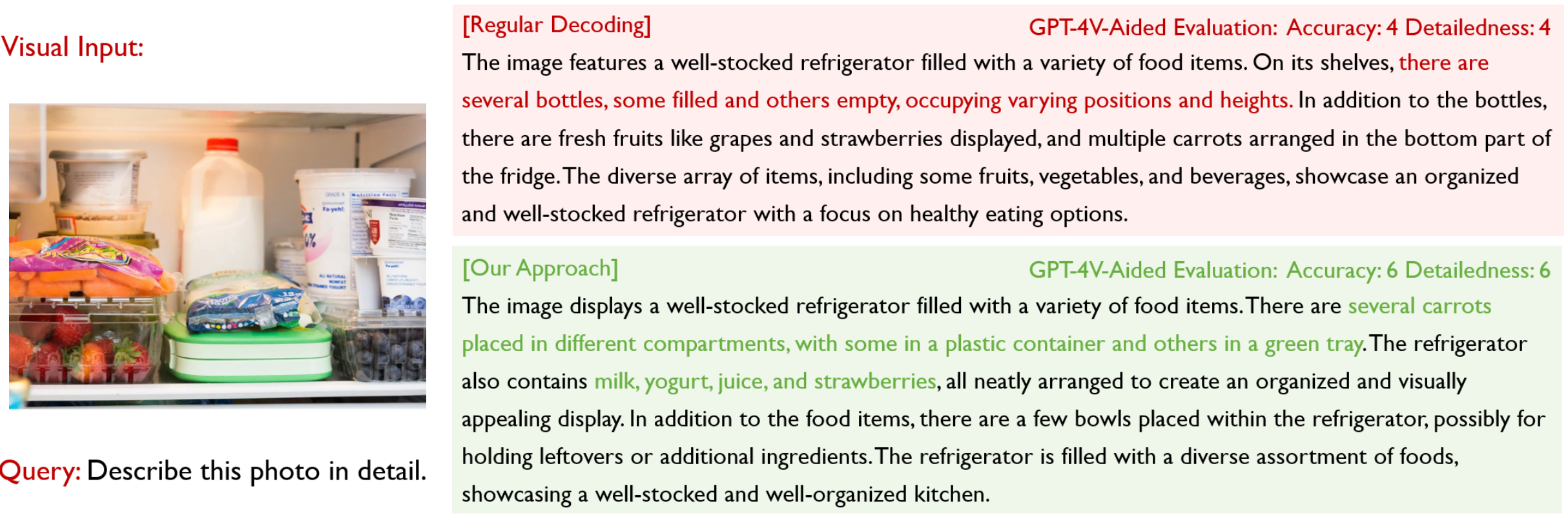}

\caption{\textbf{Case studies on the LLaVA-Bench benchmark}. We compare the responses generated by regular decoding and our method using LLaVA-1.5. GPT-4V-aided evaluation results are also provided alongside the responses. Hallucinated and accurate content is highlighted in \textcolor{darkred}{red} and \textcolor{darkgreen}{green}.} 

\label{fig:llavabench2}
\end{figure}

\begin{figure}[t]

\includegraphics[width=\linewidth]{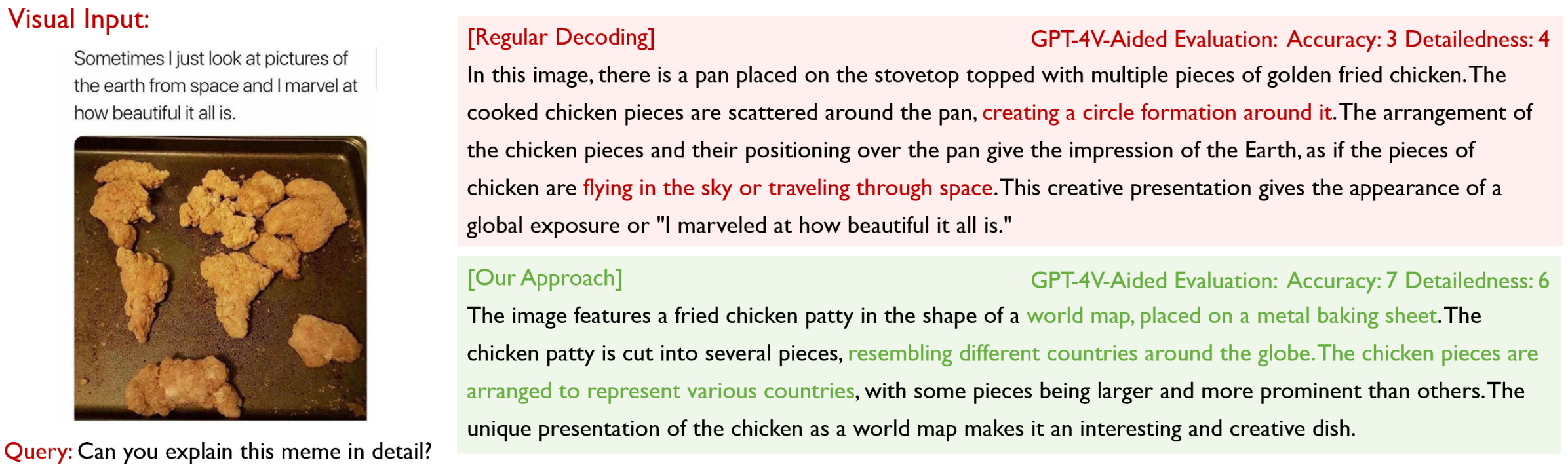}
\\[10pt]
\includegraphics[width=\linewidth]{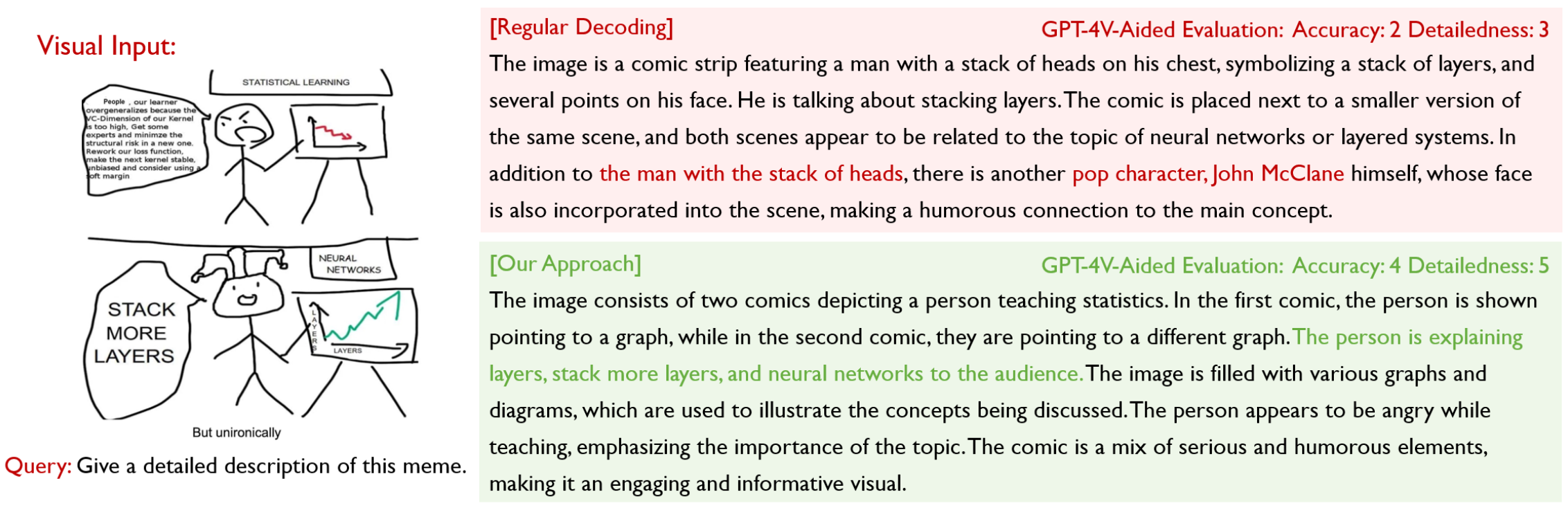}
\\[10pt]
\includegraphics[width=\linewidth]{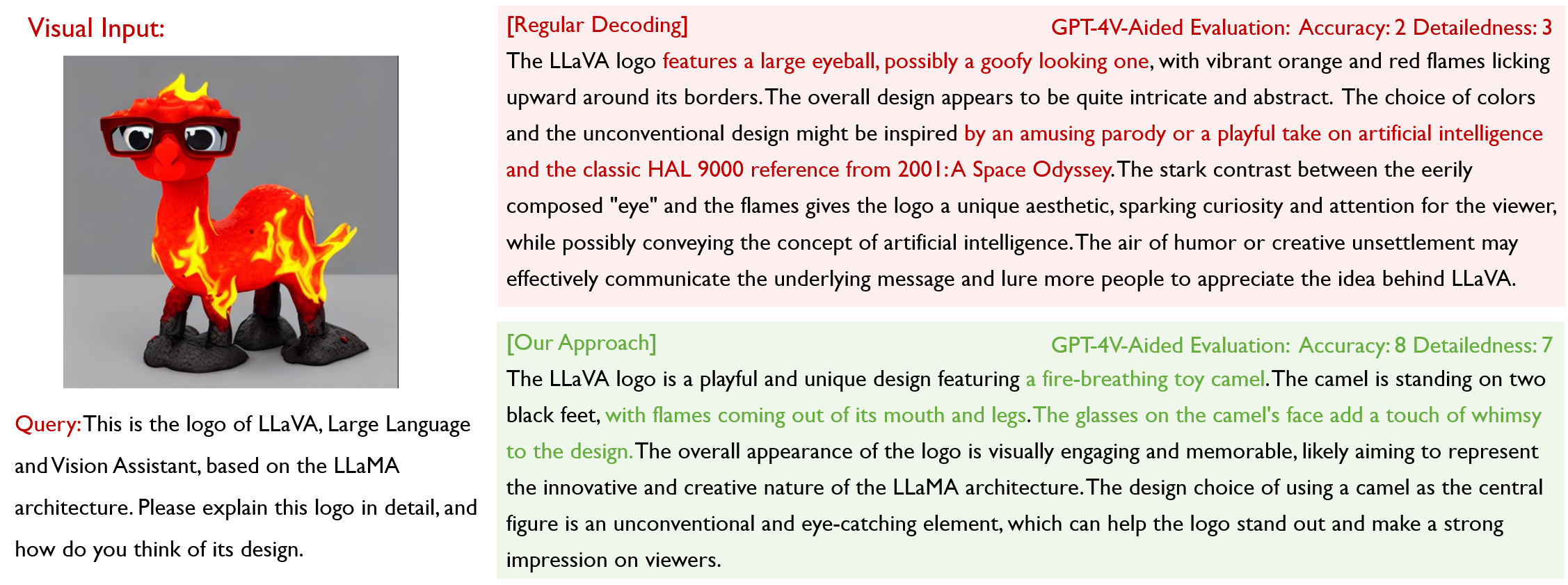}
\\[10pt]
\includegraphics[width=\linewidth]{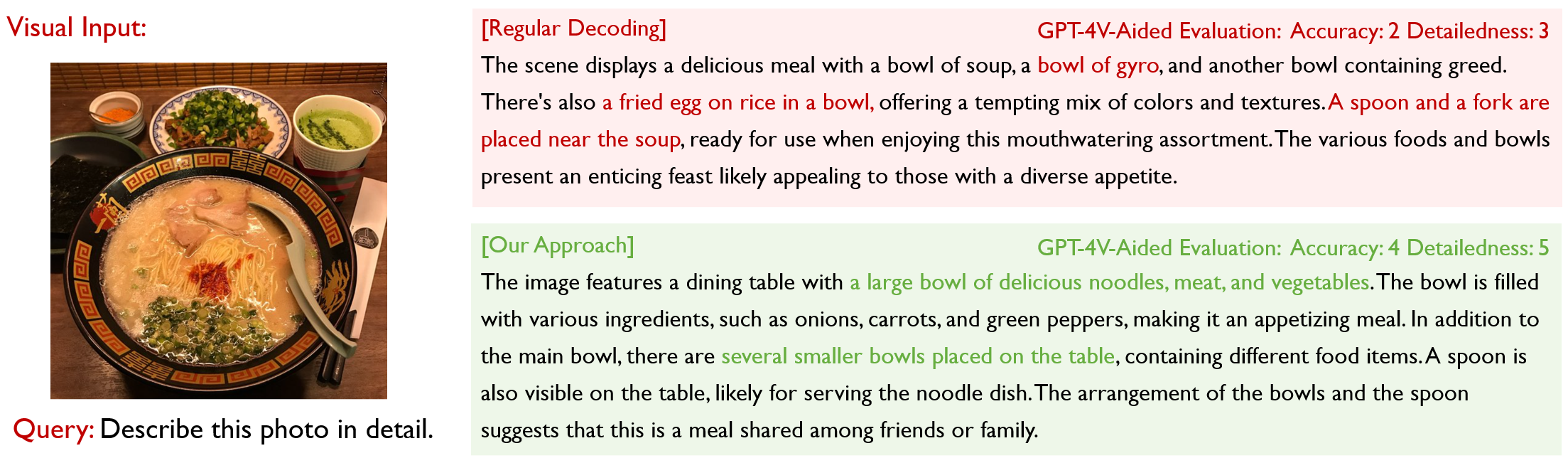}

\caption{\textbf{Case studies on the LLaVA-Bench benchmark}. We compare the responses generated by regular decoding and our method using LLaVA-1.5. GPT-4V-aided evaluation results are also provided alongside the responses. Hallucinated and accurate content is highlighted in \textcolor{darkred}{red} and \textcolor{darkgreen}{green}.} 

\label{fig:llavabench3}
\end{figure}

In Figure~\ref{fig:llavabench2} and Figure~\ref{fig:llavabench3}, we provide additional case studies on LLaVA-Bench to qualitatively demonstrate the effectiveness of our methods in mitigating hallucinations. We also included GPT-4V evaluations of accuracy and detailedness scores for each instance.

\begin{figure}[t]
\centering
\includegraphics[width=0.92\linewidth]{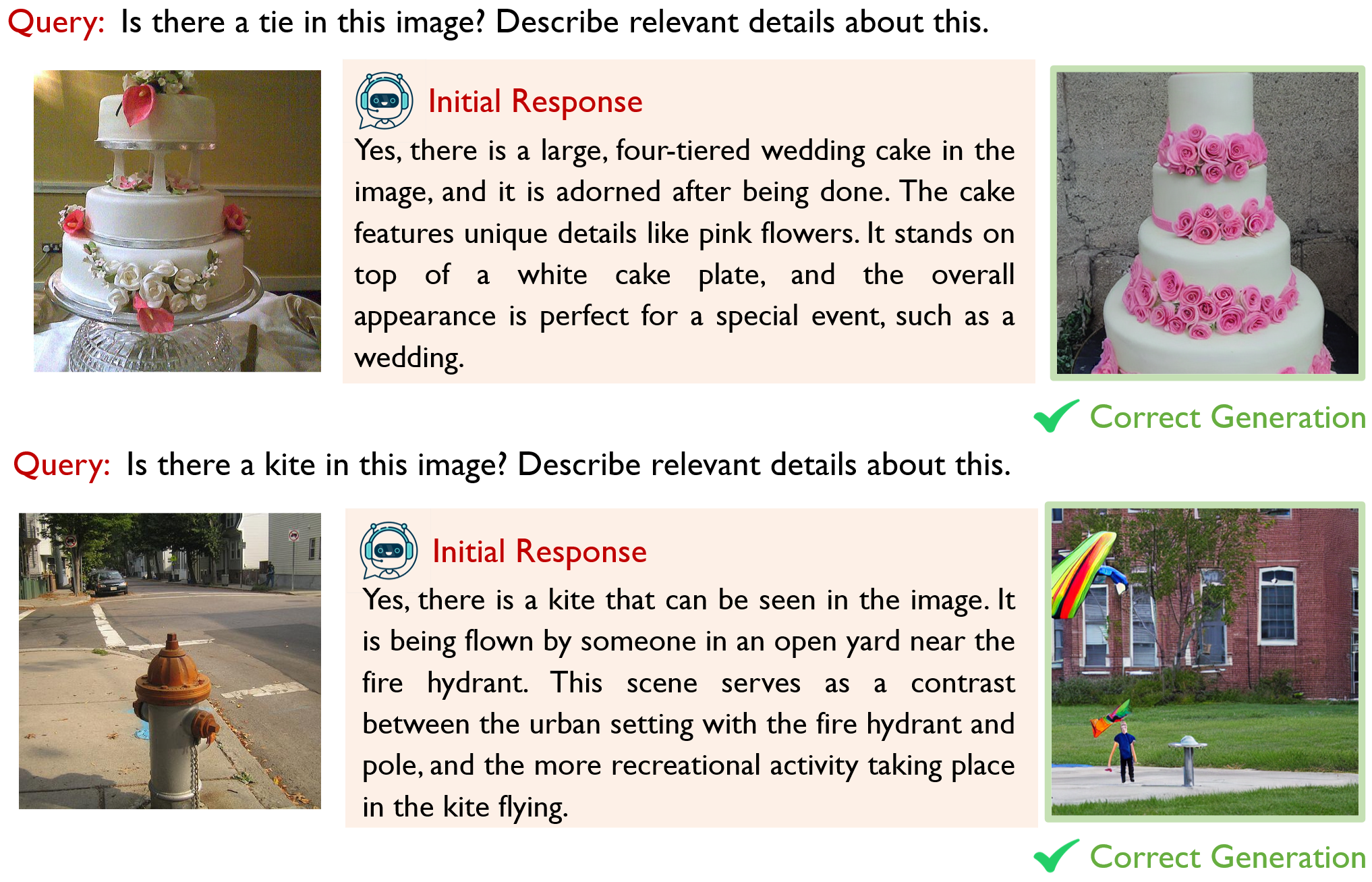}
\includegraphics[width=0.92\linewidth]{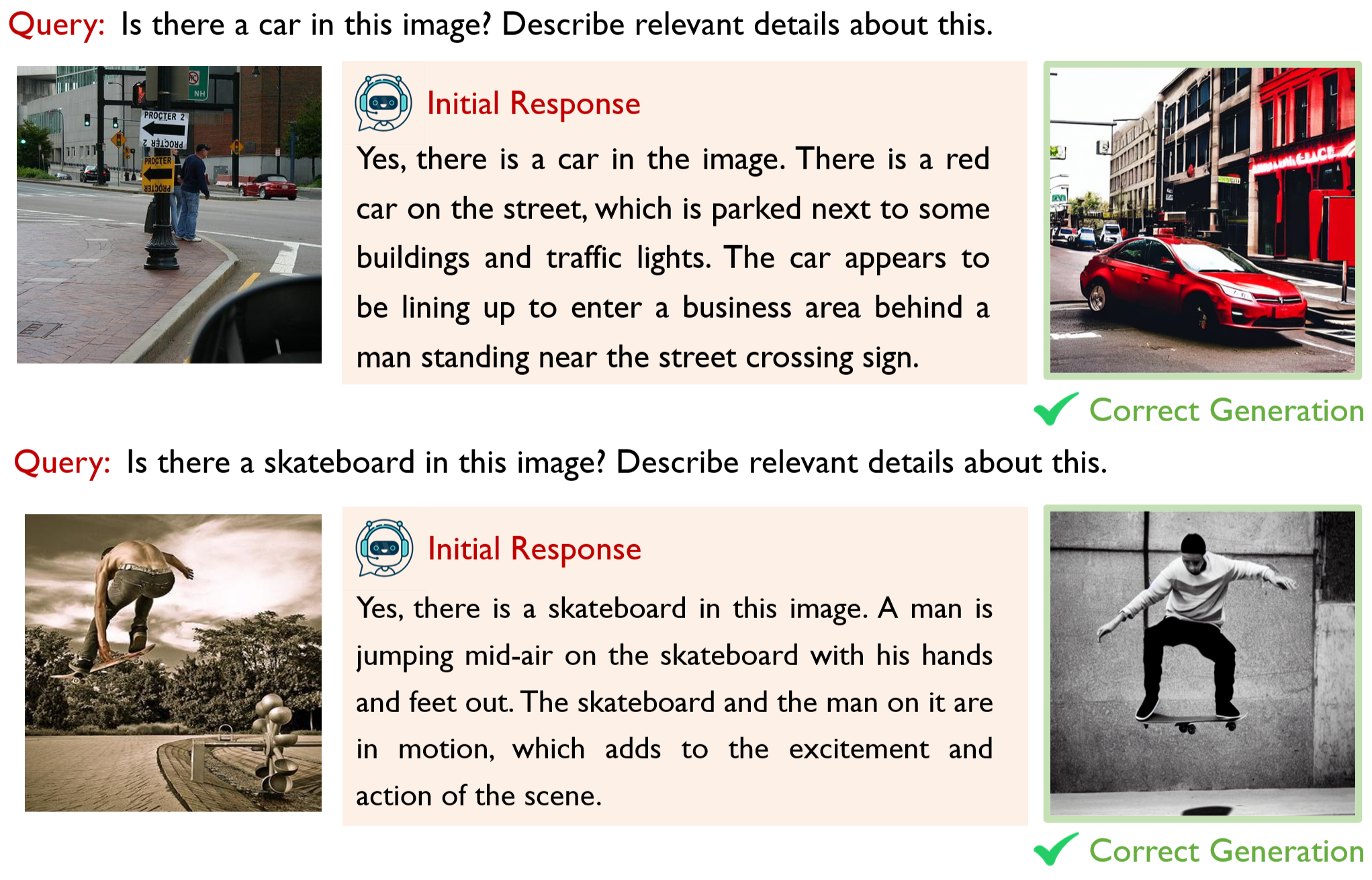}
\\[5pt]
\includegraphics[width=0.95\linewidth]{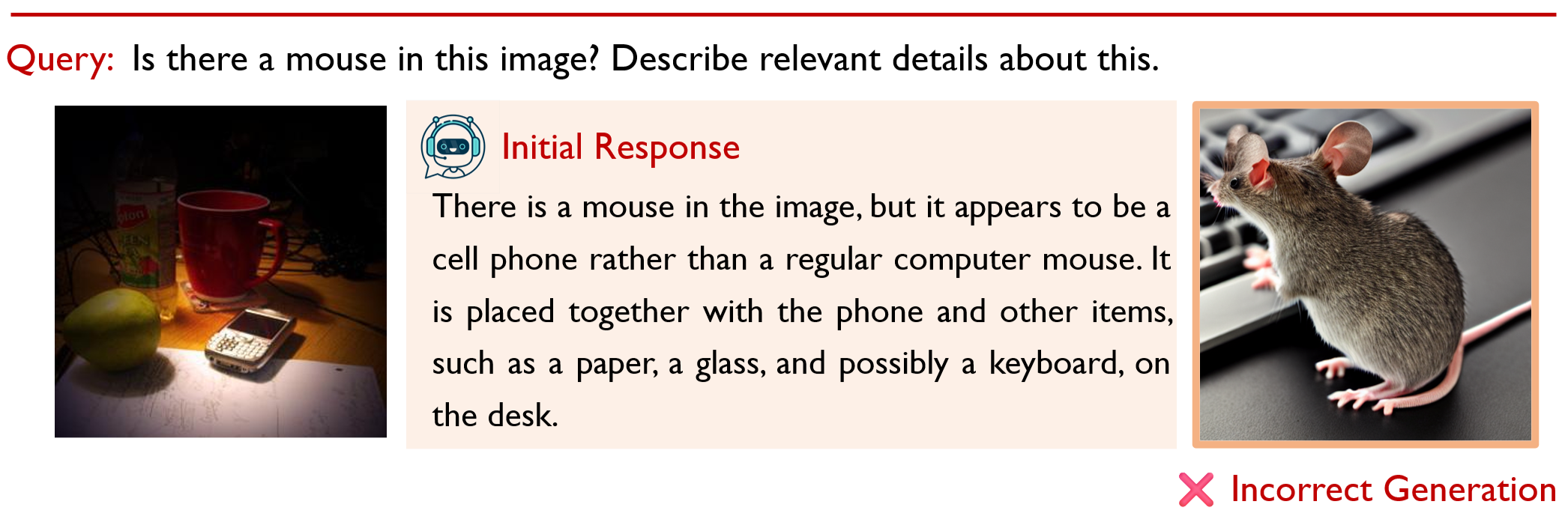}
\vspace{-10pt}
\caption{\textbf{Qualitative evaluation of images generated by the generative model on the \textit{existence} subset of the MME benchmark.} Specifically, the left displays the original image input, the middle presents the initial response generated by the LVLMs, and the right shows the image generated based on this response. This figure showcases four success cases and one failure case of our Diffusion models in generating high-quality images that align with the initial response.}
\label{fig:diffusion1}
\end{figure}

\begin{figure}[t]
\centering
\includegraphics[width=0.92\linewidth]{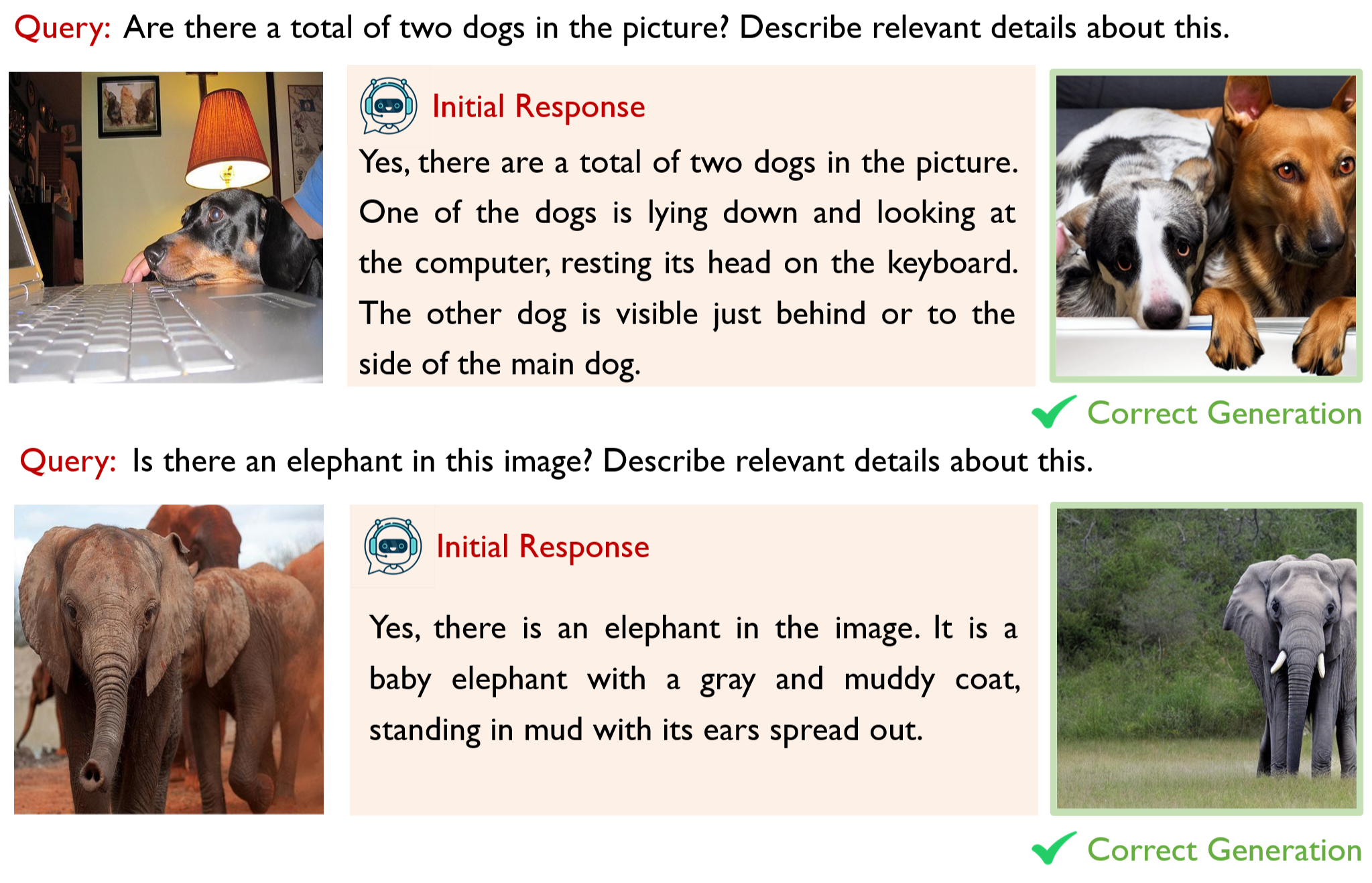}
\includegraphics[width=0.92\linewidth]{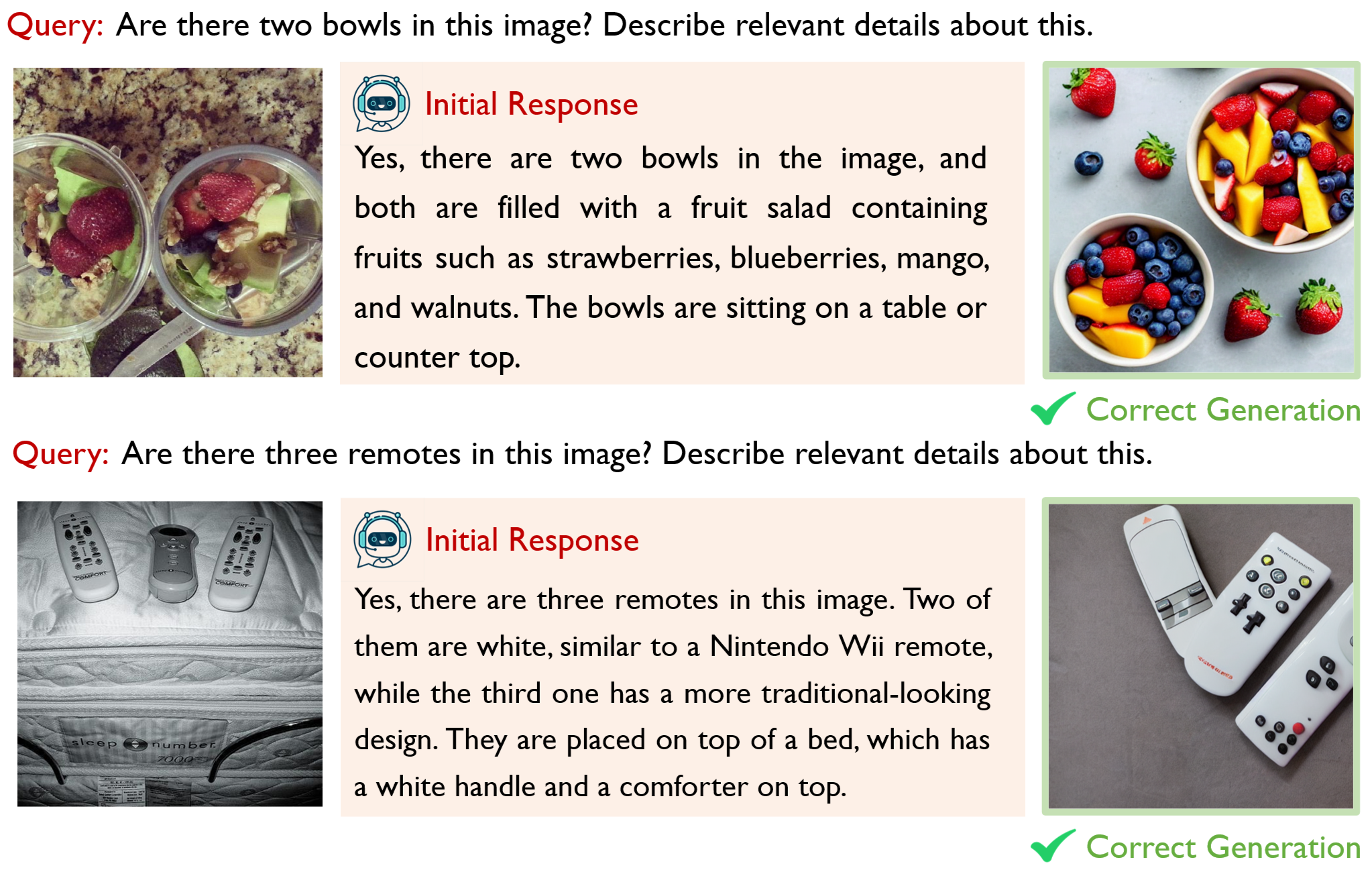}
\\[5pt]
\includegraphics[width=0.95\linewidth]{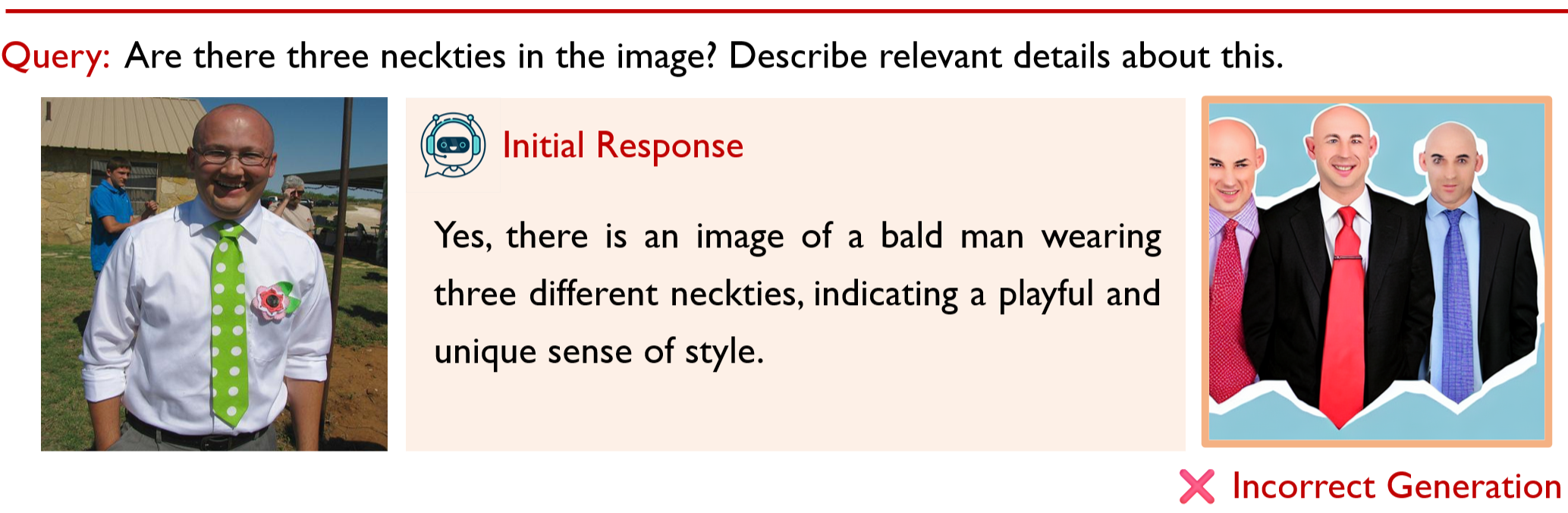}
\vspace{-10pt}
\caption{\textbf{Qualitative evaluation of images generated by the generative model on the \textit{count} subset of the MME benchmark.} Specifically, the left displays the original image input, the middle presents the initial response generated by the LVLMs, and the right  shows the image generated based on this response. This figure showcases four success cases and one failure case of our Diffusion models in generating high-quality images that align with the initial response.}
\label{fig:diffusion2}
\end{figure}

\begin{figure}[t]
\centering
\includegraphics[width=0.92\linewidth]{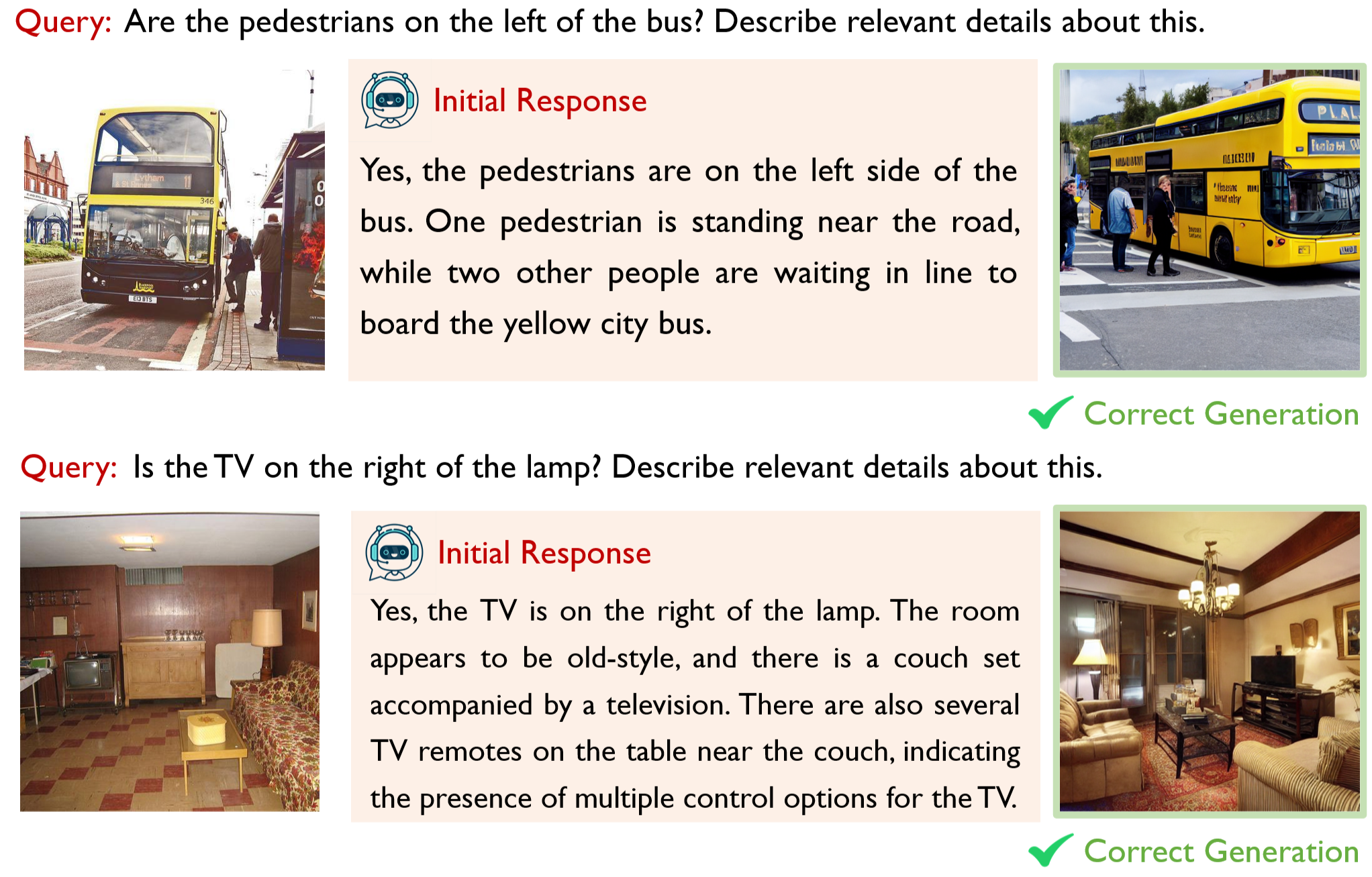}
\includegraphics[width=0.92\linewidth]{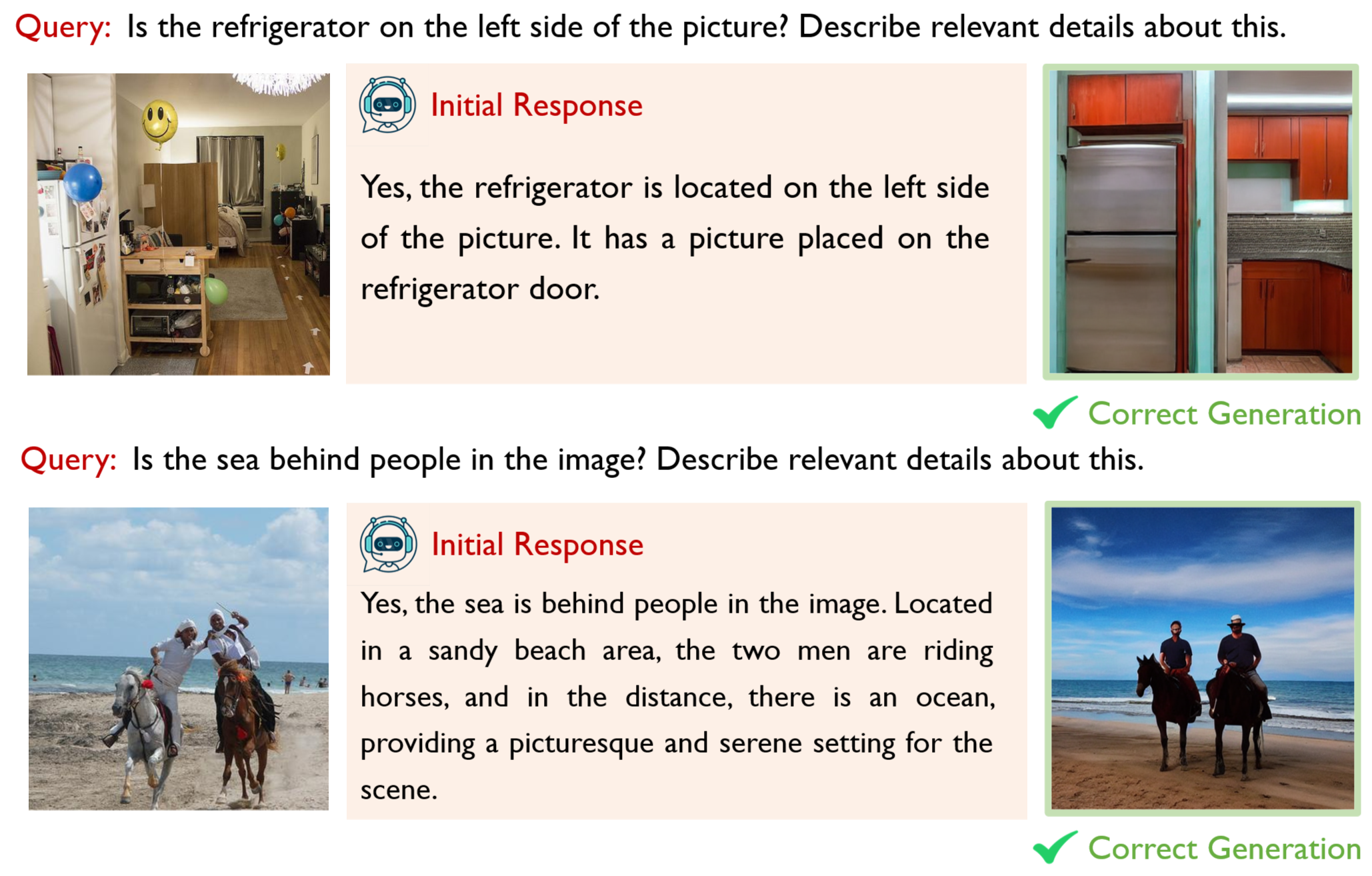}
\\[5pt]
\includegraphics[width=0.95\linewidth]{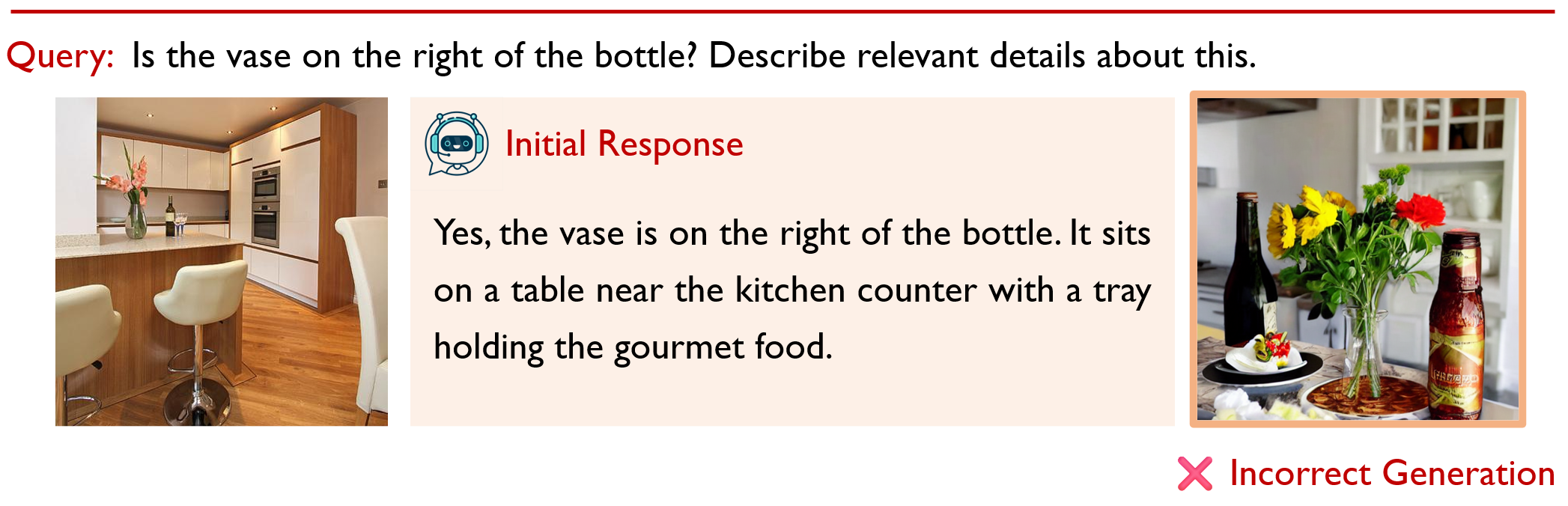}
\vspace{-10pt}
\caption{\textbf{Qualitative evaluation of images generated by the generative model on the \textit{position} subset of the MME benchmark.} Specifically, the left displays the original image input, the middle presents the initial response generated by the LVLMs, and the right  shows the image generated based on this response. This figure showcases four success cases and one failure case of our Diffusion models in generating high-quality images that align with the initial response.}
\label{fig:diffusion3}
\end{figure}

\begin{figure}[t]
\centering
\includegraphics[width=0.92\linewidth]{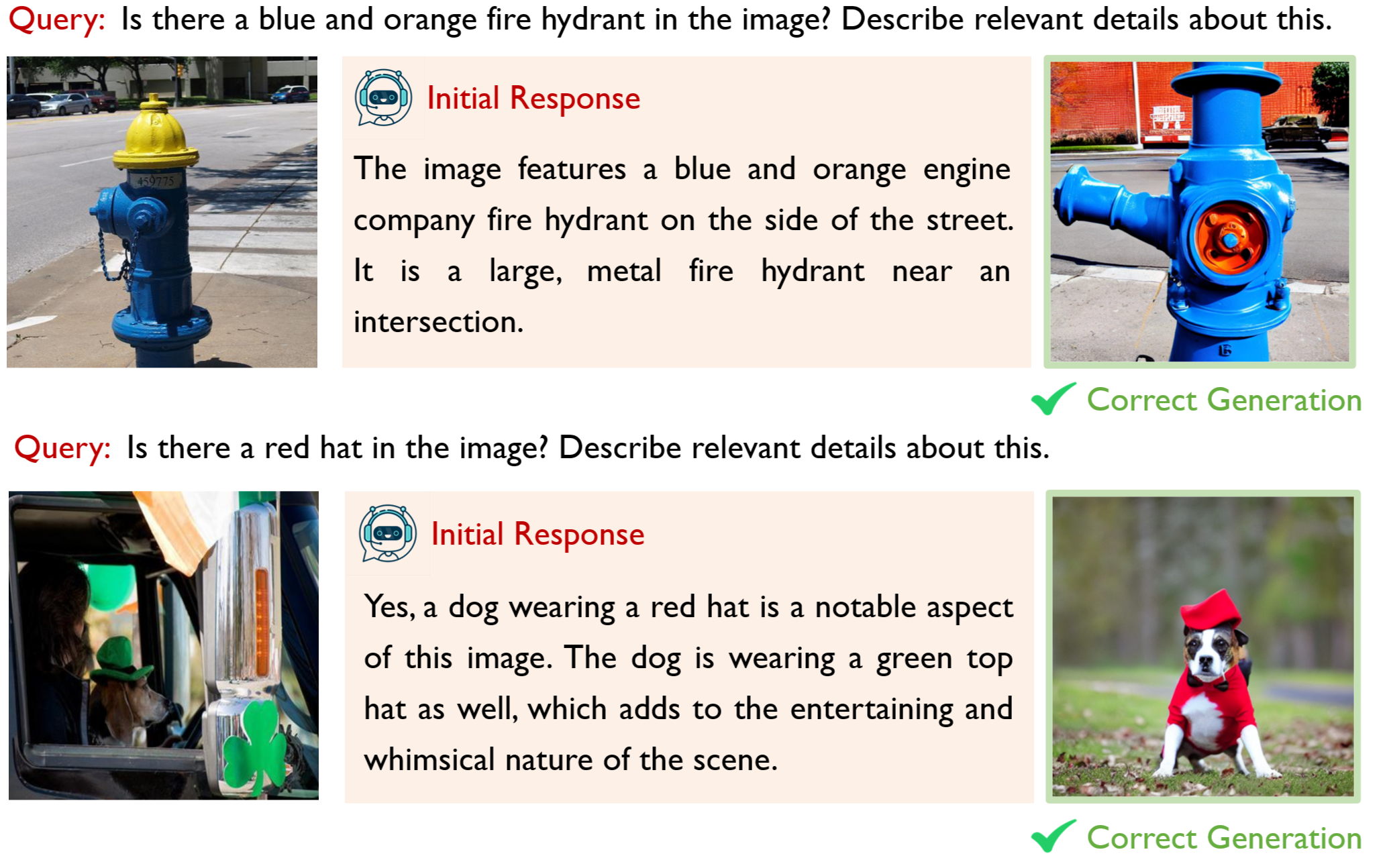}
\includegraphics[width=0.92\linewidth]{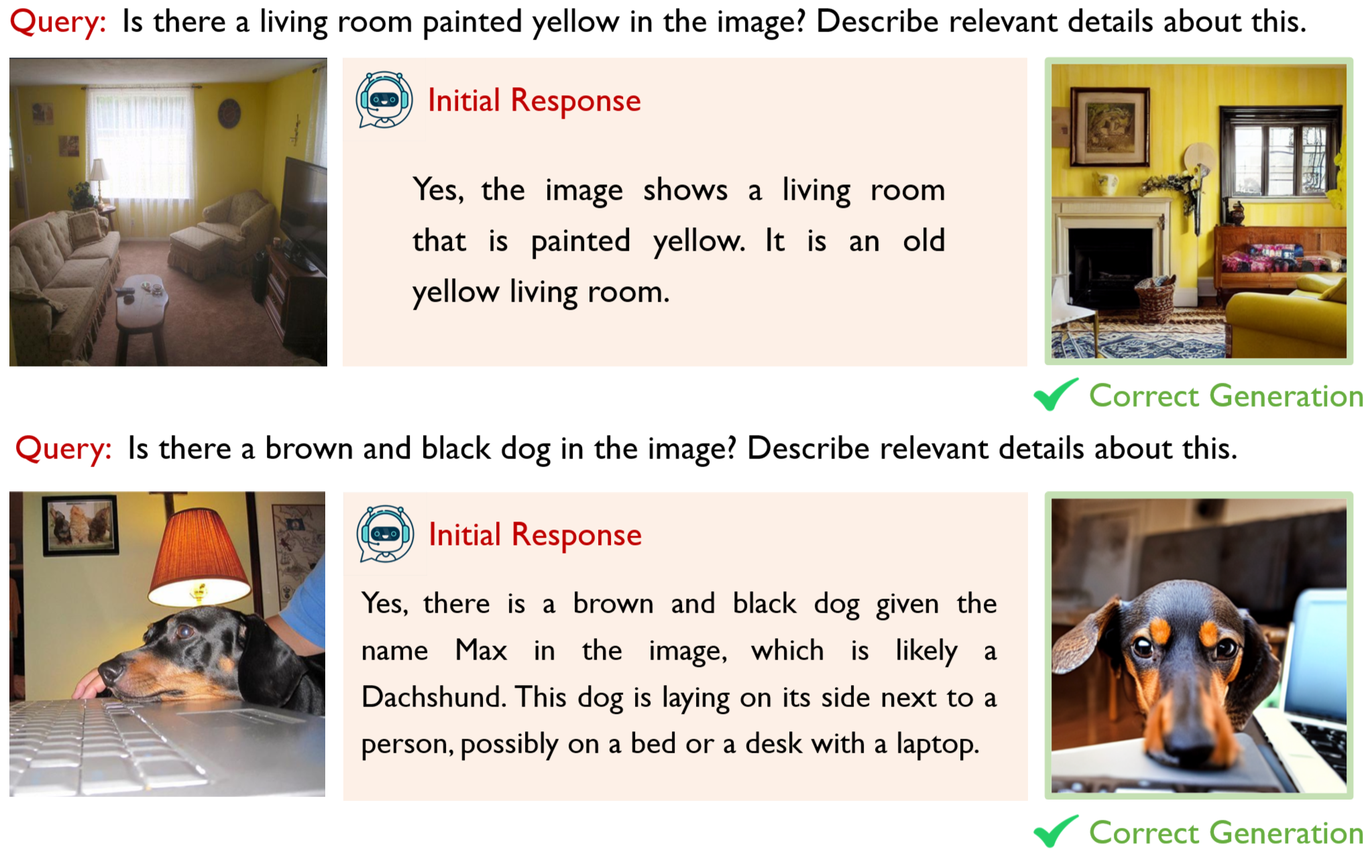}
\\[5pt]
\includegraphics[width=0.95\linewidth]{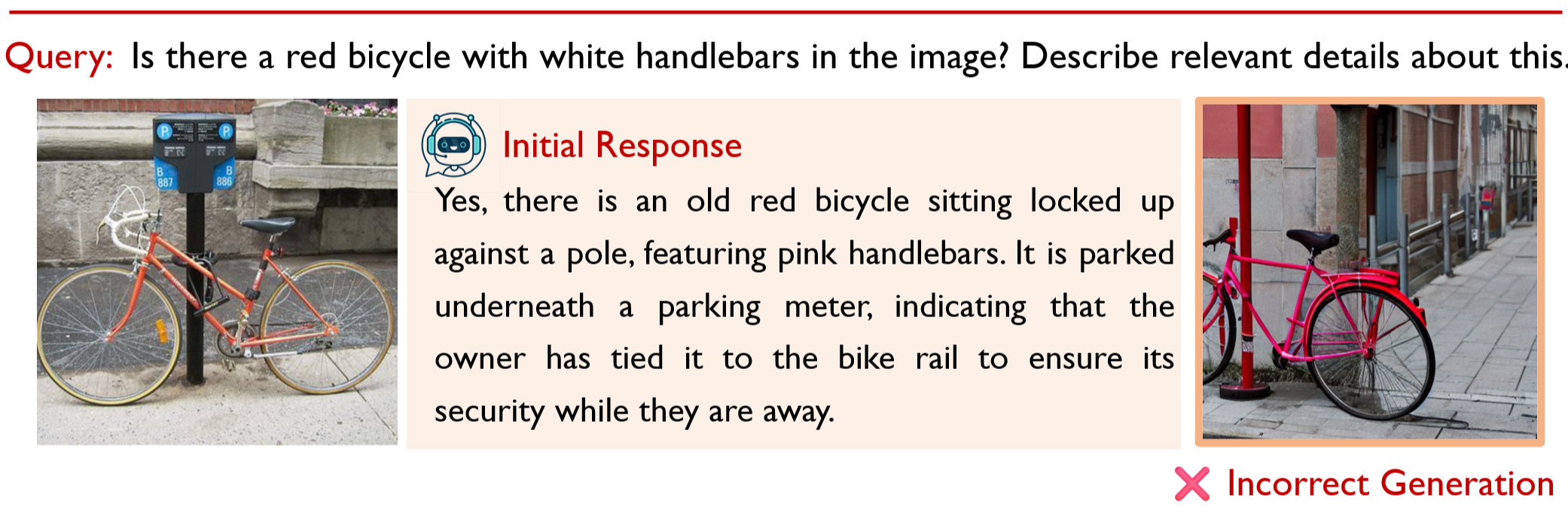}
\vspace{-10pt}
\caption{\textbf{Qualitative evaluation of images generated by the generative model on the \textit{color} subset of the MME benchmark.} Specifically, the left displays the original image input, the middle presents the initial response generated by the LVLMs, and the right  shows the image generated based on this response. This figure showcases four success cases and one failure case of our Diffusion models in generating high-quality images that align with the initial response.}
\label{fig:diffusion4}
\end{figure}

\RebuttalRevision{
In Figure~\ref{fig:diffusion1}-\ref{fig:diffusion4}, we provide qualitative evaluations of the images generated by the generative model, including both success and failure cases, across all four subsets of the MME benchmark to better understand the effectiveness of the generative models. Our results show that, despite occasional failure cases, the generative model consistently produces high-quality and realistic images that accurately visualize the initial response, providing effective self-feedback.
}

\RebuttalRevision{
\section{Future Work}
\label{sec:future}
In future work, we aim to extend the evaluation of our method to a broader range of LVLMs, such as Mini-GPT4~\citep{zhu2024minigpt} and mPLUG-Owl2~\citep{ye2024mplug}, as well as additional benchmarks, including R-Bench~\citep{wu2024evaluating}, which focuses on relation hallucination, and ROPE~\citep{chen2024multiobject}, which addresses multiple-object hallucination. This expanded evaluation will allow us to more comprehensively assess the generalizability and effectiveness of our approach across diverse models and tasks.}

\RebuttalRevision{
Furthermore, we plan to investigate integrating generative feedback directly into the instruction tuning phase. This integration has the potential to eliminate the computational overhead associated with applying our method during inference, thereby significantly improving efficiency without compromising performance. By pursuing these directions, we hope to further enhance the practical applicability and scalability of our approach.
}

\end{document}